\documentclass[twoside]{article}

\newif\ifaos \aosfalse

\ifaos
\usepackage[dvips,aos]{imsart}
\fi

\usepackage[numbers]{natbib}
\RequirePackage[OT1]{fontenc}
\RequirePackage{amsthm,amsmath,amsfonts}
\RequirePackage[colorlinks]{hyperref}
\RequirePackage{hypernat}

\RequirePackage{epsfig, graphicx, color}
\RequirePackage{times}
\RequirePackage{latexsym}
\RequirePackage{psfrag}

\usepackage{fullpage}

\ifaos
\startlocaldefs

\def\tunpar{\tau}
\def\sctail{f}
\def\scthresh{v_*}
\def\scdelta{\delta}

\newcommand{\sctailInvN}{\ensuremath{{\widebar{\numobs}_f}}}
\newcommand{\sctailInvT}{\ensuremath{\widebar{\delta}_f}}

\def\CovEstim{\hat{\CovHat}}

\def\rhoStar{\rho^{*}}
\def\SampVar{X^{(\obsind)}}
\def\SampVarStar{X^{*(\obsind)}}
\def\sampdev{W}
\def\momentpow{m}
\def\KconMom{K_{\momentpow}}

\def\csubgexpr{128(1 + 4\csubg^{2})^{2}\max_{i} (\CovMatStar_{ii})^2}

\def\csubgexprsqrtmultsix{48\sqrt{2} \,(1 + 4 \csubg^2) \,\max_{i} (\CovMatStar_{ii}) }
\def\csubgexprsqrtmulttwo{16\sqrt{2} \,(1 + 4 \csubg^2) \,\max_{i} (\CovMatStar_{ii}) }

\newcommand{\SigHat}{\ensuremath{\widehat{\Sigma}}}

\newcommand{\ThetaStar}{\ensuremath{\Theta^*}}
\newcommand{\ThetaHat}{\ensuremath{\widehat{\Theta}}}

\newcommand{\EdgeSet}{\ensuremath{E}}

\newcommand{\regpar}{\ensuremath{\lambda_\numobs}}

\newcommand{\EsetPlus}{\ensuremath{S}}
\newcommand{\EsetPlusComp}{\ensuremath{\EsetPlus^c}}

\newcommand{\Zhat}{\ensuremath{\hat Z}}

\newcommand{\ellreg}[1]{\ensuremath{\|#1\|_{1,\operatorname{off}}}}

\newlength{\widebarargwidth}
\newlength{\widebarargheight}
\newlength{\widebarargdepth}
\DeclareRobustCommand{\widebar}[1]{%
  \settowidth{\widebarargwidth}{\ensuremath{#1}}%
  \settoheight{\widebarargheight}{\ensuremath{#1}}%
  \settodepth{\widebarargdepth}{\ensuremath{#1}}%
  \addtolength{\widebarargwidth}{-0.3\widebarargheight}%
  \addtolength{\widebarargwidth}{-0.3\widebarargdepth}%
  \makebox[0pt][l]{\hspace{0.3\widebarargheight}%
    \hspace{0.3\widebarargdepth}%
    \addtolength{\widebarargheight}{0.3ex}%
    \rule[\widebarargheight]{0.95\widebarargwidth}{0.1ex}}%
  {#1}}

\makeatletter
\long\def\@makecaption#1#2{
        \vskip 0.8ex
        \setbox\@tempboxa\hbox{\small {\bf #1:} #2}
        \parindent 1.5em  
        \dimen0=\hsize
        \advance\dimen0 by -3em
        \ifdim \wd\@tempboxa >\dimen0
                \hbox to \hsize{
                        \parindent 0em
                        \hfil
                        \parbox{\dimen0}{\def\baselinestretch{0.96}\small
                                {\bf #1.} #2
                                }
                        \hfil}
        \else \hbox to \hsize{\hfil \box\@tempboxa \hfil}
        \fi
        }
\makeatother

\newcommand{\nn}{\nonumber} 
\newcommand{\trs}[2]{\left< #1 , #2 \right>}
\newcommand{\beq}{\begin{eqnarray*}}
\newcommand{\eeq}{\end{eqnarray*}}
\newcommand{\beqn}{\begin{eqnarray}}
\newcommand{\eeqn}{\end{eqnarray}}

\newcommand{\Rem}{\ensuremath{R}}

\newcommand{\Ztil}{\ensuremath{\wtil{Z}}}
\let\hat\widehat

\newcommand{\degmax}{\ensuremath{\nodedeg}}
\newcommand{\nodedeg}{\ensuremath{d}}
\newcommand{\myparagraph}[1]{\noindent \paragraph{#1}}

\newcommand{\pdim}{\mdim}

\newcommand{\CovMat}{\ensuremath{\Sigma}}

\newcommand{\Tail}{\ensuremath{\mathcal{T}}}

\newcommand{\numobs}{\ensuremath{n}}
\newcommand{\spindex}{\ensuremath{s}}
\newcommand{\spdim}{\ensuremath{s}}
\newcommand{\mdim}{\ensuremath{p}}

\theoremstyle{plain}
\newtheorem{theo}{Theorem}[section]

\newtheorem{lem}{Lemma}[section]
\newtheorem{prop}{Proposition}[section]
\newtheorem{cor}{Corollary}[section]

\theoremstyle{definition} 

\newtheorem{nota}{Notation}[section]
\newtheorem{de}{Definition}[section]
\newtheorem{exa}{Example}[section]
\newtheorem{as}{Assumption}[section]
\newtheorem{alg}{Algorithm}[section]

\newcommand{\btheo}{\begin{theo}}
\newcommand{\bde}{\begin{de}}
\newcommand{\ble}{\begin{lem}}
\newcommand{\bpr}{\begin{prop}}
\newcommand{\bno}{\begin{nota}}
\newcommand{\bex}{\begin{exa}}
\newcommand{\bcor}{\begin{cor}}
\newcommand{\spro}{\begin{proof}}
\newcommand{\bas}{\begin{as}}
\newcommand{\balg}{\begin{alg}}

\newcommand{\etheo}{\end{theo}}
\newcommand{\ede}{\end{de}}
\newcommand{\ele}{\end{lem}}
\newcommand{\epr}{\end{prop}}
\newcommand{\eno}{\end{nota}}
\newcommand{\eex}{\end{exa}}
\newcommand{\ecor}{\end{cor}}
\newcommand{\fpro}{\end{proof}}
\newcommand{\eas}{\end{as}}
\newcommand{\ealg}{\end{alg}}

\theoremstyle{plain}

\newtheorem{theos}{Theorem}

\newtheorem{lems}{Lemma}

\newtheorem{cors}{Corollary}

\theoremstyle{definition}
\newtheorem{exas}{Example}
\newtheorem{algs}{Algorithm}
\newtheorem{asss}{Assumption}
\newtheorem{defns}{Definition}

\newcommand{\btheos}{\begin{theos}}
\newcommand{\etheos}{\end{theos}}
\newcommand{\bprops}{\begin{proposition}}
\newcommand{\eprops}{\end{proposition}}
\newcommand{\bdes}{\begin{defns}}
\newcommand{\edes}{\end{defns}}
\newcommand{\blems}{\begin{lems}}
\newcommand{\elems}{\end{lems}}
\newcommand{\bcors}{\begin{cors}}
\newcommand{\ecors}{\end{cors}}
\newcommand{\bexs}{\begin{exas}}
\newcommand{\eexs}{\end{exas}}
\newcommand{\balgs}{\begin{algs}}
\newcommand{\ealgs}{\end{algs}}
\newcommand{\bass}{\begin{asss}}
\newcommand{\eass}{\end{asss}}

\newenvironment{carlist}
 {\begin{list}{$\bullet$}
 {\setlength{\topsep}{0in} \setlength{\partopsep}{0in}
  \setlength{\parsep}{0in} \setlength{\itemsep}{\parskip}
  \setlength{\leftmargin}{0.07in} \setlength{\rightmargin}{0.08in}
  \setlength{\listparindent}{0in} \setlength{\labelwidth}{0.08in}
  \setlength{\labelsep}{0.1in} \setlength{\itemindent}{0in}}}
 {\end{list}}

\newcommand{\bcar}{\begin{carlist}}
\newcommand{\ecar}{\end{carlist}}

\newenvironment{carliste}
 {\begin{list}x
 {\setlength{\topsep}{0in} \setlength{\partopsep}{0in}
  \setlength{\parsep}{0in} \setlength{\itemsep}{\parskip}
  \setlength{\leftmargin}{0.07in} \setlength{\rightmargin}{0.08in}
  \setlength{\listparindent}{0in} \setlength{\labelwidth}{0.08in}
  \setlength{\labelsep}{0.1in} \setlength{\itemindent}{0in}}}
 {\end{list}}

\newcommand{\bcare}{\begin{carliste}}
\newcommand{\ecare}{\end{carliste}}

\newcommand{\tracer}[2]{\ensuremath{\langle \!\langle {#1}, \; {#2}
\rangle \!\rangle}}

\makeatletter
\long\def\@makecaption#1#2{
        \vskip 0.8ex
        \setbox\@tempboxa\hbox{\small {\bf #1:} #2}
        \parindent 1.5em  
        \dimen0=\hsize
        \advance\dimen0 by -3em
        \ifdim \wd\@tempboxa >\dimen0
                \hbox to \hsize{
                        \parindent 0em
                        \hfil 
                        \parbox{\dimen0}{\def\baselinestretch{0.96}\small
                                {\bf #1.} #2
                                } 
                        \hfil}
        \else \hbox to \hsize{\hfil \box\@tempboxa \hfil}
        \fi
        }
\makeatother

\long\def\comment#1{}

\makeatletter
\def\@cite#1#2{[\if@tempswa #2 \fi #1]}
\makeatother

\makeatletter
\long\def\barenote#1{
    \insert\footins{\footnotesize
    \interlinepenalty\interfootnotelinepenalty 
    \splittopskip\footnotesep
    \splitmaxdepth \dp\strutbox \floatingpenalty \@MM
    \hsize\columnwidth \@parboxrestore
    {\rule{\z@}{\footnotesep}\ignorespaces
      #1\strut}}}
\makeatother

\newcommand{\bit}{\begin{itemize}}
\newcommand{\eit}{\end{itemize}}
\newcommand{\ben}{\begin{enumerate}}
\newcommand{\een}{\end{enumerate}}

\newcommand{\bear}{\begin{eqnarray}}
\newcommand{\eear}{\end{eqnarray}}

\newcommand{\df}{\ensuremath{d}}

\newcommand{\order}{{\mathcal{O}}}

\newcommand{\graph}{\ensuremath{G}}

\newcommand{\vertex}{\ensuremath{V}}
\newcommand{\edge}{\ensuremath{E}}

\newcommand{\invn}[1]{\ensuremath{{#1}^{-1}}}

\newcommand{\Exs}{\ensuremath{{\mathbb{E}}}}

\newcommand{\estart}{\begin{equation}}
\newcommand{\eend}{\end{equation}}

\newcommand{\widgraph}[2]{\includegraphics[keepaspectratio,width=#1]{#2}}

\newcommand{\defn}{\ensuremath{:  =}}

\newcommand{\bec}{\begin{center}}
\newcommand{\enc}{\end{center}}

\newcommand{\beit}{\begin{itemize}}
\newcommand{\enit}{\end{itemize}}

\newcommand{\been}{\begin{enumerate}}
\newcommand{\enen}{\end{enumerate}}

\newcommand{\comsl}{\begin{slide}}
\newcommand{\comspor}{\begin{slide*}}

\newcommand{\comsld}[2]{\begin{slide}[#1,#2]}
\newcommand{\comspord}[2]{\begin{slide*}[#1,#2]}

\newcommand{\mendsl}{\end{slide}}
\newcommand{\mendspo}{\end{slide*}}

\newcommand{\opt}[1]{\ensuremath{{#1}^{*}}}
\newcommand{\estim}[1]{\ensuremath{\widehat{#1}}}

\newcommand{\wtil}[1]{\ensuremath{\widetilde{#1}}}

\newcommand{\inv}[1]{\ensuremath{\big(#1\big)^{-1}}}

\newcommand{\real}{\ensuremath{{\mathbb{R}}}}

\DeclareMathOperator{\sign}{sign}

\def\sign{\textrm{sign}}

\newcommand{\matnorm}[2]{| \! | \! | #1 | \! | \! |_{{#2}}}
\newcommand{\vecnorm}[2]{\| #1 \|_{{#2}}}
\renewcommand{\sp}[2]{{#1}^{(#2)}}

\newcommand{\Eset}{\ensuremath{E}}

\def\conc{\Theta}

\def\invl{\frac{1}{\lambda_n}}

\newcommand\ThetaWitness{\widetilde{\Theta}}

\def\ConstStar{c_{*}}

\def\modelCompFac{K}

\endlocaldefs
\else

\fi

\newcommand{\KconCov}{\ensuremath{K_{\CovMat^*}}}

\newcommand{\KconHess}{\ensuremath{K_{\BigHess}}}

\newcommand{\E}{\ensuremath{\mathbb{E}}}

\newcommand{\BigHess}{\ensuremath{\Gamma^*}}

\newcommand{\csubg}{\ensuremath{\sigma}}

\newcommand{\myuvec}{\ensuremath{u}}
\newcommand{\Umat}{\ensuremath{U}}
\newcommand{\Vmat}{\ensuremath{V}}
\newcommand{\SamCov}{\ensuremath{\estim{\CovMat}}}

\newcommand{\mprob}{\ensuremath{\mathbb{P}}}

\newcommand{\mutinco}{\ensuremath{\alpha}} 
\newcommand{\Breg}[2]{\ensuremath{D_g}(#1 \| #2)}
\newcommand{\Symconepl}[1]{\ensuremath{\mathcal{S}^{#1}_+}}

\newcommand{\ThetaWit}{\ensuremath{\ThetaWitness}}
\newcommand{\ZWit}{\ensuremath{\widetilde{Z}}}

\newcommand{\thetamin}{\ensuremath{\theta_{\operatorname{min}}}}

\newcommand{\Res}{\ensuremath{R}}
\newcommand{\Wnoise}{\ensuremath{W}}

\newcommand{\myvec}[1]{\ensuremath{\widebar{#1}}}

\newcommand{\rad}{\ensuremath{r}}
\newcommand{\Ball}{\ensuremath{\mathbb{B}}}
\newcommand{\smallvec}{\ensuremath{\operatorname{vec}}}
\newcommand{\Term}{\ensuremath{T}}

\newcommand{\WitEvent}{\mathcal{A}}

\newcommand{\CovMatStar}{\ensuremath{\CovMat^*}}

\newcommand{\CovHat}{\ensuremath{\widehat{\CovMat}}}

\newcommand{\Model}{\ensuremath{\mathcal{M}}}

\newcommand{\obsind}{\ensuremath{k}}

\newcommand{\Xsam}[1]{\ensuremath{X^{(#1)}}}
\newcommand{\Xsambar}[1]{\ensuremath{\widebar{X}^{(#1)}}}

\newcommand{\newuvar}[1]{\ensuremath{U^{(#1)}}}
\newcommand{\newvvar}[1]{\ensuremath{V^{(#1)}}}

\newcommand{\Aij}{\ensuremath{\mathbb{A}_{ij}}}

\newcommand{\csubone}{\ensuremath{\gamma}}
\newcommand{\csubtwo}{\ensuremath{\phi}}
\newcommand{\SamDev}[1]{\ensuremath{\sampdev^{(#1)}_{ij}}}

\newcommand{\Cawful}{\ensuremath{C_1}}
\newcommand{\Cawfultwo}{\ensuremath{C_2}}
\newcommand{\Cawthree}{\ensuremath{C_3}}
\newcommand{\Cawfour}{\ensuremath{C_4}}

\begin{document}

\ifaos
\begin{frontmatter}
\title{High-dimensional covariance estimation by minimizing
$\ell_1$-penalized log-determinant divergence} \runtitle{Rates for
$\ell_1$-penalized log-determinant divergences}

\begin{aug}
\author{\fnms{Pradeep}
\snm{Ravikumar}\ead[label=e1]{pradeepr@stat.berkeley.edu}},
\author{\fnms{Martin J.}
\snm{Wainwright}\ead[label=e3]{wainwrig@stat.berkeley.edu}},\\
\author{\fnms{Garvesh}
\snm{Raskutti}\ead[label=e2]{garveshr@stat.berkeley.edu}} \and
\author{\fnms{Bin} \snm{Yu} \ead[label=e4]{binyu@stat.berkeley.edu}}

\runauthor{Ravikumar and Wainwright and Raskutti and Yu}

\affiliation{University of California, Berkeley} \address{Berkeley, CA
94720-1776 USA\\ \printead{e1}\\ \phantom{E-mail:\ }\printead*{e3}\\
\phantom{E-mail:\ }\printead*{e2}\\ \phantom{E-mail:\ }\printead*{e4}}
\end{aug}
\else
\begin{center}

{\bf{\LARGE{High-dimensional covariance estimation by minimizing
	$\ell_1$-penalized log-determinant divergence}}}

\vspace*{.2in}

\begin{tabular}{cc}
Pradeep Ravikumar$^{\dagger}$ & Martin J. Wainwright$^{\dagger,\sharp}$ \\
\texttt{pradeepr@stat.berkeley.edu} &
\texttt{wainwrig@stat.berkeley.edu} 
\end{tabular}
\vspace*{.06in}

\begin{tabular}{cc}

Garvesh Raskutti$^{\dagger}$ & Bin Yu$^{\dagger,\sharp}$ \\
\texttt{garveshr@stat.berkeley.edu} &
\texttt{binyu@stat.berkeley.edu}
\end{tabular}

\vspace{.5cm}

\begin{tabular}{c}
Department of Statistics$^\dagger$, and \\
Department of Electrical Engineering and Computer Sciences$^\sharp$ \\
University of California, Berkeley \\
Berkeley, CA  94720
\end{tabular}

\vspace{.2cm}

\today

\vspace{.2cm}

\end{center}
\fi

\begin{abstract}%

Given i.i.d. observations of a random vector $X \in \mathbb{R}^p$, we
study the problem of estimating both its covariance matrix $\Sigma^*$,
and its inverse covariance or concentration matrix \mbox{$\Theta^* =
(\Sigma^*)^{-1}$.}  We estimate $\Theta^*$ by minimizing an
$\ell_1$-penalized log-determinant Bregman divergence; in the
multivariate Gaussian case, this approach corresponds to
$\ell_1$-penalized maximum likelihood, and the structure of $\Theta^*$
is specified by the graph of an associated Gaussian Markov random
field.  We analyze the performance of this estimator under
high-dimensional scaling, in which the number of nodes in the graph
$p$, the number of edges $s$ and the maximum node degree $d$, are
allowed to grow as a function of the sample size $n$.  In addition to
the parameters $(p,s,d)$, our analysis identifies other key quantities
that control rates: (a) the $\ell_\infty$-operator norm of the true
covariance matrix $\Sigma^*$; and (b) the $\ell_\infty$ operator norm
of the sub-matrix $\Gamma^*_{S S}$, where $S$ indexes the graph edges,
and $\Gamma^* = (\Theta^*)^{-1} \otimes (\Theta^*)^{-1}$; and (c) a
mutual incoherence or irrepresentability measure on the matrix
$\Gamma^*$ and (d) the rate of decay $1/\sctail(\numobs,\scdelta)$ on
the probabilities $ \{|\widehat{\Sigma}^n_{ij}- \Sigma^*_{ij}| >
\delta \}$, where $\widehat{\Sigma}^n$ is the sample covariance based
on $n$ samples.  Our first result establishes consistency of our
estimate $\widehat{\Theta}$ in the elementwise maximum-norm. This in
turn allows us to derive convergence rates in Frobenius and spectral
norms, with improvements upon existing results for graphs with maximum
node degrees $\degmax = o(\sqrt{\spindex})$.  In our second result, we
show that with probability converging to one, the estimate
$\widehat{\Theta}$ correctly specifies the zero pattern of the
concentration matrix $\Theta^*$.  We illustrate our theoretical
results via simulations for various graphs and problem parameters,
showing good correspondences between the theoretical predictions and
behavior in simulations.
\end{abstract}

\ifaos


\end{frontmatter}
\fi

\section{Introduction}

The area of high-dimensional statistics deals with estimation in the
``large \pdim, small \numobs'' setting, where $\pdim$ and $\numobs$
correspond, respectively, to the dimensionality of the data and the
sample size. Such high-dimensional problems arise in a variety of
applications, among them remote sensing, computational biology and
natural language processing, where the model dimension may be
comparable or substantially larger than the sample size.  It is
well-known that such high-dimensional scaling can lead to dramatic
breakdowns in many classical procedures.  In the absence of additional
model assumptions, it is frequently impossible to obtain consistent
procedures when $\pdim \gg \numobs$. Accordingly, an active line of
statistical research is based on imposing various restrictions on the
model----for instance, sparsity, manifold structure, or graphical
model structure----and then studying the scaling behavior of different
estimators as a function of sample size $\numobs$, ambient dimension
$\pdim$ and additional parameters related to these structural
assumptions.

In this paper, we study the following problem: given $\numobs$ i.i.d.
observations $\{\SampVar\}_{\obsind=1}^{\numobs}$ of a zero mean
random vector $X \in \real^{\pdim}$, estimate both its covariance
matrix \mbox{$\CovMatStar$,} and its inverse covariance or
concentration matrix $\ThetaStar \defn \inv{\CovMatStar}$.  Perhaps
the most natural candidate for estimating $\CovMatStar$ is the
empirical sample covariance matrix, but this is known to behave poorly
in high-dimensional settings. For instance, when $\pdim/\numobs
\rightarrow c > 0$, and the samples are drawn i.i.d. from a
multivariate Gaussian distribution, neither the eigenvalues nor the
eigenvectors of the sample covariance matrix are consistent estimators
of the population
versions~\cite{Johnstone2001,JohnstoneLu2004}. Accordingly, many
regularized estimators have been proposed to estimate the covariance
or concentration matrix under various model assumptions.  One natural
model assumption is that reflected in shrinkage estimators, such as in
the work of \citet{LedoitWolf2003}, who proposed to shrink the sample
covariance to the identity matrix. An alternative model assumption,
relevant in particular for time series data, is that the covariance or
concentration matrix is banded, meaning that the entries decay based
on their distance from the diagonal. \citet{Furrer2007} proposed to
shrink the covariance entries based on this distance from the
diagonal. \citet{Wu2003} and \citet{Huang2006} estimate these banded
concentration matrices by using thresholding and $\ell_1$-penalties
respectively, as applied to a Cholesky factor of the inverse
covariance matrix. \citet{BickelLevina2008} prove the consistency of
these banded estimators so long as $\frac{(\log\,\pdim)^2}{\numobs}
\rightarrow 0$ and the model covariance matrix is banded as well, but
as they note, these estimators depend on the presented order of the
variables.

A related class of models are based on positing some kind of sparsity,
either in the covariance matrix, or in the inverse covariance.
\citet{BickelLevina2007} study thresholding estimators of covariance
matrices, assuming that each row satisfies an $\ell_q$-ball sparsity
assumption.  In independent work, \citet{Karoui2007} also studied
thresholding estimators of the covariance, but based on an alternative
notion of sparsity, one which captures the number of closed paths of
any length in the associated graph.  Other work has studied models in
which the inverse covariance or concentration matrix has a sparse
structure.  As will be clarified in the next section, when the random
vector is multivariate Gaussian, the set of non-zero entries in the
concentration matrix correspond to the set of edges in an associated
Gaussian Markov random field (GMRF).  In this setting, imposing
sparsity on the concentration matrix can be interpreted as requiring
that the graph underlying the GMRF have relatively few edges.  A line
of recent papers~\citep{AspreBanG2008,FriedHasTib2007,YuanLin2007}
have proposed an estimator that minimizes the Gaussian negative
log-likelihood regularized by the $\ell_1$ norm of the entries (or the
off-diagonal entries) of the concentration matrix. The resulting
optimization problem is a log-determinant program, which can be solved
in polynomial time with interior point methods~\citep{Boyd02}, or by
faster co-ordinate descent
algorithms~\citep{AspreBanG2008,FriedHasTib2007}.  In recent work,
\citet{Rothman2007} have analyzed some aspects of high-dimensional
behavior of this estimator; assuming that the minimum and maximum
eigenvalues of $\CovMatStar$ are bounded, they show that consistent
estimates can be achieved in Frobenius and operator norm, in
particular at the rate $\order(\sqrt{\frac{(\spdim + \pdim) \log
\pdim}{\numobs}})$.

The focus of this paper is the problem of estimating the concentration
matrix $\Theta^*$ under sparsity conditions.  We do not impose
specific distributional assumptions on $X$ itself, but rather analyze the estimator in terms of 
the tail behavior of the maximum deviation $\max_{i,j}
|\estim{\Sigma}^n_{ij} - \CovMatStar_{ij}|$ of the sample and
population covariance matrices.  To estimate $\Theta^*$, we
consider minimization of an $\ell_1$-penalized log-determinant Bregman
divergence, which is equivalent to the usual $\ell_1$-penalized
maximum likelihood when $X$ is multivariate Gaussian.  We analyze the
behavior of this estimator under high-dimensional scaling, in which
the number of nodes $\pdim$ in the graph, and the maximum node degree
$\degmax$ are all allowed to grow as a function of the sample size
$\numobs$.

In addition to the triple $(\numobs, \pdim, \degmax)$, we also
explicitly keep track of certain other measures of model complexity,
that could potentially scale as well.  The first of these measures is
the $\ell_\infty$-operator norm of the covariance matrix
$\CovMatStar$, which we denote by $\KconCov \defn
\matnorm{\CovMatStar}{\infty}$.  The next quantity involves the
Hessian of the log-determinant objective function, $\BigHess \defn
(\ThetaStar)^{-1} \otimes (\ThetaStar)^{-1}$.  When the distribution
of $X$ is multivariate Gaussian, this Hessian has the more explicit
representation $\BigHess_{(j,k), (\ell, m)} = \operatorname{cov}\{X_j
X_k, \; X_\ell X_m \}$, showing that it measures the covariances of
the random variables associated with each edge of the graph.  For this
reason, the matrix $\BigHess$ can be viewed as an edge-based
counterpart to the usual node-based covariance matrix $\CovMatStar$.
Using $\EsetPlus$ to index the variable pairs $(i,j)$ associated with
non-zero entries in the inverse covariance.  our analysis involves the
quantity $\KconHess = \matnorm{(\BigHess_{\EsetPlus
\EsetPlus})^{-1}}{\infty}$. Finally, we also impose a mutual
incoherence or irrepresentability condition on the Hessian $\BigHess$;
this condition is similar to assumptions imposed on $\CovMatStar$ in
previous work~\cite{Tropp2006,Zhao06,MeinsBuhl2006,Wainwright2006_new} on
the Lasso.  We provide some examples where the Lasso
irrepresentability condition holds, but our corresponding condition on
$\BigHess$ fails; however, we do not know currently whether one
condition strictly dominates the other.

Our first result establishes consistency of our estimator
$\estim{\Theta}$ in the elementwise maximum-norm, providing a rate
that depends on the tail behavior of the entries in the random matrix
$\SamCov^\numobs - \CovMatStar$. For the special case of sub-Gaussian
random vectors with concentration matrices having at most $d$
non-zeros per row, a corollary of our analysis is consistency in
spectral norm at rate \mbox{$\matnorm{\estim{\Theta} - \Theta^*}{2} =
\order(\sqrt{(\degmax^2 \,\log \pdim)/\numobs})$,} with high
probability, thereby strengthening previous
results~\cite{Rothman2007}.  Under the milder restriction of each
element of $X$ having bounded $4\momentpow$-th moment, the rate in
spectral norm is substantially slower---namely,
\mbox{$\matnorm{\estim{\Theta} - \ThetaStar}{2} = \order(\degmax\,
\pdim^{1/2m}/\sqrt{\numobs})$}---highlighting that the familiar
logarithmic dependence on the model size $\pdim$ is linked to
particular tail behavior of the distribution of $X$.  Finally, we show
that under the same scalings as above, with probability converging to
one, the estimate $\estim{\Theta}$ correctly specifies the zero
pattern of the concentration matrix $\Theta^*$.

The remainder of this paper is organized as follows.  In
Section~\ref{SecBackground}, we set up the problem and give some
background.  Section~\ref{SecResult} is devoted to statements of our
main results, as well as discussion of their consequences.
Section~\ref{SecProof} provides an outline of the proofs, with the
more technical details deferred to appendices.  In
Section~\ref{SecExperiments}, we report the results of some simulation
studies that illustrate our theoretical predictions.

\myparagraph{Notation} For the convenience of the reader, we summarize
here notation to be used throughout the paper. Given a vector $\myuvec
\in \real^\df$ and parameter $a \in [1, \infty]$, we use
$\|\myuvec\|_a$ to denote the usual $\ell_a$ norm.  Given a matrix
$\Umat \in \real^{p \times p}$ and parameters $a,b \in [1, \infty]$,
we use $\matnorm{\Umat}{a,b}$ to denote the induced matrix-operator
norm $\max_{\|y\|_a = 1} \|\Umat y\|_b$; see \citet{Horn1985} for
background.  Three cases of particular importance in this paper are
the \emph{spectral norm} $\matnorm{\Umat}{2}$, corresponding to the
maximal singular value of $\Umat$; the
\emph{$\ell_\infty/\ell_\infty$-operator norm}, given by
\begin{eqnarray}
\label{EqnLinfOp}
\matnorm{\Umat}{\infty} & \defn & \max \limits_{j=1, \ldots, p}
\sum_{k=1}^p |\Umat_{jk}|,
\end{eqnarray}
and the \emph{$\ell_1/\ell_1$-operator norm}, given by
$\matnorm{\Umat}{1} = \matnorm{\Umat^T}{\infty}$.  Finally, we use
$\|\Umat\|_\infty$ to denote the element-wise maximum $\max_{i,j}
|\Umat_{ij}|$; note that this is not a matrix norm, but rather a norm
on the vectorized form of the matrix. For any matrix $\Umat \in
\real^{\pdim \times \pdim}$, we use $\smallvec(\Umat)$ or equivalently
$\widebar{\Umat} \in \real^{\pdim^2}$ to denote its \emph{vectorized
form}, obtained by stacking up the rows of $\Umat$.  We use
$\tracer{\Umat}{\Vmat} \defn \sum_{i,j} \Umat_{ij} \Vmat_{ij}$ to
denote the \emph{trace inner product} on the space of symmetric
matrices. Note that this inner product induces the \emph{Frobenius
norm} $\matnorm{\Umat}{F} \defn \sqrt{\sum_{i,j}
\Umat_{ij}^2}$. Finally, for asymptotics, we use the following
standard notation: we write $f(n) = \order(g(n))$ if $f(n) \leq c
g(n)$ for some constant $c < \infty$, and $f(n) = \Omega(g(n))$ if
$f(n) \geq c' g(n)$ for some constant $c' > 0$.  The notation
\mbox{$f(n) \asymp g(n)$} means that \mbox{$f(n) = \order(g(n))$} and
\mbox{$f(n) = \Omega(g(n))$.}

\section{Background and problem set-up}
\label{SecBackground}

Let $X = (X_1, \ldots, X_\pdim)$ be a zero mean $\pdim$-dimensional
random vector. The focus of this paper is the problem of estimating
the covariance matrix $\CovMatStar \defn \Exs[ X X^T]$ and
concentration matrix $\ThetaStar \defn \invn{\CovMatStar}$ of the
random vector $X$ given $\numobs$ i.i.d. observations
$\{X^{(\obsind)}\}_{\obsind=1}^{\numobs}$.  In this section, we
provide background, and set up this problem more precisely.  We begin
with background on Gaussian graphical models, which provide one
motivation for the estimation of concentration matrices.  We then
describe an estimator based based on minimizing an $\ell_1$
regularized log-determinant divergence; when the data are drawn from a
Gaussian graphical model, this estimator corresponds to
$\ell_1$-regularized maximum likelihood.  We then discuss the
distributional assumptions that we make in this paper.

\subsection{Gaussian graphical models}

One motivation for this paper is the problem of Gaussian graphical
model selection.  A graphical model or a Markov random field is a
family of probability distributions for which the conditional
independence and factorization properties are captured by a graph. Let
$X = (X_1, X_2, \ldots, X_\pdim)$ denote a zero-mean Gaussian random
vector; its density can be parameterized by the inverse covariance or
\emph{concentration matrix} $\ThetaStar = (\CovMatStar)^{-1} \in
\Symconepl{\pdim}$, and can be written as
\begin{eqnarray}
\label{EqnDefnGaussMRF}
f(x_1, \ldots, x_\pdim; \ThetaStar) & = & \frac{1}{\sqrt{(2 \pi)^\pdim
 \det((\ThetaStar)^{-1})}} \; \exp \big\{ -\frac{1}{2} x^T \ThetaStar
 x \big \}.
\end{eqnarray}
\begin{figure}[ht]
\begin{center}
\begin{tabular}{ccc}
\psfrag{#1#}{$1$} \psfrag{#2#}{$2$} \psfrag{#3#}{$3$}
\psfrag{#4#}{$4$} \psfrag{#5#}{$5$}
\raisebox{.2in}{\widgraph{0.3\textwidth}{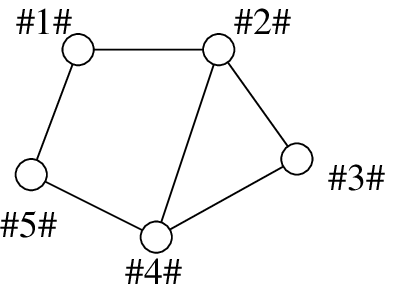}} & &
\widgraph{.25\textwidth}{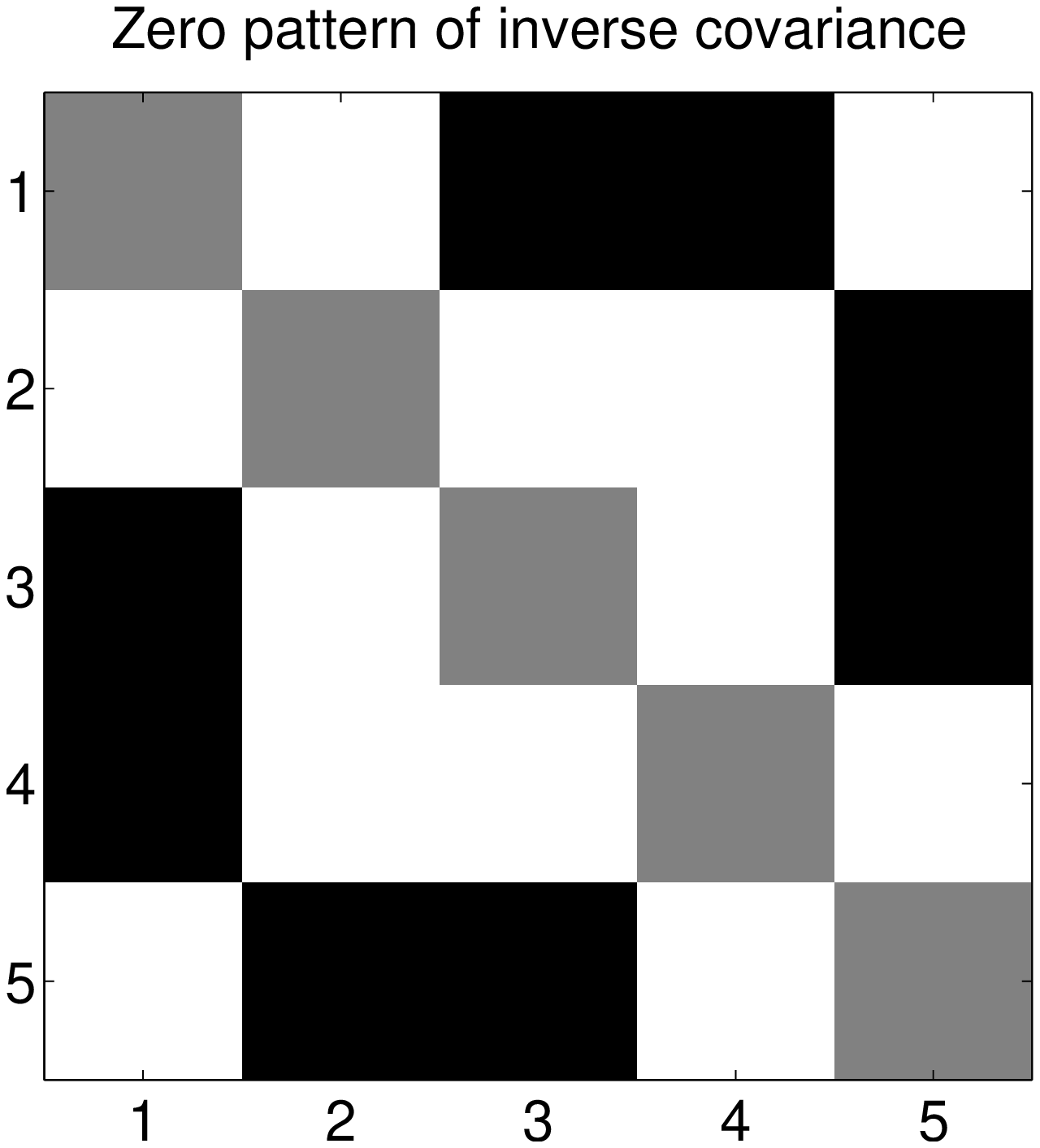} \\
(a) & & (b)
\end{tabular}
\end{center}
\caption{(a) Simple undirected graph.  A Gauss Markov random field has
a Gaussian variable $X_i$ associated with each vertex $i \in \vertex$.
This graph has $\pdim = 5$ vertices, maximum degree $d = 3$ and $s=6$
edges.  (b) Zero pattern of the inverse covariance $\ThetaStar$
associated with the GMRF in (a).  The set $\EdgeSet(\ThetaStar)$
corresponds to the off-diagonal non-zeros (white blocks); the diagonal
is also non-zero (grey squares), but these entries do not correspond
to edges.  The black squares correspond to non-edges, or zeros in
$\ThetaStar$.}
\label{FigMarkov}
\end{figure}

We can relate this Gaussian distribution of the random vector $X$ to a
graphical model as follows.  Suppose we are given an undirected graph
$\graph = (\vertex, \edge)$ with vertex set $\vertex = \{1, 2, \ldots,
\pdim \}$ and edge\footnote{As a remark on notation, we would like to
contrast the notation for the edge-set $\edge$ from the notation for
an expectation of a random variable, $\mathbb{E}(\cdot)$.} set
$\edge$, so that each variable $X_i$ is associated with a
corresponding vertex $i \in \vertex$. The Gaussian Markov random field
(GMRF) associated with the graph $\graph$ over the random vector $X$
is then the family of Gaussian distributions with concentration
matrices $\ThetaStar$ that respect the edge structure of the graph, in
the sense that $\ThetaStar_{ij} = 0$ if $(i,j) \notin
\edge$. Figure~\ref{FigMarkov} illustrates this correspondence between
the graph structure (panel (a)), and the sparsity pattern of the
concentration matrix $\ThetaStar$ (panel (b)).  The problem of
estimating the entries of the concentration matrix $\ThetaStar$
corresponds to estimating the Gaussian graphical model instance, while
the problem of estimating the off-diagonal zero-pattern of the
concentration matrix----that is, the set
\begin{eqnarray}
\label{EqnDefnEdgeSet}
\edge(\ThetaStar) & \defn & \{i, j \in \vertex \mid \, i \neq j,
\ThetaStar_{ij} \neq 0 \}
\end{eqnarray}
corresponds to the problem of Gaussian graphical \emph{model
selection}. 

With a slight abuse of notation, we define the \emph{sparsity index}
$\spindex \defn |\edge(\ThetaStar)|$ as the total number of non-zero
elements in off-diagonal positions of $\ThetaStar$; equivalently, this
corresponds to twice the number of edges in the case of a Gaussian
graphical model. We also define the \emph{maximum degree or row
cardinality}
\begin{eqnarray}
\label{EqnDefnDegmax}
\degmax & \defn & \max_{i = 1, \ldots, \pdim } \biggr|\big \{ j \in
\vertex \, \mid \, \ThetaStar_{ij} \neq 0 \big\} \biggr|,
\end{eqnarray}
corresponding to the maximum number of non-zeros in any row of
$\ThetaStar$; this corresponds to the maximum degree in the graph of
the underlying Gaussian graphical model.  Note that we have included
the diagonal entry $\ThetaStar_{ii}$ in the degree count,
corresponding to a self-loop at each vertex.

It is convenient throughout the paper to use graphical terminology,
such as degrees and edges, even though the distributional assumptions
that we impose, as described in Section~\ref{SecDistAssum}, are milder
and hence apply even to distributions that are not Gaussian MRFs.

\subsection{$\ell_1$-penalized log-determinant divergence}

An important set in this paper is the cone
\begin{eqnarray}
\Symconepl{\pdim} & \defn & \big \{ A \in \real^{\pdim \times \pdim}
\mid A = A^T, \; A \succeq 0 \big \},
\end{eqnarray}
formed by all symmetric positive semi-definite matrices in $\pdim$
dimensions. We assume that the covariance matrix $\CovMatStar$ and
concentration matrix $\ThetaStar$ of the random vector $X$ are
strictly positive definite, and so lie in the interior of this cone
$\Symconepl{\pdim}$. 

The focus of this paper is a particular type of $M$-estimator for the
concentration matrix $\ThetaStar$, based on minimizing a Bregman
divergence between symmetric matrices.  A function is of Bregman type
if it is strictly convex, continuously differentiable and has bounded
level sets~\cite{Bregman67a,Censor}.  Any such function induces a
\emph{Bregman divergence} of the form $\Breg{A}{B} = g(A) - g(B) -
\trs{\nabla g(B)}{A-B}$.  From the strict convexity of $g$, it follows
that $\Breg{A}{B} \geq 0$ for all $A$ and $B$, with equality if and
only if $A = B$.

As a candidate Bregman function, consider the log-determinant barrier
function, defined for any matrix $A \in \Symconepl{\pdim}$ by
\begin{eqnarray}
\label{EqnDefnLogDet}
g(A) & \defn &  \begin{cases} - \log \det(A)  & \mbox{if $A \succ 0$} \\
                              + \infty & \mbox{otherwise.}
                \end{cases}
\end{eqnarray}
As is standard in convex analysis, we view this function as taking
values in the extended reals $\real_* = \real \cup \{+\infty \}$.
With this definition, the function $g$ is strictly convex, and its
domain is the set of strictly positive definite matrices.  Moreover,
it is continuously differentiable over its domain, with $\nabla g(A) =
- A^{-1}$; see Boyd and Vandenberghe~\cite{Boyd02} for further
discussion.  The Bregman divergence corresponding to this
log-determinant Bregman function $g$ is given by
\begin{eqnarray}
\label{EqnDefnBreg}
\Breg{A}{B} & \defn & - \log \det A + \log \det B +
\tracer{B^{-1}}{A-B},
\end{eqnarray}
valid for any $A, B \in \Symconepl{\pdim}$ that are strictly positive
definite.  This divergence suggests a natural way to estimate
concentration matrices---namely, by minimizing the divergence
$\Breg{\ThetaStar}{\Theta}$---or equivalently, by minimizing the
function
\begin{equation}
\label{EqnPop}
\min_{\Theta \succ 0 } \big \{ \tracer{\Theta}{\CovMatStar} - \log
\det \Theta \big \},
\end{equation}
where we have discarded terms independent of $\Theta$, and used the
fact that the inverse of the concentration matrix is the covariance matrix
(i.e., $(\ThetaStar)^{-1} = \CovMatStar= \Exs[X X^T]$).  Of course,
the convex program~\eqref{EqnPop} cannot be solved without knowledge
of the true covariance matrix $\CovMatStar$, but one can take the
standard approach of replacing $\CovMatStar$ with an empirical
version, with the possible addition of a regularization term.

In this paper, we analyze a particular instantiation of this strategy.
Given $\numobs$ samples, we define the \emph{sample covariance matrix}
\begin{eqnarray}
\label{EqnDefnSamCov}
\SamCov^\numobs & \defn & \frac{1}{\numobs} \sum_{\obsind=1}^\numobs
\sp{X}{\obsind} (\sp{X}{\obsind})^T.
\end{eqnarray}
To lighten notation, we occasionally drop the superscript $\numobs$,
and simply write $\SigHat$ for the sample covariance.  We also define
the \emph{off-diagonal $\ell_1$ regularizer}
\begin{eqnarray}
\ellreg{\Theta} & \defn & \sum_{i \neq j} |\Theta_{ij}|,
\end{eqnarray}
where the sum ranges over all $i, j = 1, \ldots, \pdim$ with $i \neq
j$.  Given some regularization constant $\regpar > 0$, we consider
estimating $\ThetaStar$ by solving the following
\emph{$\ell_1$-regularized log-determinant program}:
\begin{eqnarray}
\label{EqnGaussMLE}
\ThetaHat & \defn & \arg\min_{\Theta \succ 0} \big \{
\tracer{\Theta}{\SigHat^\numobs} - \log \det(\Theta) + \regpar
\ellreg{\Theta} \big \}.
\end{eqnarray}
As shown in Appendix~\ref{AppLemMLECharac}, for any $\regpar > 0$ and
sample covariance matrix $\SigHat^\numobs$ with strictly positive
diagonal, this convex optimization problem has a unique optimum, so
there is no ambiguity in equation~\eqref{EqnGaussMLE}.  When the data
is actually drawn from a multivariate Gaussian distribution, then the
problem~\eqref{EqnGaussMLE} is simply $\ell_1$-regularized maximum
likelihood.

\def\Conste{c}
\def\Constp{c}

\subsection{Tail conditions}
\label{SecDistAssum}

In this section, we describe the tail conditions that underlie our
analysis.  Since the estimator ~\eqref{EqnGaussMLE} is based on using
the sample covariance $\SamCov^\numobs$ as a surrogate for the
(unknown) covariance $\CovMatStar$, any type of consistency requires
bounds on the difference $\SamCov^\numobs - \CovMatStar$.  In
particular, we define the following tail condition:
\bdes[Tail conditions]
\label{DefnTail}
The random vector $X$ satisfies tail condition $\Tail(\sctail,
\scthresh)$ if there exists a constant $\scthresh \in (0, \infty]$ and
a function $\sctail: \mathbb{N} \times (0,\infty) \rightarrow (0,
\infty)$ such that for any $(i,j) \in \vertex \times \vertex$:
\begin{eqnarray}
\label{EqnSamTail}
\mprob[|\CovHat^\numobs_{ij} - \CovMatStar_{ij}| \geq \scdelta] & \leq
& 1/\sctail(\numobs,\scdelta) \qquad \mbox{for all $\scdelta \in (0,
1/\scthresh]$.}
\end{eqnarray}
We adopt the convention $1/0 \defn + \infty$, so that the value
$\scthresh = 0$ indicates the inequality holds for any $\scdelta \in
(0,\infty)$.
\edes

\newcommand{\garvesh}{\ensuremath{a}}

Two important examples of the tail function $\sctail$ are the
following:
\begin{enumerate} 
\item[(a)] an \emph{exponential-type tail function}, meaning that
$\sctail(\numobs,\scdelta) = \exp(\Conste \, \numobs\,
\scdelta^{\garvesh})$, for some scalar $\Conste > 0$, and exponent
$\garvesh > 0$; and
\item[(b)] a \emph{polynomial-type tail function}, meaning that
$\sctail(\numobs,\scdelta) = \Constp \, \numobs^{\momentpow} \,
\scdelta^{2\momentpow}$, for some positive integer $\momentpow \in
\mathbb{N}$ and scalar $\Constp > 0$.
\end{enumerate}
As might be expected, if $X$ is multivariate Gaussian, then the
deviations of sample covariance matrix have an exponential-type tail
function with $\garvesh = 2$.  A bit more generally, in the following
subsections, we provide broader classes of distributions whose sample
covariance entries satisfy exponential and a polynomial tail bounds
(see Lemmata~\ref{LEM_SAM_COV_BOUND_SUBG}
and~\ref{LEM_SAM_COV_BOUND_MOMENT} respectively).

Given a larger number of samples $\numobs$, we expect the tail
probability bound $1/\sctail(\numobs,\scdelta)$ to be smaller, or
equivalently, for the tail function $\sctail(\numobs,\scdelta)$ to
larger.  Accordingly, we require that $\sctail$ is monotonically
increasing in $\numobs$, so that for each fixed $\scdelta >0$, we can
define the inverse function
\begin{eqnarray}
\label{EqnSamTailN}
\sctailInvN(r; \scdelta) & \defn & \arg \max \big \{ n \; \mid \;
\sctail(\numobs, \scdelta) \leq r \big \}.
\end{eqnarray}
Similarly, we expect that $\sctail$ is monotonically increasing in
$\scdelta$, so that for each fixed $\numobs$, we can define the
inverse in the second argument
\begin{eqnarray}
\label{EqnSamTailT}
\sctailInvT(r; \numobs) & \defn & \arg \max \big \{ \scdelta \; \mid
\; \sctail(\numobs, \scdelta) \leq r \big \}.
\end{eqnarray}
For future reference, we note a simple consequence of the monotonicity
of the tail function $\sctail$---namely
\begin{eqnarray}
\label{EqnMonot}
\numobs > \sctailInvN( \delta, r) \quad \mbox{for some $\delta > 0$} &
\Longrightarrow & \sctailInvT(\numobs,r) \leq \delta.
\end{eqnarray}
The inverse functions $\sctailInvN$ and $\sctailInvT$ play an
important role in describing the behavior of our estimator.  We
provide concrete examples in the following two subsections.

\subsubsection{Sub-Gaussian distributions}
\label{SecSamCovBoundSubG}
In this subsection, we study the case of i.i.d. observations of
sub-Gaussian random variables.
\bdes
A zero-mean random variable $Z$ is \emph{sub-Gaussian} if there exists
a constant $\csubg\in (0, \infty)$ such that
\begin{eqnarray}
\label{EqnDefnSubgauss}
\E[\exp(t Z)] & \leq & \exp(\csubg^2 \, t^2/2) \qquad \mbox{for all
$t \in \real$.}
\end{eqnarray}
\edes
By the Chernoff bound, this upper bound~\eqref{EqnDefnSubgauss} on the
moment-generating function implies a two-sided tail bound of the form
\begin{eqnarray}
\label{EqnSubgaussChern}
\mprob[|Z| > z] & \leq & 2 \exp \big(- \frac{z^2}{2 \csubg^2}\big).
\end{eqnarray}
Naturally, any zero-mean Gaussian variable with variance $\sigma^2$
satisfies the bounds~\eqref{EqnDefnSubgauss}
and~\eqref{EqnSubgaussChern}.  In addition to the Gaussian case, the
class of sub-Gaussian variates includes any bounded random variable
(e.g., Bernoulli, multinomial, uniform), any random variable with
strictly log-concave density~\cite{BulKoz,Ledoux01}, and any finite
mixture of sub-Gaussian variables.

The following lemma, proved in
Appendix~\ref{APP_LEM_SAM_COV_BOUND_SUBG}, shows that the entries of
the sample covariance based on i.i.d. samples of sub-Gaussian random
vector satisfy an exponential-type tail bound with exponent $\garvesh
= 2$.  The argument is along the lines of a result due to Bickel and
Levina~\cite{BickelLevina2007}, but with more explicit control of the
constants in the error exponent:
\blems
\label{LEM_SAM_COV_BOUND_SUBG}
Consider a zero-mean random vector $(X_1, \ldots, X_\pdim)$ with
covariance $\CovMatStar$ such that each $X_i/\sqrt{\CovMatStar_{ii}}$
is sub-Gaussian with parameter $\csubg$.  Given $\numobs$
i.i.d. samples, the associated sample covariance $\CovHat^\numobs$
satisfies the tail bound
\begin{eqnarray*}
\mprob \big[ |\CovHat^\numobs_{ij }- \CovMatStar_{ij}| > \scdelta
 \big] & \leq & 4 \exp \big \{- \frac{\numobs \scdelta^2}{ \csubgexpr }
 \big \},
\end{eqnarray*}
for all $\scdelta \in \big(0, \max_{i}(\CovMatStar_{ii})\,8(1 + 4
\csubg^2)\big)$.

\elems
Thus, the sample covariance entries the tail condition $\Tail(\sctail,
\scthresh)$ with $\scthresh = \big[\max_{i}(\CovMatStar_{ii})\,8(1 + 4
\csubg^2)\big]^{-1}$, and an exponential-type tail function with
$\garvesh = 2$---namely
\begin{eqnarray}
\label{EqnSubgaussF}
\qquad \sctail(\numobs, \scdelta) = \frac{1}{4} \exp( \ConstStar
\numobs \scdelta^2), & \mbox{with}  & \ConstStar = \big[ \csubgexpr \big]^{-1}
\end{eqnarray}
A little calculation shows that the associated inverse functions take
the form
\begin{equation}
\label{EqnExpInverse}
\sctailInvT(r; \numobs ) \, = \, \sqrt{\frac{\log(4 \,r)}{\ConstStar \,
\numobs}}, \quad \mbox{and} \quad \sctailInvN(r; \scdelta) \, = \,
\frac{\log(4 \, r)}{\ConstStar \scdelta^2}.
\end{equation}

\subsubsection{Tail bounds with moment bounds}
\label{SecSamCovBoundMoment}

In the following lemma, proved in
Appendix~\ref{APP_LEM_SAM_COV_BOUND_MOMENT}, we show that given
i.i.d. observations from random variables with bounded moments, the
sample covariance entries satisfy a polynomial-type tail bound. See
the papers~\cite{Zhao06,Karoui2007} for related results on tail bounds
for variables with bounded moments.
\blems
\label{LEM_SAM_COV_BOUND_MOMENT}
Suppose there exists a positive integer $m$ and scalar $\KconMom \in
\real$ such that for $i = 1,\hdots,\pdim$,
\begin{eqnarray}
\label{EqnBoundedMoments}
\Exs \biggr[ \big(\frac{X_i}{\sqrt{\CovMatStar_{ii}}} \big)^{4 \momentpow}
\biggr] & \leq &   \KconMom.
\end{eqnarray}
For i.i.d. samples $\{\Xsam{\obsind}_i \}_{\obsind=1}^\numobs$, the
sample covariance matrix $\CovHat^\numobs$ satisfies the bound
\begin{eqnarray}
\mprob \big [ \Big| \CovHat^\numobs_{ij} - \CovMatStar_{ij} \Big)\Big|
  > \scdelta \big] & \leq & \frac{\big\{\momentpow^{2\momentpow+1}
  2^{2\momentpow} (\max_i \CovMatStar_{ii}) ^{2\momentpow}\, (\KconMom
  + 1 ) \big\}}{\numobs^{\momentpow}\, \scdelta^{2\momentpow}}.
\end{eqnarray}
\elems
Thus, in this case, the sample covariance satisfies the tail condition
$\Tail(\sctail, \scthresh)$ with $\scthresh = 0$, so that the bound
holds for all $\scdelta \in (0,\infty)$, and with the polynomial-type
tail function
\begin{equation}
\label{EqnPolyF}
\sctail(\numobs,\scdelta) = \ConstStar \numobs^{\momentpow}
\scdelta^{2\momentpow} \quad \mbox{where $\ConstStar =
1/\big\{\momentpow^{2\momentpow+1} 2^{2\momentpow} (\max_i
\CovMatStar_{ii}) ^{2\momentpow}\, (\KconMom + 1 ) \big\}$.}
\end{equation}
Finally, a little calculation shows that in this case, the inverse
tail functions take the form
\begin{equation}
\label{EqnPolyInverse}
\sctailInvT(\numobs,r) \, = \,
\frac{(r/\ConstStar)^{1/2\momentpow}}{\sqrt{\numobs}}, \quad
\mbox{and} \quad \sctailInvN(\scdelta,r) \, = \,
\frac{(r/\ConstStar)^{1/\momentpow}}{\scdelta^{2}}.
\end{equation}

\section{Main results and some consequences}
\label{SecResult}

In this section, we state our main results, and discuss some of their
consequences.  We begin in Section~\ref{SecAssumptions} by stating
some conditions on the true concentration matrix $\ThetaStar$ required in
our analysis, including a particular type of incoherence or
irrepresentability condition.  In Section~\ref{SecEllinf}, we state
our first main result---namely, Theorem~\ref{ThmMain} on consistency
of the estimator $\ThetaHat$, and the rate of decay of its error in
elementwise $\ell_\infty$ norm.  Section~\ref{SecModelCons} is devoted
to Theorem~\ref{ThmModel} on the model selection consistency of the
estimator.  Section~\ref{SecInco} is devoted the relation between the
log-determinant estimator and the ordinary Lasso (neighborhood-based
approach) as methods for graphical model selection; in addition, we
illustrate our irrepresentability assumption for some simple graphs.
Finally, in Section~\ref{SecFrob}, we state and prove some corollaries
of Theorem~\ref{ThmMain}, regarding rates in Frobenius and operator
norms.

\subsection{Conditions on covariance and Hessian}
\label{SecAssumptions}

Our results involve some quantities involving the Hessian of the
log-determinant barrier~\eqref{EqnDefnLogDet}, evaluated at the true
concentration matrix $\ThetaStar$.  Using standard results on matrix
derivatives~\citep{Boyd02}, it can be shown that this Hessian takes
the form
\begin{eqnarray}
\label{EqnDefnHess}
\BigHess & \defn & \nabla^2_{\Theta} g(\Theta) \Big |_{\Theta =
  \ThetaStar} \; = \; \invn{\ThetaStar} \otimes \invn{\ThetaStar},
\end{eqnarray}
where $\otimes$ denotes the Kronecker matrix product.  By definition,
$\BigHess$ is a $\pdim^{2} \times \pdim^{2}$ matrix indexed by vertex
pairs, so that entry $\BigHess_{(j,k), (\ell, m)}$ corresponds to the
second partial derivative $ \frac{\partial^2 g}{\partial \Theta_{jk}
\partial \Theta_{\ell m}}$, evaluated at $\Theta = \ThetaStar$.  When
$X$ has multivariate Gaussian distribution, then $\BigHess$ is the
Fisher information of the model, and by standard results on cumulant
functions in exponential families~\cite{Brown86}, we have the more
specific expression $\BigHess_{(j,k), (\ell, m)} =
\operatorname{cov}\{X_j X_k, \; X_\ell X_m \}$.  For this reason,
$\BigHess$ can be viewed as an edge-based counterpart to the usual
covariance matrix $\CovMatStar$.

We define the set of non-zero off-diagonal entries in the model
concentration matrix $\ThetaStar$:
\begin{eqnarray}
\Eset(\ThetaStar) & \defn & \{ (i,j) \in \vertex \times \vertex \,
	\mid \, i \neq j, \ThetaStar_{ij} \neq 0 \},
\end{eqnarray}
and let $\EsetPlus(\ThetaStar) = \{ \Eset(\ThetaStar) \cup \{(1,1),
\ldots, (\pdim, \pdim) \}$ be the augmented set including the
diagonal.  We let $\EsetPlusComp(\ThetaStar)$ denote the complement of
$\EsetPlus(\ThetaStar)$ in the set $\{1, \ldots, \pdim \} \times \{1,
\ldots, \pdim\}$, corresponding to all pairs $(\ell, m)$ for which
$\ThetaStar_{\ell m} = 0$.  When it is clear from context, we shorten
our notation for these sets to $\EsetPlus$ and $\EsetPlusComp$,
respectively.  Finally, for any two subsets $T$ and $T'$ of $\vertex
\times \vertex$, we use $\BigHess_{T T'}$ to denote the $|T| \times
|T'|$ matrix with rows and columns of $\BigHess$ indexed by $T$ and
$T'$ respectively.

Our main results involve the $\ell_\infty/\ell_\infty$ norm applied to
the covariance matrix $\CovMatStar$, and to the inverse of a sub-block
of the Hessian $\BigHess$.  In particular, we define
\begin{eqnarray}
\label{EqnCovConst}
\KconCov & \defn & \matnorm{\CovMatStar}{\infty} \; = \; \Big(
 \max_{i=1, \ldots ,\pdim} \sum_{j=1}^\pdim |\CovMatStar_{ij}| \Big),
\end{eqnarray}
corresponding to the $\ell_\infty$-operator norm of the true
covariance matrix $\CovMatStar$, and
\begin{eqnarray}
\KconHess & \defn & \matnorm{(\BigHess_{\EsetPlus
\EsetPlus})^{-1}}{\infty} \; = \; \matnorm{([ {\ThetaStar}^{-1}
\otimes {\ThetaStar}^{-1}]_{\EsetPlus \EsetPlus})^{-1}}{\infty}.
\end{eqnarray}
Our analysis keeps explicit track of these quantities, so that they
can scale in a non-trivial manner with the problem dimension $\pdim$.

We assume the Hessian satisfies the following type of \emph{mutual
incoherence or irrepresentable condition}:
\bass
\label{AssInco}
There exists some $\mutinco \in (0,1]$ such that
\begin{eqnarray}
\label{EqnInco}
\matnorm{\BigHess_{\EsetPlusComp \EsetPlus} (\BigHess_{\EsetPlus
\EsetPlus})^{-1}}{\infty} & \leq & (1 - \mutinco).
\end{eqnarray}
\eass

The underlying intuition is that this assumption imposes control on
the influence that the non-edge terms, indexed by $\EsetPlusComp$, can
have on the edge-based terms, indexed by $\EsetPlus$.  It is worth
noting that a similar condition for the Lasso, with the covariance
matrix $\Sigma^*$ taking the place of the matrix $\BigHess$ above, is
necessary and sufficient for support recovery using the ordinary
Lasso~\cite{MeinsBuhl2006,Tropp2006,Wainwright2006_new,Zhao06}.  See
Section~\ref{SecInco} for illustration of the form taken by
Assumption~\ref{AssInco} for specific graphical models.

A remark on notation: although our analysis allows the quantities
$\KconCov, \KconHess$ as well as the model size $\pdim$ and maximum
node-degree $\degmax$ to grow with the sample size $\numobs$, we
suppress this dependence on $\numobs$ in their notation.

\subsection{Rates in elementwise $\ell_\infty$-norm}
\label{SecEllinf}

We begin with a result that provides sufficient conditions on the
sample size $\numobs$ for bounds in the elementwise
$\ell_\infty$-norm.  This result is stated in terms of the tail
function $\sctail$, and its inverses $\sctailInvN$ and $\sctailInvT$ (equations~\eqref{EqnSamTailN} and~\eqref{EqnSamTailT}), and so covers a general range of possible
tail behaviors.  So as to make it more concrete, we follow the general
statement with corollaries for the special cases of exponential-type
and polynomial-type tail functions, corresponding to sub-Gaussian and
moment-bounded variables respectively.  

In the theorem statement, the choice of regularization constant
$\regpar$ is specified in terms of a user-defined parameter $\tunpar >
2$.  Larger choices of $\tunpar$ yield faster rates of convergence in
the probability with which the claims hold, but also lead to more
stringent requirements on the sample size.
\btheos
\label{ThmMain}
Consider a distribution satisfying the incoherence
assumption~\eqref{EqnInco} with parameter \mbox{$\mutinco \in (0,1]$,}
and the tail condition~\eqref{EqnSamTail} with parameters
$\Tail(\sctail, \scthresh)$.  Let $\ThetaHat$ be the unique optimum of
the log-determinant program~\eqref{EqnGaussMLE} with regularization
parameter \mbox{$\regpar = (8/\mutinco) \,
\sctailInvT(\numobs,\pdim^{\tunpar})$} for some $\tunpar > 2$.  Then,
if the sample size is lower bounded as
\begin{eqnarray}
\label{EqnSampleBound}
\numobs & > & \sctailInvN \Biggr( 1 \Big/\max \Big \{ \scthresh,\; 6
 \big(1 + 8\mutinco^{-1} \big) \: \degmax\, \max\{\KconCov
 \KconHess,\KconCov^{3}\KconHess^{2} \} \Big \} , \; \;\pdim^{\tunpar}
 \Biggr),
\end{eqnarray}
then with probability greater than $1-1/\pdim^{\tunpar - 2}
\rightarrow 1$, we have:
\begin{enumerate}
\item[(a)] The estimate $\ThetaHat$ satisfies the elementwise
$\ell_\infty$-bound:
\begin{eqnarray}
\label{EqnEllinfBound}
\| \estim{\Theta} - \ThetaStar\|_\infty & \leq & \big\{2 \big(1 + 8 \mutinco^{-1}\big)\KconHess\big\}\;
\sctailInvT(\numobs,\pdim^{\tunpar}).
\end{eqnarray}
\item[(b)] It specifies an edge set $\Eset(\ThetaHat)$ that is a
subset of the true edge set $\Eset(\ThetaStar)$, and includes all
edges $(i,j)$ with $|\ThetaStar_{ij}| >  \big\{2 \big(1 + 8 \mutinco^{-1}\big)\KconHess\big\}\; \sctailInvT(\numobs,\pdim^{\tunpar})$.
\end{enumerate}
\etheos

If we assume that the various quantities $\KconHess, \KconCov,
\mutinco$ remain constant as a function of $(\numobs, \pdim,
\degmax)$, we have the elementwise $\ell_\infty$ bound \mbox{$\|
\estim{\Theta} - \ThetaStar\|_\infty =
\order(\sctailInvT(\numobs,\pdim^{\tunpar}))$}, so that the inverse
tail function $\sctailInvT(\numobs,\pdim^{\tunpar})$ (see
equation~\eqref{EqnSamTailT}) specifies rate of convergence in the
element-wise $\ell_\infty$-norm.  In the following section, we derive
the consequences of this $\ell_\infty$-bound for two specific tail
functions, namely those of exponential-type with $\garvesh = 2$, and
polynomial-type tails (see Section~\ref{SecDistAssum}).  Turning to
the other factors involved in the theorem statement, the quantities
$\KconCov$ and $\KconHess$ measure the sizes of the entries in the
covariance matrix $\CovMatStar$ and inverse Hessian $(\BigHess)^{-1}$
respectively.  Finally, the factor $(1 + \frac{8}{\mutinco})$ depends
on the irrepresentability assumption~\ref{AssInco}, growing in
particular as the incoherence parameter $\mutinco$ approaches $0$.

\subsubsection{Exponential-type tails}

We now discuss the consequences of Theorem~\ref{ThmMain} for
distributions in which the sample covariance satisfies an
exponential-type tail bound with exponent $\garvesh = 2$.  In
particular, recall from Lemma~\ref{LEM_SAM_COV_BOUND_SUBG} that
such a tail bound holds when the variables are sub-Gaussian.
\def\ConstN{c_1}
\def\ConstTheta{c_2}

\bcors 
\label{CorEllinfSubg}
Under the same conditions as Theorem~\ref{ThmMain}, suppose moreover
that the variables $X_i/\sqrt{\CovMatStar_{ii}}$ are sub-Gaussian with
parameter $\csubg$, and the samples are drawn independently.  Then if
the sample size $\numobs$ satisfies the bound
\begin{eqnarray}
\numobs & > & \Cawful \; \degmax^2 \, (1 + \frac{8}{\mutinco})^2
\;\big (\tunpar \log \pdim + \log 4 \big)
\end{eqnarray}
where $\Cawful \defn \big\{\csubgexprsqrtmultsix \, \max\{\KconCov
\KconHess,\KconCov^{3}\KconHess^{2} \} \big\}^{2}$, then with
probability greater than $1-1/\pdim^{\tunpar -2}$, the estimate
$\ThetaHat$ satisfies the bound,
\begin{eqnarray*}
\| \estim{\Theta} - \ThetaStar\|_\infty & \leq & 
\big\{\csubgexprsqrtmulttwo \, (1 + 8\mutinco^{-1})\KconHess\big\}\; \sqrt{\frac{\tunpar \log \pdim + \log 4}{\numobs}}.
\end{eqnarray*}
\ecors
\spro
From Lemma~\ref{LEM_SAM_COV_BOUND_SUBG}, when the rescaled variables
$X_i/\sqrt{\CovMatStar_{ii}}$ are sub-Gaussian with parameter
$\csubg$, the sample covariance entries satisfies a tail bound
$\Tail(\sctail, \scthresh)$ with with $\scthresh = \big[\max_{i}(\CovMatStar_{ii})\,8(1 + 4 \csubg^2)\big]^{-1}$ 
and $\sctail(\numobs,\scdelta) = (1/4) \exp(\ConstStar \numobs \scdelta^2)$, where \mbox{$\ConstStar =
\big[\csubgexpr\big]^{-1}$.}  As a consequence, for this particular model, the
inverse functions $\sctailInvT(\numobs,\pdim^{\tunpar})$ and
$\sctailInvN(\scdelta,\pdim^{\tunpar})$ take the form
\begin{subequations}
\label{EqnExpTailInv}
\begin{eqnarray}
\sctailInvT(\numobs,\pdim^{\tunpar}) & = &
\sqrt{\frac{\log(4\,\pdim^{\tunpar})}{\ConstStar \, \numobs}} \; = \;
\sqrt{ \csubgexpr} \; \sqrt{\frac{\tunpar \log \pdim + \log 4}{\numobs}},\\
\sctailInvN(\scdelta,\pdim^{\tunpar}) &= &
\frac{\log(4\,\pdim^{\tunpar})} {\ConstStar \scdelta^2} \; = \; 
\csubgexpr  \;
\biggr(\frac{\tunpar \log \pdim + \log 4}{\scdelta^{2}}\biggr).
\end{eqnarray}
\end{subequations}
Substituting these forms into the claim of Theorem~\ref{ThmMain} and
doing some simple algebra yields the stated corollary.
\fpro

When $\KconHess, \KconCov, \mutinco$ remain constant as a function of
$(\numobs, \pdim, \degmax)$, the corollary can be summarized
succinctly as a sample size of \mbox{$\numobs = \Omega(\degmax^2 \log
\pdim)$} samples ensures that an elementwise
$\ell_\infty$ bound \mbox{$\| \estim{\Theta} - \ThetaStar\|_\infty =
\order\big( \sqrt{\frac{\log \pdim}{\numobs}}\big)$} holds with high probability. 
In practice, one frequently considers graphs with maximum node degrees
$\degmax$ that either remain bounded, or that grow sub-linearly with
the graph size (i.e., $\degmax = o(\pdim)$). In such cases, the sample
size allowed by the corollary can be substantially smaller than the
graph size, so that for sub-Gaussian random variables, the method can
succeed in the $\pdim \gg \numobs$ regime.

\subsubsection{Polynomial-type tails}

We now state a corollary for the case of a polynomial-type tail
function, such as those ensured by the case of random variables with
appropriately bounded moments.
\bcors
\label{CorEllinfPoly}
Under the assumptions of Theorem~\ref{ThmMain}, suppose the rescaled
variables $X_i/\sqrt{\CovMatStar_{ii}}$ have $4\momentpow^{th}$
moments upper bounded by $\KconMom$, and the sampling is i.i.d.
Then if the sample size $\numobs$ satisfies the bound
\begin{eqnarray}
\label{EqnPolyTailSampSize}
\numobs & > & \Cawfultwo \, \degmax^{2}\, \big(1 + \frac{8}{\mutinco}
\big)^2\, \pdim^{\tunpar/\momentpow},
\end{eqnarray}
where $\Cawfultwo \defn \big\{12 \momentpow \,[\momentpow (\KconMom +
  1)]^{\frac{1}{2\momentpow}}\, \max_i(\CovMatStar_{ii})\max
\{\KconCov^{2} \KconHess,\KconCov^{4} \KconHess^{2} \} \big\}^{2}$,
then with probability greater than $1-1/\pdim^{\tunpar -2}$, the
estimate $\ThetaHat$ satisfies the bound,
\begin{eqnarray*}
\| \estim{\Theta} - \ThetaStar\|_\infty & \leq & \{4\momentpow
 [\momentpow (\KconMom + 1)]^{\frac{1}{2\momentpow}}\, \big(1 +
 \frac{8}{\mutinco} \big) \KconHess\}\;
 \sqrt{\frac{\pdim^{\tunpar/\momentpow}}{\numobs}}.
\end{eqnarray*}
\ecors
\begin{proof}
Recall from Lemma~\ref{LEM_SAM_COV_BOUND_MOMENT} that when the
rescaled variables $X_i/\sqrt{\CovMatStar_{ii}}$ have bounded
$4\momentpow^{th}$ moments, then the sample covariance $\CovHat$
satisfies the tail condition $\Tail(\sctail, \scthresh)$ with
$\scthresh = 0$, and with $\sctail(\numobs,\scdelta) = \ConstStar
\numobs^{\momentpow} \scdelta^{2\momentpow}$ with $\ConstStar$ defined as
$\ConstStar = 1/\big\{\momentpow^{2\momentpow+1} 2^{2\momentpow} (\max_i
\CovMatStar_{ii}) ^{2\momentpow}\, (\KconMom + 1 ) \big\}$. As a
consequence, for this particular model, the inverse functions take the
form
\begin{subequations}
\label{EqnPolyTailInv}
\begin{eqnarray}
\sctailInvT(\numobs,\pdim^{\tunpar}) & = &
\frac{(\pdim^{\tunpar}/\ConstStar)^{1/2\momentpow}}{\sqrt{\numobs}}
\,=\, \{2\momentpow [\momentpow (\KconMom +
1)]^{\frac{1}{2\momentpow}} \max_i \CovMatStar_{ii} \}\;
\sqrt{\frac{\pdim^{\tunpar/\momentpow}}{\numobs}},\\
\sctailInvN(\scdelta,\pdim^{\tunpar}) & = &
\frac{(\pdim^{\tunpar}/\ConstStar)^{1/\momentpow}}{\scdelta^{2}} \,=\,
\{2\momentpow [\momentpow (\KconMom + 1)]^{\frac{1}{2\momentpow}}
\max_i \CovMatStar_{ii} \}^{2}\;
\big(\frac{\pdim^{\tunpar/\momentpow}}{\scdelta^{2}}\big).
\end{eqnarray}
\end{subequations}
The claim then follows by substituting these expressions into Theorem~\ref{ThmMain} and performing some algebra.
\end{proof}

When the quantities $(\KconHess, \KconCov, \mutinco)$ remain constant
as a function of $(\numobs, \pdim, \degmax)$,
Corollary~\ref{CorEllinfPoly} can be summarized succinctly as
\mbox{$\numobs = \Omega(\degmax^2 \, \pdim^{\tunpar/\momentpow})$}
samples are sufficient to achieve a convergence rate in elementwise
$\ell_\infty$-norm of the order \mbox{$\| \estim{\Theta} -
\ThetaStar\|_\infty = \order\big(
\sqrt{\frac{\pdim^{\tunpar/\momentpow}}{\numobs}}\big)$,} with high
probability.  Consequently, both the required sample size and the rate
of convergence of the estimator are polynomial in the number of
variables $\pdim$.  It is worth contrasting these rates with the case
of sub-Gaussian random variables, where the rates have only
logarithmic dependence on the problem size $\pdim$.

\subsection{Model selection consistency}
\label{SecModelCons}

Part (b) of Theorem~\ref{ThmMain} asserts that the edge set
$\Eset(\ThetaHat)$ returned by the estimator is contained within the
true edge set $\Eset(\ThetaStar)$---meaning that it correctly
\emph{excludes} all non-edges---and that it includes all edges that
are ``large'', relative to the $\sctailInvT(\numobs,\pdim^{\tunpar})$
decay of the error.  The following result, essentially a minor
refinement of Theorem~\ref{ThmMain}, provides sufficient conditions
linking the sample size $\numobs$ and the minimum value
\begin{eqnarray}
\label{EqnDefnThetaMin}
\thetamin & \defn & \min_{(i,j) \in \Eset(\ThetaStar)} |\ThetaStar_{ij}|
\end{eqnarray}
for model selection consistency.  More precisely, define the event
\begin{eqnarray}
\Model(\ThetaHat; \ThetaStar) & \defn & \big \{ \sign(\ThetaHat_{ij})
= \sign(\ThetaStar_{ij}) \quad \forall (i,j) \in \Eset(\ThetaStar)
\big \}
\end{eqnarray}
that the estimator $\ThetaHat$ has the same edge set as $\ThetaStar$,
and moreover recovers the correct signs on these edges.  With this
notation, we have:
\btheos
\label{ThmModel}

Under the same conditions as Theorem~\ref{ThmMain}, suppose that
the sample size satisfies the lower bound
\begin{eqnarray}
\label{EqnNumobsModel}
\numobs & > & \sctailInvN \Biggr( 1 \big/\max \Big \{ 2 \KconHess (1 +
8\mutinco^{-1})\, \thetamin^{-1}, \; \scthresh, \; 6 \big (1 +
8\mutinco^{-1} \big) \: \degmax\, \max\{\KconCov
\KconHess,\KconCov^{3}\KconHess^{2} \} \Big\} , \; \;\pdim^{\tunpar}
\Biggr).
\end{eqnarray}
Then the estimator is model selection consistent with high probability
as $\pdim \rightarrow \infty$,
\begin{eqnarray}
\mprob \big[ \Model(\ThetaHat; \ThetaStar) \big] & \geq & 1 -
1/\pdim^{\tunpar - 2} \; \rightarrow \; 1.
\end{eqnarray}
\etheos

In comparison to Theorem~\ref{ThmMain}, the sample size
requirement~\eqref{EqnNumobsModel} differs only in the additional term
$\frac{2 \KconHess (1 + \frac{8}{\mutinco})}{\thetamin}$ involving the
minimum value.  This term can be viewed as constraining how quickly
the minimum can decay as a function of $(\numobs, \pdim)$, as we
illustrate with some concrete tail functions.

\subsubsection{Exponential-type tails} 

Recall the setting of Section~\ref{SecSamCovBoundSubG}, where the
random variables $\{\SampVar_{i}/\sqrt{\CovMatStar_{ii}}\}$ are
sub-Gaussian with parameter $\csubg$.  Let us suppose that the
parameters $(\KconHess, \KconCov, \mutinco)$ are viewed as constants
(not scaling with $(\pdim, \degmax)$.  Then, using the
expression~\eqref{EqnExpTailInv} for the inverse function
$\sctailInvN$ in this setting, a corollary of Theorem~\ref{ThmModel}
is that a sample size 
\begin{eqnarray}
\label{EqnModelSampSub}
\numobs & = & \Omega \big( (\degmax^2 + \thetamin^{-2}) \, \tunpar \log
\pdim \big)
\end{eqnarray}
is sufficient for model selection consistency with probability greater
than $1-1/\pdim^{\tunpar-2}$.  Alternatively, we can state that $\numobs
= \Omega(\tunpar \degmax^2 \log \pdim)$ samples are sufficient, as
along as the minimum value scales as \mbox{$\thetamin =
\Omega(\sqrt{\frac{\log \pdim}{\numobs}})$.}

\subsubsection{Polynomial-type tails} 

Recall the setting of Section~\ref{SecSamCovBoundMoment}, where the
rescaled random variables $X_i/\sqrt{\CovMatStar_{ii}}$ have bounded
$4\momentpow^{th}$ moments. Using the expression~\eqref{EqnPolyTailInv}
for the inverse function $\sctailInvN$ in this setting, a corollary of
Theorem~\ref{ThmModel} is that a sample size
\begin{eqnarray}
\label{EqnModelSampPoly}
\numobs & = & \Omega\big( (\degmax^2 + \thetamin^{-2})\,
\pdim^{\tunpar/\momentpow} \big)
\end{eqnarray}
is sufficient for model selection consistency with probability greater
than $1-1/\pdim^{\tunpar-2}$.  Alternatively, we can state than $\numobs
= \Omega(\degmax^2 \pdim^{\tunpar/\momentpow})$ samples are
sufficient, as long as the minimum value scales as \mbox{$\thetamin =
\Omega(\pdim^{\tunpar/(2\momentpow)}/{\sqrt{\numobs}})$.}

\subsection{Comparison to neighbor-based graphical model selection}
\label{SecInco}

Suppose that $X$ follows a multivariate Gaussian distribution, so that
the structure of the concentration matrix $\ThetaStar$ specifies the
structure of a Gaussian graphical model.  In this case, it is
interesting to compare our sufficient conditions for graphical model
consistency of the log-determinant approach, as specified in
Theorem~\ref{ThmModel}, to those of the neighborhood-based method,
first proposed by \citet{MeinsBuhl2006}. The latter method estimates
the full graph structure by performing an $\ell_1$-regularized linear
regression (Lasso)---of the form $X_i = \sum_{j \neq i} \theta_{ij}
X_j + W$--- of each node on its neighbors and using the support of the
estimated regression vector $\theta$ to predict the neighborhood set.
These neighborhoods are then combined, by either an OR rule or an AND
rule, to estimate the full graph.  Various aspects of the
high-dimensional model selection consistency of the Lasso are now
understood~\cite{MeinsBuhl2006,Wainwright2006_new,Zhao06}; for
instance, it is known that mutual incoherence or irrepresentability
conditions are necessary and sufficient for its
success~\cite{Tropp2006,Zhao06}. In terms of scaling,
Wainwright~\cite{Wainwright2006_new} shows that the Lasso succeeds
with high probability if and only if the sample size scales as
\mbox{$\numobs \asymp c (\{\degmax + \theta_{\operatorname{min}}^{-2}
\} \log \pdim)$,} where $c$ is a constant determined by the covariance
matrix $\CovMatStar$.  By a union bound over the $\pdim$ nodes in the
graph, it then follows that the neighbor-based graph selection method
in turn succeeds with high probability if $\numobs = \Omega(\{\degmax
+ \thetamin^{-2} \} \log \pdim)$.

For comparison, consider the application of Theorem~\ref{ThmModel} to
the case where the variables are sub-Gaussian (which includes the
Gaussian case). For this setting, we have seen that the scaling
required by Theorem~\ref{ThmModel} is $\numobs = \Omega( \{ \degmax^2
+ \thetamin^{-2} \} \log \pdim)$, so that the dependence of the
log-determinant approach in $\thetamin$ is identical, but it depends
quadratically on the maximum degree $\degmax$.  We suspect that that
the quadratic dependence $\degmax^2$ might be an artifact of our
analysis, but have not yet been able to reduce it to $\degmax$.
Otherwise, the primary difference between the two methods is in the
nature of the irrepresentability assumptions that are imposed: our
method requires Assumption~\ref{AssInco} on the Hessian $\BigHess$,
whereas the neighborhood-based method imposes this same type of
condition on a set of $\pdim$ covariance matrices, each of size
$(\pdim -1) \times (\pdim-1)$, one for each node of the graph.  Below
we show two cases where the Lasso irrepresentability condition holds,
while the log-determinant requirement fails.  However, in general, we
do not know whether the log-determinant irrepresentability strictly
dominates its analog for the Lasso.

\subsubsection{Illustration of irrepresentability: Diamond graph} 
\begin{figure}[htb]
\begin{center}
\begin{tabular}{ccc}
\psfrag{#1#}{$1$} \psfrag{#2#}{$2$} \psfrag{#3#}{$3$}
\psfrag{#4#}{$4$} \widgraph{.3\textwidth}{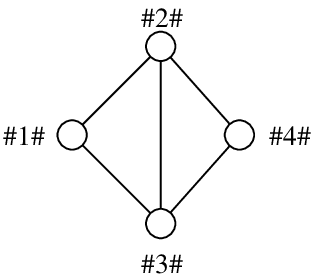} & \hspace*{.2in}
& \psfrag{#1#}{$1$} \psfrag{#2#}{$2$} \psfrag{#3#}{$3$}
\psfrag{#4#}{$4$} \widgraph{.3\textwidth}{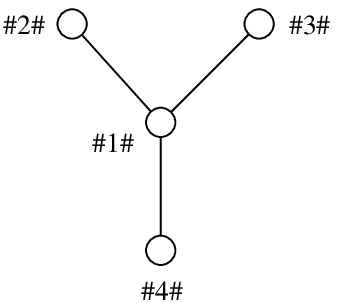} \\
(a) & & (b)
\end{tabular}
\end{center}
\caption{(a) Graph of the example discussed by~\citet{Meins2008}.  (b)
A simple $4$-node star graph.}
\label{FigSimpGraph}
\end{figure}

Consider the following Gaussian graphical model example from
\citet{Meins2008}.  Figure~\ref{FigSimpGraph}(a) shows a
diamond-shaped graph $G = (V,E)$, with vertex set $V = \{1,2,3,4\}$
and edge-set as the fully connected graph over $V$ with the edge
$(1,4)$ removed.
The covariance matrix $\CovMatStar$ is parameterized by the
correlation parameter $\rho \in [0,1/\sqrt{2}]$: the diagonal entries
are set to $\CovMatStar_{i} = 1$, for all $i \in \vertex$; the entries
corresponding to edges are set to $\CovMatStar_{ij} = \rho$ for $(i,j)
\in \edge \backslash \{(2,3)\}$, $\CovMatStar_{23} = 0$; and finally
the entry corresponding to the non-edge is set as $\CovMatStar_{14} =
2 \rho^2$. \citet{Meins2008} showed that the $\ell_1$-penalized
log-determinant estimator $\ThetaHat$ fails to recover the graph
structure, for any sample size, if $\rho > -1 + (3/2)^{1/2} \approx
0.23$.  It is instructive to compare this necessary condition to the
sufficient condition provided in our analysis, namely the incoherence
Assumption~\ref{AssInco} as applied to the Hessian $\BigHess$.  For
this particular example, a little calculation shows that
Assumption~\ref{AssInco} is equivalent to the constraint
\begin{eqnarray*}
4 |\rho| (|\rho| + 1) & < & 1,
\end{eqnarray*}
an inequality which holds for all $\rho \in (-0.2017, 0.2017)$.  Note
that the upper value $0.2017$ is just below the necessary threshold
discussed by \citet{Meins2008}. On the other hand, the
irrepresentability condition for the Lasso requires only that $2
|\rho| < 1$, i.e., $\rho \in (-0.5,0.5)$.  Thus, in the regime $|\rho|
\in [0.2017,0.5)$, the Lasso irrepresentability condition holds while
the log-determinant counterpart fails.

\subsubsection{Illustration of irrepresentability: Star graphs} 
A second interesting example is the star-shaped graphical model,
illustrated in Figure~\ref{FigSimpGraph}(b), which consists of a
single hub node connected to the rest of the spoke nodes. We consider
a four node graph, with vertex set $V = \{1,2,3,4\}$ and edge-set $E =
\{(1,s) \mid s \in \{2,3,4\}\}$. The covariance matrix $\CovMatStar$
is parameterized the correlation parameter $\rho \in [-1,1]$: the
diagonal entries are set to $\CovMatStar_{ii} = 1$, for all $i \in V$;
the entries corresponding to edges are set to $\CovMatStar_{ij} =
\rho$ for $(i,j) \in E$; while the non-edge entries are set as
$\CovMatStar_{ij} = \rho^2$ for $(i,j) \notin E$.  Consequently, for
this particular example, Assumption~\ref{AssInco} reduces to the
constraint $|\rho| (|\rho| + 2) < 1$, which holds for all $\rho \in
(-0.414, 0.414)$. The irrepresentability condition for the Lasso on
the other hand allows the full range $\rho \in (-1,1)$. Thus there is
again a regime, $|\rho| \in [0.414,1)$, where the Lasso
irrepresentability condition holds while the log-determinant
counterpart fails.

\subsection{Rates in Frobenius and spectral norm}
\label{SecFrob}
We now derive some corollaries of Theorem~\ref{ThmMain} concerning
estimation of $\ThetaStar$ in Frobenius norm, as well as the spectral
norm.  Recall that $\spindex = |\Eset(\ThetaStar)|$ denotes the total
number of off-diagonal non-zeros in $\ThetaStar$.
\bcors
\label{CorOperatorNorm} 
Under the same assumptions as Theorem~\ref{ThmMain}, with probability
at least $1 - 1/\pdim^{\tunpar - 2}$, the
estimator $\estim{\Theta}$ satisfies
\begin{subequations}
\begin{eqnarray}
\label{EqnPrecFrob}
\matnorm{\ThetaHat - \ThetaStar}{F} & \leq & \big\{2 \KconHess \big(1
 + \frac{8}{\mutinco}\big)\big\} \, \sqrt{\spindex +\pdim} \;
 \sctailInvT(\numobs,\pdim^{\tunpar}), \qquad \mbox{and} \\
\label{EqnPrecSpectral}
\matnorm{\ThetaHat - \ThetaStar}{2} & \leq & \big\{2 \KconHess \big(1
 + \frac{8}{\mutinco}\big) \big\}\, \min \{\sqrt{\spindex +\pdim}, \,
 \degmax \} \; \sctailInvT(\numobs,\pdim^{\tunpar}).
\end{eqnarray}
\end{subequations}
\ecors
\spro
\newcommand{\Upper}{\ensuremath{\nu}}

With the shorthand notation $\Upper \defn 2 \KconHess (1 + 8/\mutinco)
\; \sctailInvT(\numobs,\pdim^{\tunpar})$, Theorem~\ref{ThmMain}
guarantees that, with probability at least $1 - 1/\pdim^{\tunpar -
2}$, $\|\ThetaHat - \ThetaStar\|_\infty \leq \Upper$.  Since the edge
set of $\ThetaHat$ is a subset of that of $\ThetaStar$, and
$\ThetaStar$ has at most $\pdim + \spindex$ non-zeros (including the
diagonal), we conclude that
\begin{eqnarray*}
\matnorm{\ThetaHat - \ThetaStar}{F} & = & \big[ \sum_{i=1}^\pdim
(\ThetaHat_{ii} - \ThetaStar_{ii})^2 + \sum_{(i,j) \in \Eset}
(\ThetaHat_{ij} - \ThetaStar_{ij})^2 \big]^{1/2} \\
& \leq & \Upper \; \sqrt{\spindex + \pdim},
\end{eqnarray*}
from which the bound~\eqref{EqnPrecFrob} follows.  On the other hand,
for a symmetric matrix, we have
\begin{eqnarray}\label{EqnPrecInftyOp}
\matnorm{\ThetaHat- \ThetaStar}{2} & \leq & \matnorm{\ThetaHat -
\ThetaStar}{\infty} \; \leq \; \degmax \Upper,
\end{eqnarray}
using the definition of the $\Upper_\infty$-operator norm, and the fact
that $\ThetaHat$ and $\ThetaStar$ have at most $\degmax$ non-zeros per
row.  Since the Frobenius norm upper bounds the spectral norm, the
bound~\eqref{EqnPrecSpectral} follows.

\fpro

\subsubsection{Exponential-type tails}
For the exponential tail function case where the rescaled random
variables $X_i/\sqrt{\CovMatStar_{ii}}$ are sub-Gaussian with
parameter $\csubg$, we can use the expression~\eqref{EqnExpTailInv}
for the inverse function $\sctailInvT$ to derive rates in Frobenius
and spectral norms.  When the quantities $\KconHess, \KconCov,
\mutinco$ remain constant, these bounds can be summarized succinctly
as a sample size \mbox{$\numobs = \Omega(\degmax^2 \log \pdim)$} is
sufficient to guarantee the bounds
\begin{subequations}
\begin{eqnarray}
\matnorm{\estim{\Theta} - \ThetaStar}{F} & = &
\order\biggr(\sqrt{\frac{(\spindex + \pdim)\,\log \pdim}{\numobs}}\,\biggr),
\quad \mbox{and} \\
\matnorm{\estim{\Theta} - \ThetaStar}{2} & = &
\order\biggr(\sqrt{\frac{\min\{\spindex + \pdim,\,\degmax^{2}\} \, \log
\pdim}{\numobs}}\,\biggr),
\end{eqnarray}
\end{subequations}
with probability at least $1 - 1/\pdim^{\tunpar - 2}$.

\subsubsection{Polynomial-type tails}

Similarly, let us again consider the polynomial tail case, in which
the rescaled variates $X_i/\sqrt{\CovMatStar_{ii}}$ have bounded $4
\momentpow^{th}$ moments and the samples are drawn i.i.d.  Using the
expression~\eqref{EqnPolyTailInv} for the inverse function we can
derive rates in the Frobenius and spectral norms.  When the quantities
$\KconHess, \KconCov, \mutinco$ are viewed as constant, we are
guaranteed that a sample size \mbox{$\numobs = \Omega(\degmax^2 \,
\pdim^{\tunpar/\momentpow})$} is sufficient to guarantee the bounds
\begin{subequations}
\begin{eqnarray}
 \matnorm{\estim{\Theta} - \ThetaStar}{F} & = &
 \order\biggr(\sqrt{\frac{(\spindex +
 \pdim)\,\pdim^{\tunpar/\momentpow}}{\numobs}}\,\biggr),\textrm{ and } \\
\matnorm{\estim{\Theta} - \ThetaStar}{2} & = &
\order\biggr(\sqrt{\frac{\min\{\spindex + \pdim,\,\degmax^{2}\} \,
\pdim^{\tunpar/\momentpow}}{\numobs}}\,\biggr),
\end{eqnarray}
\end{subequations}
with probability at least $1 - 1/\pdim^{\tunpar - 2}$.


\subsection{Rates for the covariance matrix estimate}
Finally, we describe some bounds on the estimation of the covariance
matrix $\CovMatStar$.  By Lemma~\ref{LEM_MLE_CHARAC}, the estimated
concentration matrix $\ThetaHat$ is positive definite, and hence can
be inverted to obtain an estimate of the covariance matrix, which we
denote as $\CovEstim \defn (\ThetaHat)^{-1}$.

\bcors
\label{CorCovBound}
Under the same assumptions as Theorem~\ref{ThmMain}, with probability
at least $1 - 1/\pdim^{\tunpar - 2}$, the following bounds hold.
\begin{enumerate}
\item[(a)] The element-wise $\ell_{\infty}$ norm of the deviation
 $(\CovEstim - \CovMatStar)$ satisfies the bound
\begin{eqnarray}
\label{EqnCovInftyBound}
\vecnorm{\CovEstim - \CovMatStar}{\infty} & \leq & \Cawthree,
	[\sctailInvT(\numobs,\pdim^{\tunpar})] + \Cawfour
	\degmax\, [\sctailInvT(\numobs,\pdim^{\tunpar})]^{2}
\end{eqnarray}
where $\Cawthree = 2 \KconCov^{2} \KconHess\Big(1 +
\frac{8}{\mutinco}\Big)$ and $\Cawfour = 6 \KconCov^{3}
\KconHess^{2}\Big(1 + \frac{8}{\mutinco}\Big)^{2}$.
\item[(b)] The $\ell_2$ operator-norm of the deviation $(\CovEstim -
\CovMatStar)$ satisfies the bound
\begin{eqnarray}
\label{EqnCovSpectralBound}
\matnorm{\CovEstim - \CovMatStar}{2} & \leq & \Cawthree \, \degmax \,
	[\sctailInvT(\numobs,\pdim^{\tunpar})] + \Cawfour \degmax^{2}
	\, [\sctailInvT(\numobs,\pdim^{\tunpar})]^{2}.
\end{eqnarray}
\end{enumerate}
\ecors
The proof involves certain lemmata and derivations that are parts of
the proofs of Theorems~\ref{ThmMain} and \ref{ThmModel}, so that we
defer it to Section~\ref{SecCorCovProof}.

\section{Proofs of main result}
\label{SecProof}

In this section, we work through the proofs of Theorems~\ref{ThmMain}
and~\ref{ThmModel}.  We break down the proofs into a sequence of
lemmas, with some of the more technical aspects deferred to
appendices.

Our proofs are based on a technique that we call a \emph{primal-dual
witness method}, used previously in analysis of the
Lasso~\cite{Wainwright2006_new}.  It involves following a specific
sequence of steps to construct a pair $(\ThetaWitness, \ZWit)$ of
symmetric matrices that together satisfy the optimality conditions
associated with the convex program~\eqref{EqnGaussMLE} \emph{with high
probability}.  Thus, when the constructive procedure succeeds,
$\ThetaWitness$ is \emph{equal} to the unique solution
$\estim{\Theta}$ of the convex program~\eqref{EqnGaussMLE}, and
$\ZWit$ is an optimal solution to its dual.  In this way, the
estimator $\estim{\Theta}$ inherits from $\ThetaWitness$ various
optimality properties in terms of its distance to the truth
$\ThetaStar$, and its recovery of the signed sparsity pattern.  To be
clear, our procedure for constructing $\ThetaWitness$ is \emph{not} a
practical algorithm for solving the log-determinant
problem~\eqref{EqnGaussMLE}, but rather is used as a proof technique
for certifying the behavior of the $M$-estimator~\eqref{EqnGaussMLE}.

\subsection{Primal-dual witness approach}
\label{SecPrimalDualWitness}
As outlined above, at the core of the primal-dual witness method are the standard convex
optimality conditions that characterize the optimum $\ThetaHat$ of the
convex program~\eqref{EqnGaussMLE}.  For future reference, we note
that the sub-differential of the norm $\ellreg{\cdot}$ evaluated at
some $\Theta$ consists the set of all symmetric matrices $Z \in
\real^{\pdim \times \pdim}$ such that
\begin{eqnarray}\label{EqnSubGradDefn}
Z_{ij} & = & \begin{cases} 0 & \mbox{if $i =j$} \\ \sign(\Theta_{ij})
& \mbox{if $i \neq j$ and $\Theta_{ij} \neq 0$} \\ \in [-1, +1] &
\mbox{if $i \neq j$ and $\Theta_{ij} = 0$.}
             \end{cases}
\end{eqnarray}
The following result is proved in Appendix~\ref{AppLemMLECharac}:
\blems
\label{LEM_MLE_CHARAC}
For any $\regpar > 0$ and sample covariance $\SigHat$ with strictly
positive diagonal, the $\ell_1$-regularized log-determinant
problem~\eqref{EqnGaussMLE} has a unique solution $\ThetaHat \succ 0$
characterized by
\begin{eqnarray}
\label{EqnZeroSubgrad}
\SigHat - \ThetaHat^{-1} + \regpar \Zhat & = & 0,
\end{eqnarray}
where $\Zhat$ is an element of the subdifferential $\partial
\ellreg{\ThetaHat}$.  
\elems

Based on this lemma, we construct the primal-dual witness solution
$(\ThetaWitness, \ZWit)$ as follows:
\begin{enumerate}
\item[(a)]  We determine the
matrix $\ThetaWitness$ by solving the restricted log-determinant
problem
\begin{eqnarray}
\label{EqnRestricted}
\ThetaWitness & \defn & \arg \min_{\Theta \succ 0, \;
\Theta_{\EsetPlusComp} = 0} \big \{\tracer{\Theta}{\SigHat} - \log
\det(\Theta) + \regpar \ellreg{\Theta} \big \}.
\end{eqnarray}
Note that by construction, we have $\ThetaWitness \succ 0$, and
moreover $\ThetaWitness_{\EsetPlusComp} = 0$.
\item[(b)] We choose $\ZWit_{\EsetPlus}$ as a member of the
sub-differential of the regularizer $\ellreg{\cdot}$, evaluated
at $\ThetaWit$.
\item[(c)] We set $\ZWit_{\EsetPlusComp}$ as
\begin{eqnarray}
 \ZWit_{\EsetPlusComp} &=& \frac{1}{\regpar}\big\{-
 \SigHat_{\EsetPlusComp} + [\invn{\ThetaWit}]_{\EsetPlusComp}\big\},
\end{eqnarray}
which ensures that constructed matrices $(\ThetaWit,\ZWit)$ satisfy
the optimality condition~\eqref{EqnZeroSubgrad}.
\item[(d)]  We verify the \emph{strict dual feasibility} condition
\begin{eqnarray*}
|\ZWit_{ij}| & < & 1 \quad \mbox{for all $(i,j) \in \EsetPlusComp$}.
\end{eqnarray*}

\end{enumerate}
To clarify the nature of the construction, steps (a) through (c)
suffice to obtain a pair $(\ThetaWit,\ZWit)$ that satisfy the
optimality conditions~\eqref{EqnZeroSubgrad}, but do \emph{not}
guarantee that $\ZWit$ is an element of sub-differential $\partial
\ellreg{\ThetaWit}$.  By construction, specifically step (b) of the
construction ensures that the entries $\ZWit$ in $\EsetPlus$ satisfy
the sub-differential conditions, since $\ZWit_{\EsetPlus}$ is a member
of the sub-differential of $\partial \ellreg{\ThetaWit_{\EsetPlus}}$.
The purpose of step (d), then, is to verify that the remaining
elements of $\ZWit$ satisfy the necessary conditions to belong
to the sub-differential.

If the primal-dual witness construction succeeds, then it acts as a
\emph{witness} to the fact that the solution $\ThetaWit$ to the
restricted problem~\eqref{EqnRestricted} is equivalent to the solution
$\ThetaHat$ to the original (unrestricted)
problem~\eqref{EqnGaussMLE}.  We exploit this fact in our proofs of
Theorems~\ref{ThmMain} and \ref{ThmModel} that build on this: we first show
that the primal-dual witness technique succeeds with high-probability,
from which we can conclude that the support of the optimal solution
$\ThetaHat$ is contained within the support of the true $\ThetaStar$.
In addition, we exploit the characterization of $\ThetaHat$ provided
by the primal-dual witness construction to establish the elementwise
$\ell_\infty$ bounds claimed in Theorem~\ref{ThmMain}.
Theorem~\ref{ThmModel} requires checking, in addition, that certain
sign consistency conditions hold, for which we require lower bounds on
the value of the minimum value $\thetamin$.

In the analysis to follow, some additional notation is useful.  We let
$\Wnoise$ denote the ``effective noise'' in the sample covariance
matrix $\SigHat$, namely
\begin{eqnarray}
\label{EqnWdefn}
\Wnoise & \defn & \SigHat - (\ThetaStar)^{-1}.
\end{eqnarray}
Second, we use $\Delta = \ThetaWitness - \ThetaStar$ to measure the
discrepancy between the primal witness matrix $\ThetaWit$ and the
truth $\ThetaStar$.  Finally, recall the log-determinant barrier $g$
from equation~\eqref{EqnDefnLogDet}. We let $\Res(\Delta)$ denote the
difference of the gradient $\nabla g(\ThetaWit) =
\invn{\ThetaWitness}$ from its first-order Taylor expansion around
$\ThetaStar$. Using known results on the first and second derivatives
of the log-determinant function (see p. 641 in Boyd and
Vandenberghe~\cite{Boyd02}), this remainder takes the form
\begin{eqnarray}
\label{EqnRdefn}
\Res(\Delta) & = & \invn{\ThetaWitness} - \invn{\ThetaStar} +
{\ThetaStar}^{-1} \Delta {\ThetaStar}^{-1}.
\end{eqnarray}

\subsection{Auxiliary results}

We begin with some auxiliary lemmata, required in the proofs of our
main theorems.  In Section~\ref{SecStrictDual}, we provide sufficient
conditions on the quantities $\Wnoise$ and $\Res$ for the strict dual
feasibility condition to hold. In Section~\ref{SecRemainder}, we
control the remainder term $\Res(\Delta)$ in terms of $\Delta$, while
in Section~\ref{SecEllinfBound}, we control $\Delta$ itself, providing
elementwise $\ell_\infty$ bounds on $\Delta$.  In
Section~\ref{SecSignConsis}, we show that under appropriate conditions
on the minimum value $\thetamin$, the bounds in the earlier lemmas
guarantee that the sign consistency condition holds.  All of the
analysis in these sections is \emph{deterministic} in nature.  In
Section~\ref{SecNoise}, we turn to the probabilistic component of the
analysis, providing control of the noise $\Wnoise$ in the sample
covariance matrix.  Finally, the proofs of Theorems~\ref{ThmMain}
and~\ref{ThmModel} follows by using this probabilistic control of
$\Wnoise$ and the stated conditions on the sample size to show that
the deterministic conditions hold with high probability.

\subsubsection{Sufficient conditions for strict dual feasibility}
\label{SecStrictDual}

We begin by stating and proving a lemma that provides sufficient
(deterministic) conditions for strict dual feasibility to hold, so
that $\|\ZWit_{\EsetPlusComp}\|_\infty < 1$.
\blems[Strict dual feasibility]
\label{LemStrictDual}
Suppose that
\begin{eqnarray}
\label{EqnDetSuff}
\max \big \{ \|\Wnoise\|_\infty, \; \|\Res(\Delta)\|_\infty \big \}  &
\leq & \frac{\mutinco \, \regpar}{8}.
\end{eqnarray}
Then the matrix $\ZWit_{\EsetPlusComp}$ constructed in step (c)
satisfies $\|\ZWit_{\EsetPlusComp}\|_\infty < 1$, and therefore
$\ThetaWitness = \ThetaHat$.
\elems
\begin{proof}
Using the definitions~\eqref{EqnWdefn} and~\eqref{EqnRdefn}, we can
re-write the stationary condition~\eqref{EqnZeroSubgrad} in an
alternative but equivalent form
\begin{eqnarray}
\label{EqnLME}
\invn{\ThetaStar} \Delta \invn{\ThetaStar} + \Wnoise - \Res(\Delta) +
\regpar \Ztil & = 0.
\end{eqnarray}
This is a linear-matrix equality, which can be re-written as an
ordinary linear equation by ``vectorizing'' the matrices.  We use the
notation $\smallvec(A)$, or equivalently $\myvec{A}$ for the
$\pdim^2$-vector version of the matrix $A \in \real^{\pdim \times
\pdim}$, obtained by stacking up the rows into a single column vector.
In vectorized form, we have
\begin{equation*}
\smallvec \big(\invn{\ThetaStar} \Delta \invn{\ThetaStar} \big) = \big
(\invn{\ThetaStar} \otimes \invn{\ThetaStar} \big) \myvec{\Delta} \; =
\; \BigHess \myvec{\Delta}.
\end{equation*}
In terms of the disjoint decomposition $\EsetPlus$ and
$\EsetPlusComp$, equation~\eqref{EqnLME} can be re-written as two
blocks of linear equations as follows:
\begin{subequations}
\begin{eqnarray}
\label{EqnStatBlockS} \BigHess_{\EsetPlus
\EsetPlus} \myvec{\Delta}_{\EsetPlus} + \myvec{\Wnoise}_{\EsetPlus} -
\myvec{\Res}_{\EsetPlus} + \regpar \myvec{\ZWit}_{\EsetPlus} & = & 0
\\
\label{EqnStatBlockScomp}
\BigHess_{\EsetPlusComp \EsetPlus} \myvec{\Delta}_{\EsetPlus} +
\myvec{\Wnoise}_{\EsetPlusComp} - \myvec{\Res}_{\EsetPlusComp} +
\regpar \myvec{\ZWit}_{\EsetPlusComp} & = & 0.
\end{eqnarray}
\end{subequations}
Here we have used the fact that $\Delta_{\EsetPlusComp} = 0$ by
construction.

Since $\BigHess_{\EsetPlus \EsetPlus}$ is invertible, we can solve for
$\myvec{\Delta}_{\EsetPlus}$ from equation~\eqref{EqnStatBlockS} as
follows:
\begin{eqnarray*}
\myvec{\Delta}_{\EsetPlus} & = & \inv{\BigHess_{\EsetPlus\EsetPlus}}
\big[-\myvec{\Wnoise}_{\EsetPlus} + \myvec{\Res}_{\EsetPlus} - \regpar
\myvec{\ZWit_\EsetPlus} \big].
\end{eqnarray*}
Substituting this expression into equation~\eqref{EqnStatBlockScomp},
we can solve for $\ZWit_{\EsetPlusComp}$ as follows:
\begin{eqnarray}
\myvec{\ZWit}_{\EsetPlusComp} &= & -\frac{1}{\regpar}
\BigHess_{\EsetPlusComp\EsetPlus}\myvec{\Delta}_{\EsetPlus} +
\frac{1}{\regpar} \myvec{\Res}_{\EsetPlusComp} - \frac{1}{\regpar}
\myvec{\Wnoise}_{\EsetPlusComp} \nonumber\\
& = & -\frac{1}{\regpar} \BigHess_{\EsetPlusComp\EsetPlus}
\inv{\BigHess_{\EsetPlus\EsetPlus}}(\myvec{\Wnoise}_{\EsetPlus} -
\myvec{\Res}_{\EsetPlus}) + \BigHess_{\EsetPlusComp\EsetPlus}
\inv{\BigHess_{\EsetPlus\EsetPlus}} \myvec{\ZWit}_{\EsetPlus} -
\frac{1}{\regpar} (\myvec{\Wnoise}_{\EsetPlusComp} -
\myvec{\Res}_{\EsetPlusComp}).
\end{eqnarray}
Taking the $\ell_\infty$ norm of both sides yields
\begin{multline*}
\|\myvec{\ZWit}_{\EsetPlusComp}\|_{\infty} \leq \invl
\matnorm{\BigHess_{\EsetPlusComp\EsetPlus}
\inv{\BigHess_{\EsetPlus\EsetPlus}}}{\infty}
(\|\myvec{\Wnoise}_{\EsetPlus}\|_{\infty} +
\|\myvec{\Res}_{\EsetPlus}\|_{\infty}) \\
+ \matnorm{\BigHess_{\EsetPlusComp\EsetPlus}
\inv{\BigHess_{\EsetPlus\EsetPlus}}}{\infty}
\|\myvec{\ZWit}_{\EsetPlus}\|_{\infty} + \invl
(\|\myvec{\Wnoise}_{\EsetPlus}\|_{\infty} +
\|\myvec{\Res}_{\EsetPlus}\|_{\infty}).
\end{multline*}
Recalling Assumption~\ref{AssInco}---namely, that
$\matnorm{\BigHess_{\EsetPlusComp\EsetPlus}
\inv{\BigHess_{\EsetPlus\EsetPlus}}}{\infty} \le (1 - \mutinco)$---we
have
\begin{eqnarray*}
\|\myvec{\ZWit}_{\EsetPlusComp}\|_{\infty} & \leq &
\frac{2-\mutinco}{\regpar} \,
(\|\myvec{\Wnoise}_{\EsetPlus}\|_{\infty} +
\|\myvec{\Res}_{\EsetPlus}\|_{\infty}) + (1 - \mutinco),
\end{eqnarray*}
where we have used the fact that $\|\myvec{\ZWit}_\EsetPlus\|_{\infty}
\leq 1$, since $\ZWit$ belongs to the sub-differential of the norm
$\ellreg{\cdot}$ by construction.  Finally, applying
assumption~\eqref{EqnDetSuff} from the lemma statement, we have
\begin{eqnarray*}
\|\myvec{\ZWit}_{\EsetPlusComp}\|_{\infty} & \leq &
\frac{(2-\mutinco)}{\regpar} \, \big( \frac{\mutinco \regpar}{4}) +
(1-\mutinco) \\
& \leq &  \frac{\mutinco}{2} + (1-\mutinco)  \: < \; 1,
\end{eqnarray*}
as claimed.

\end{proof}


\subsubsection{Control of remainder term}
\label{SecRemainder}

Our next step is to relate the behavior of the remainder
term~\eqref{EqnRdefn} to the deviation $\Delta = \ThetaWit -
\ThetaStar$.
\blems[Control of remainder]
\label{LEM_R_CONV}
Suppose that the elementwise $\ell_\infty$ bound $\|\Delta\|_\infty
\leq \frac{1}{3 \, \KconCov \degmax}$ holds.  Then:
\begin{eqnarray}
\label{EqnRemExpand}
\Res(\Delta) & = & \invn{\opt{\conc}} \Delta \invn{\opt{\conc}} \Delta
J \invn{\opt{\conc}},
\end{eqnarray}
where $J \defn \sum_{k=0}^{\infty} (-1)^{k} \big(\invn{\opt{\conc}}
\Delta\big)^{k}$ has norm $\matnorm{J^T}{\infty} \leq 3/2$.  Moreover,
in terms of the elementwise $\ell_\infty$-norm, we have
\begin{eqnarray}
\label{EqnRemBound}
\| \Res(\Delta)\|_\infty & \leq & \frac{3}{2} \degmax
\|\Delta\|_\infty^2 \; \KconCov^3.
\end{eqnarray}
\elems
We provide the proof of this lemma in Appendix~\ref{APP_LEM_R_CONV}.  It
is straightforward, based on standard matrix expansion techniques.

\subsubsection{Sufficient conditions for $\ell_\infty$ bounds}
\label{SecEllinfBound}

Our next lemma provides control on the deviation \mbox{$\Delta =
\ThetaWitness - \ThetaStar$,} measured in elementwise $\ell_\infty$
norm.
\blems[Control of $\Delta$]
\label{LEM_D_CONV}
Suppose that
\begin{eqnarray}
\label{EqnDconvAss}
r \defn 2 \KconHess \big( \|\Wnoise\|_\infty + \regpar \big) & \leq & \min
\big \{ \frac{1}{3 \KconCov \degmax}, \; \frac{1}{3 \KconCov^3
\;\KconHess \degmax} \big \}.
\end{eqnarray}
Then we have the elementwise $\ell_\infty$ bound
\begin{eqnarray}
\label{EqnDconvBound}
\|\Delta \|_\infty = \| \ThetaWit - \ThetaStar \|_\infty & \leq & r.
\end{eqnarray}
\elems

We prove the lemma in Appendix~\ref{APP_LEM_D_CONV}; at a high level,
the main steps involved are the following.  We begin by noting that
$\ThetaWitness_{\EsetPlusComp} = \ThetaStar_{\EsetPlusComp} = 0$, so
that $\vecnorm{\Delta}{\infty} =
\vecnorm{\Delta_{\EsetPlus}}{\infty}$.  Next, we characterize
$\ThetaWit_{\EsetPlus}$ in terms of the zero-gradient condition
associated with the restricted problem~\eqref{EqnRestricted}.  We then
define a continuous map $F: \Delta_{\EsetPlus} \mapsto
F(\Delta_{\EsetPlus})$ such that its fixed points are equivalent to
zeros of this gradient expression in terms of $\Delta_\EsetPlus =
\ThetaWit_\EsetPlus - \ThetaStar_{\EsetPlus}$.  We then show that the
function $F$ maps the $\ell_\infty$-ball
\begin{eqnarray}
\label{EqnDefnRad}
\Ball(\rad) & \defn & \{ \Theta_{\EsetPlus} \mid \|\Theta_\EsetPlus
\|_\infty \leq \rad \}, \qquad \mbox{with $\rad \defn 2 \KconHess \big
(\|\Wnoise\|_\infty + \regpar \big)$},
\end{eqnarray}
onto itself.  Finally, with these results in place, we can apply
Brouwer's fixed point theorem (e.g., p. 161; Ortega and
Rheinboldt~\cite{OrtegaR70}) to conclude that $F$ does indeed have a
fixed point inside $\Ball(\rad)$.  

\subsubsection{Sufficient conditions for sign consistency}
\label{SecSignConsis}

We now show how a lower bound on the minimum value $\thetamin$, when
combined with Lemma~\ref{LEM_D_CONV}, allows us to guarantee
\emph{sign consistency} of the primal witness matrix
$\ThetaWit_{\EsetPlus}$.
\blems[Sign Consistency]
\label{LemSignConsis}
Suppose the minimum absolute value $\thetamin$ of non-zero entries
in the true concentration matrix $\ThetaStar$ is lower bounded as
\begin{eqnarray}
\label{EqnThetaminLow}
\thetamin & \geq & 4 \KconHess \big (\|\Wnoise\|_\infty + \regpar
\big),
\end{eqnarray}
then \mbox{$\sign(\ThetaWit_\EsetPlus) = \sign(\ThetaStar_\EsetPlus)$}
holds.  
\elems
This claim follows from the bound~\eqref{EqnThetaminLow} combined with
the bound~\eqref{EqnDconvBound} ,which together imply that for all
$(i,j) \in \EsetPlus$, the estimate $\ThetaWit_{ij}$ cannot differ
enough from $\ThetaStar_{ij}$ to change sign.


\subsubsection{Control of noise term}
\label{SecNoise}

The final ingredient required for the proofs of Theorems~\ref{ThmMain}
and \ref{ThmModel} is control on the sampling noise $\Wnoise = \SamCov
- \CovMatStar$.  This control is specified in terms of the decay
function $\sctail$ from equation~\eqref{EqnSamTail}.
\blems[Control of Sampling Noise]
\label{LemWconv}
For any $\tunpar > 2$ and sample size $\numobs$ such that
$\sctailInvT(\numobs,\pdim^\tunpar) \le 1/\scthresh$, we have
\begin{eqnarray}
\mprob\biggr[ \|\Wnoise\|_\infty \geq
\sctailInvT(\numobs,\pdim^{\tunpar}) \biggr] & \leq &
\frac{1}{\pdim^{\tunpar - 2}} \; \rightarrow \; 0.
\end{eqnarray}

\elems
\begin{proof}
Using the definition~\eqref{EqnSamTail} of the decay function
$\sctail$, and applying the union bound over all $\pdim^2$ entries of
the noise matrix, we obtain that for all $\scdelta \leq 1/\scthresh$,
\begin{eqnarray*}
\mprob \big[\max_{i, j} |\Wnoise_{ij}| \geq \scdelta \big] & \leq &
\pdim^2/\sctail(\numobs,\scdelta).
\end{eqnarray*}
Setting $\scdelta = \sctailInvT(\numobs,\pdim^{\tunpar})$ yields that
\begin{eqnarray*}
\mprob \big[\max_{i, j} |\Wnoise_{ij}| \geq
\sctailInvT(\numobs,\pdim^{\tunpar}) \big] & \leq & \pdim^2/
\big[\sctail(\numobs,\sctailInvT(\numobs,\pdim^{\tunpar})) \big] \; =
\; 1/\pdim^{\tunpar - 2},
\end{eqnarray*}
as claimed.  Here the last equality follows since
$\sctail(\numobs,\sctailInvT(\numobs,\pdim^{\tunpar})) =
\pdim^\tunpar$, using the definition~\eqref{EqnSamTailT} of the
inverse function $\sctailInvT$.
\end{proof}

\def\scdeltaNew{\scdelta'}
\def\numobsNew{\numobs'}

\subsection{Proof of Theorem~\ref{ThmMain}}
\label{SecThmMain}

We now have the necessary ingredients to prove
Theorem~\ref{ThmMain}. We first show that with high probability the
witness matrix $\ThetaWit$ is equal to the solution $\ThetaHat$ to the
original log-determinant problem~\eqref{EqnGaussMLE}, in particular by
showing that the primal-dual witness construction (described in in
Section~\ref{SecPrimalDualWitness}) succeeds with high probability.
Let $\WitEvent$ denote the event that $\|\Wnoise\|_\infty \leq
\sctailInvT(\numobs,\pdim^{\tunpar})$.  Using the monotonicity of the
inverse tail function~\eqref{EqnMonot}, the lower lower
bound~\eqref{EqnSampleBound} on the sample size $\numobs$ implies that
$\sctailInvT(\numobs,\pdim^{\tunpar}) \le 1/\scthresh$.  Consequently,
Lemma~\ref{LemWconv} implies that $\mprob(\WitEvent) \ge 1 -
\frac{1}{\pdim^{\tunpar - 2}}$.  Accordingly, we condition on the
event $\WitEvent$ in the analysis to follow.

We proceed by verifying that assumption~\eqref{EqnDetSuff} of
Lemma~\ref{LemStrictDual} holds.  Recalling the choice of
regularization penalty \mbox{$\regpar = (8/\mutinco)\,
\sctailInvT(\numobs,\pdim^{\tunpar})$,} we have $\|\Wnoise\|_\infty
\leq (\mutinco/8) \regpar$.  In order to establish
condition~\eqref{EqnDetSuff} it remains to establish the bound
$\|\Res(\Delta)\|_\infty \leq \frac{\mutinco \,\regpar}{8}$.  We do so
in two steps, by using Lemmas~\ref{LEM_D_CONV} and~\ref{LEM_R_CONV}
consecutively.  First, we show that the
precondition~\eqref{EqnDconvAss} required for Lemma~\ref{LEM_D_CONV}
to hold is satisfied under the specified conditions on $\numobs$ and
$\regpar$. From Lemma~\ref{LemWconv} and our choice of regularization
constant \mbox{$\regpar = (8/\mutinco)\,
\sctailInvT(\numobs,\pdim^{\tunpar})$,}
\begin{eqnarray*}
2 \KconHess \big( \,\|\Wnoise\|_\infty + \regpar \big) & \leq & 2
\KconHess \Big(1 + \frac{8}{\mutinco}\Big) \,
\sctailInvT(\numobs,\pdim^{\tunpar}),
\end{eqnarray*}
provided $\sctailInvT(\numobs,\pdim^{\tunpar}) \le 1/\scthresh$.  From
the lower bound~\eqref{EqnSampleBound} and the
monotonicity~\eqref{EqnMonot} of the tail inverse functions, we have
\begin{eqnarray}
\label{EqnDelPreqBound}
2 \KconHess \Big(1 + \frac{8}{\mutinco}\Big) \,
\sctailInvT(\numobs,\pdim^{\tunpar}) & \leq & \min \big\{ \frac{1}{3
\KconCov \degmax}, \; \frac{1}{3 \KconCov^3 \;\KconHess \degmax}
\big\},
\end{eqnarray}
showing that the assumptions of Lemma~\ref{LEM_D_CONV} are satisfied.
Applying this lemma, we conclude that
\begin{eqnarray}
\label{EqnDelBound}
\| \Delta \|_\infty & \leq & 2 \KconHess \big( \, \|\Wnoise\|_\infty +
\regpar \big) \; \leq \; 2 \KconHess \Big(1 + \frac{8}{\mutinco}\Big)
\, \sctailInvT(\numobs,\pdim^{\tunpar}).
\end{eqnarray}

Turning next to Lemma~\ref{LEM_R_CONV}, we see that its assumption
$\|\Delta\|_{\infty} \leq \frac{1}{3 \, \KconCov \degmax}$ holds, by
applying equations~\eqref{EqnDelPreqBound} and \eqref{EqnDelBound}.
Consequently, we have
\begin{eqnarray*}
\|\Res(\Delta)\|_\infty & \leq & \frac{3}{2} \,\degmax\;
\|\Delta\|_\infty^2 \; \KconCov^3 \\
& \leq & 6 \KconCov^3 \KconHess^2 \, \degmax \, \Big(1 +
\frac{8}{\mutinco}\Big)^2 [\sctailInvT(\numobs,\pdim^{\tunpar})]^{2}
\\
& = & \Biggr \{ 6 \KconCov^3 \KconHess^2 \, \degmax \, \Big(1 +
\frac{8}{\mutinco}\Big)^2 \sctailInvT(\numobs,\pdim^{\tunpar}) \Biggr
\} \frac{\mutinco \regpar}{8} \\
& \leq & \frac{\mutinco \regpar}{8},
\end{eqnarray*}
as required, where the final inequality follows from our
condition~\eqref{EqnSampleBound} on the sample size, and the
monotonicity property~\eqref{EqnMonot}.  

Overall, we have shown that the assumption~\eqref{EqnDetSuff} of
Lemma~\ref{LemStrictDual} holds, allowing us to conclude that
$\ThetaWit = \ThetaHat$. The estimator $\ThetaHat$ then satisfies the
$\ell_\infty$-bound~\eqref{EqnDelBound} of $\ThetaWit$, as claimed in
Theorem~\ref{ThmMain}(a), and moreover, we have
$\ThetaHat_{\EsetPlusComp} = \ThetaWit_{\EsetPlusComp} = 0$, as
claimed in Theorem~\ref{ThmMain}(b).  Since the above was conditioned
on the event $\WitEvent$, these statements hold with probability
$\mprob(\WitEvent) \ge 1 - \frac{1}{\pdim^{\tunpar - 2}}$.

\subsection{Proof of Theorem~\ref{ThmModel}}
\label{SecThmModel}

We now turn to the proof of Theorem~\ref{ThmModel}. A little
calculation shows that the assumed lower bound~\eqref{EqnNumobsModel}
on the sample size $\numobs$ and the monotonicity
property~\eqref{EqnMonot} together guarantee that
\begin{eqnarray*}
\thetamin & > & 4 \KconHess \Big(1 +\frac{8}{\mutinco}\Big) \,
\sctailInvT(\numobs,\pdim^{\tunpar})
\end{eqnarray*}
Proceeding as in the proof of Theorem~\ref{ThmMain}, with probability
at least $1 - 1/\pdim^{\tunpar - 2}$, we have the equality
\mbox{$\ThetaWit = \ThetaHat$,} and also that $\|\ThetaWit -
\ThetaStar\|_\infty \leq \thetamin/2$.  Consequently,
Lemma~\ref{LemSignConsis} can be applied, guaranteeing that
$\sign(\ThetaStar_{ij}) = \sign(\ThetaWit_{ij})$ for all $(i,j) \in
\Eset$.  Overall, we conclude that with probability at least $1 -
1/\pdim^{\tunpar - 2}$, the sign consistency condition
$\sign(\ThetaStar_{ij}) = \sign(\ThetaHat_{ij})$ holds for all $(i,j)
\in \Eset$, as claimed.


\def\DeltaHat{\hat{\Delta}} \def\Lin{L}

\subsection{Proof of Corollary~\ref{CorCovBound}}
\label{SecCorCovProof}
With the shorthand $\DeltaHat = \ThetaHat - \ThetaStar$, we have
\begin{equation*}
\CovEstim - \CovMatStar = (\ThetaStar + \DeltaHat)^{-1} -
\inv{\ThetaStar}.
\end{equation*}
From the definition~\eqref{EqnRdefn} of the residual $R(\cdot)$, this
difference can be written as
\begin{eqnarray}
\label{EqnCovDev}
\CovEstim - \CovMatStar & = & - \invn{\ThetaStar} \DeltaHat
\invn{\ThetaStar} + R(\DeltaHat).
\end{eqnarray}

Proceeding as in the proof of Theorem~\ref{ThmMain} we condition on
the event $\WitEvent = \{ \|\Wnoise\|_\infty \leq
\sctailInvT(\numobs,\pdim^{\tunpar}) \}$, and which holds with
probability $\mprob(\WitEvent) \ge 1 - \frac{1}{\pdim^{\tunpar - 2}}$.
As in the proof of that theorem, we are guaranteed that the
assumptions of Lemma~\ref{LEM_R_CONV} are satisfied, allowing us to
conclude
\begin{align}\label{EqnRequiv}
\Res(\DeltaHat) = \invn{\opt{\conc}} \DeltaHat \invn{\opt{\conc}}
\DeltaHat J \invn{\opt{\conc}},
\end{align}
where $J \defn \sum_{k=0}^{\infty} (-1)^{k} \big(\invn{\opt{\conc}}
\DeltaHat\big)^{k}$ has norm $\matnorm{J^T}{\infty} \leq 3/2$.

We begin by proving the bound~\eqref{EqnCovInftyBound}.  From
equation~\eqref{EqnCovDev}, we have $\vecnorm{\CovEstim -
\CovMatStar}{\infty} \leq \vecnorm{L(\DeltaHat)}{\infty} +
\vecnorm{R(\DeltaHat)}{\infty}$.  From Lemma~\ref{LEM_R_CONV}, we have
the elementwise $\ell_\infty$-norm bound
\begin{eqnarray*}
\| \Res(\DeltaHat)\|_\infty & \leq & \frac{3}{2} \degmax
\|\DeltaHat\|_\infty^2 \; \KconCov^3.
\end{eqnarray*}
The quantity $L(\DeltaHat)$ in turn can be bounded as follows,
\begin{eqnarray*}
\vecnorm{L(\DeltaHat)}{\infty} & = &
\max_{ij}\big|e_{i}^{T}\invn{\ThetaStar} \DeltaHat \invn{\ThetaStar}
e_{j}\big|\\ &\le& \max_{i} \|e_{i}^{T}\invn{\ThetaStar}\|_{1}
\max_{j} \|\DeltaHat \invn{\ThetaStar} e_j\|_{\infty}\\ &\le& \max_{i}
\|e_{i}^{T}\invn{\ThetaStar}\|_{1} \|\DeltaHat\|_{\infty} \|\max_{j}
\|\invn{\ThetaStar} e_j\|_{1}
\end{eqnarray*}
where we used the inequality that $\|\DeltaHat u \|_{\infty} \le
\|\DeltaHat\|_{\infty} \|u\|_{1}$.  Simplifying further, we obtain
\begin{eqnarray*}
\vecnorm{L(\DeltaHat)}{\infty} & \leq &
\matnorm{\invn{\ThetaStar}}{\infty} \|\DeltaHat\|_{\infty}
\matnorm{\invn{\ThetaStar}}{1}\\ 
& \leq & \matnorm{\invn{\ThetaStar}}{\infty}^{2}
\|\DeltaHat\|_{\infty} \\ 
& \leq & \KconCov^{2} \|\DeltaHat\|_{\infty},
\end{eqnarray*}
where we have used the fact that $\matnorm{\invn{\ThetaStar}}{1} =
\matnorm{[\invn{\ThetaStar}]^{T}}{\infty} =
\matnorm{\invn{\ThetaStar}}{\infty}$, which follows from the symmetry
of $\invn{\ThetaStar}$.  Combining the pieces, we obtain
\begin{eqnarray}
\label{EqnCovDevInftya}
\vecnorm{\CovEstim - \CovMatStar}{\infty} &\le &
\vecnorm{L(\DeltaHat)}{\infty} + \vecnorm{R(\DeltaHat)}{\infty}\\
& \leq & \KconCov^{2} \|\DeltaHat\|_{\infty} + \frac{3}{2} \degmax
\KconCov^3 \|\DeltaHat\|_\infty^2. \nonumber
\end{eqnarray}
The claim then follows from the elementwise $\ell_\infty$-norm
bound~\eqref{EqnEllinfBound} from Theorem~\ref{ThmMain}.

Next, we establish the bound~\eqref{EqnCovSpectralBound} in spectral
norm.  Taking the $\ell_{\infty}$ operator norm of both sides of
equation~\eqref{EqnCovDev} yields the inequality $\matnorm{\CovEstim -
\CovMatStar}{\infty} \leq \matnorm{L(\DeltaHat)}{\infty} +
\matnorm{R(\DeltaHat)}{\infty}$. Using the
expansion~\eqref{EqnRequiv}, and the sub-multiplicativity of the
$\ell_{\infty}$ operator norm, we obtain
\begin{eqnarray*}
\matnorm{\Res(\DeltaHat)}{\infty} & \leq &
\matnorm{\invn{\opt{\conc}}}{\infty} \matnorm{\DeltaHat}{\infty}
\matnorm{\invn{\opt{\conc}}}{\infty} \matnorm{\DeltaHat}{\infty}
\matnorm{J}{\infty} \matnorm{\invn{\opt{\conc}}}{\infty}\\ &\le&
\matnorm{\invn{\opt{\conc}}}{\infty}^{3} \matnorm{J}{\infty}
\matnorm{\DeltaHat}{\infty}^{2}\\ 
& \leq & \frac{3}{2} \KconCov^{3} \matnorm{\DeltaHat}{\infty}^{2},
\end{eqnarray*}
where the last inequality uses the bound $\matnorm{J}{\infty} \leq
3/2$.  (Proceeding as in the proof of Lemma~\ref{LEM_R_CONV}, this
bound holds conditioned on $\WitEvent$, and for the sample size
specified in the theorem statement.)  In turn, the term $L(\DeltaHat)$
can be bounded as
\begin{eqnarray*}
\matnorm{L(\Delta)}{\infty} & \leq & \matnorm{\invn{\ThetaStar}
\DeltaHat \invn{\ThetaStar}}{\infty}\\ &\le&
\matnorm{\invn{\ThetaStar}}{\infty}^{2} \matnorm{\DeltaHat}{\infty}\\
& \leq & \KconCov^{2} \matnorm{\DeltaHat}{\infty},
\end{eqnarray*}
where the second inequality uses the sub-multiplicativity of the
$\ell_\infty$-operator norm.  Combining the pieces yields
\begin{eqnarray}
\label{EqnCovDevInftyOp}
\matnorm{\CovEstim - \CovMatStar}{\infty} & \leq &
\matnorm{L(\DeltaHat)}{\infty} + \matnorm{R(\DeltaHat)}{\infty} \;
\leq \; \KconCov^{2} \matnorm{\DeltaHat}{\infty} + \frac{3}{2} \KconCov^3
\|\DeltaHat\|_\infty^2.
\end{eqnarray}
Conditioned on the event $\WitEvent$, we have the
bound~\eqref{EqnPrecInftyOp} on the $\ell_\infty$-operator norm
\begin{eqnarray*}
\matnorm{\DeltaHat}{\infty} & \leq & 2 \KconHess \Big(1 +
\frac{8}{\mutinco}\Big) \, \degmax \,
\sctailInvT(\numobs,\pdim^{\tunpar}).
\end{eqnarray*}
Substituting this bound, as well as the elementwise $\ell_\infty$-norm
bound~\eqref{EqnEllinfBound} from Theorem~\ref{ThmMain}, into the
bound~\eqref{EqnCovDevInftyOp} yields the stated claim.


\section{Experiments}
\label{SecExperiments}

In this section, we illustrate our results with various experimental
simulations, reporting results in terms of the probability of correct
model selection (Theorem~\ref{ThmModel}) or the $\ell_\infty$-error
(Theorem~\ref{ThmMain}).  For these illustrations, we study the case
of Gaussian graphical models, and results for three different classes
of graphs, namely chains, grids, and star-shaped graphs. We also
consider various scalings of the quantities which affect the
performance of the estimator: in addition the triple $(\numobs, \pdim,
\degmax)$, we also report some results concerning the role of the
parameters $\KconCov$, $\KconHess$ and $\thetamin$ that we have identified in the
main theorems. For all results
reported here, we solved the resulting $\ell_1$-penalized
log-determinant program~\eqref{EqnGaussMLE} using the \texttt{glasso}
program of \citet{FriedHasTib2007}, which builds on the block
co-ordinate descent algorithm of~\citet{AspreBanG2008}.

\begin{figure}
\begin{center}
\begin{tabular}{ccccc}
\raisebox{.6in}{\widgraph{0.25\textwidth}{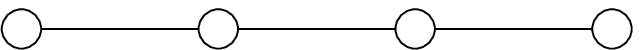}}& &
\raisebox{.0in}{\widgraph{0.28\textwidth}{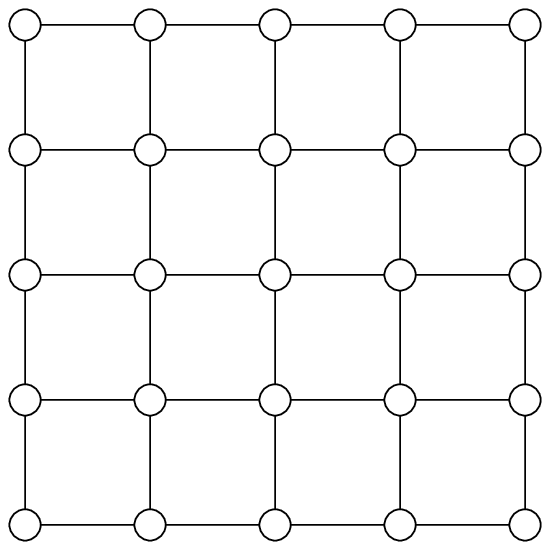}}& &
\raisebox{.1in}{\widgraph{0.25\textwidth}{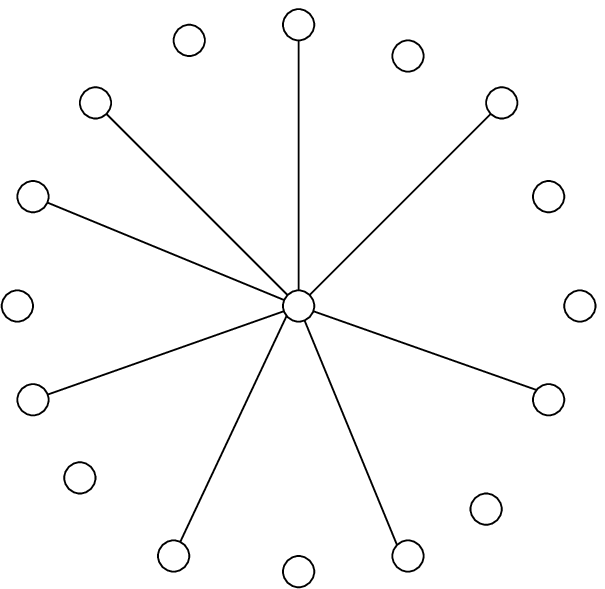}}\\
(a) && (b) && (c)
\end{tabular}
\caption{Illustrations of different graph classes used in simulations.
(a) Chain ($\degmax = 2$).  (b) Four-nearest neighbor grid ($\degmax =
4$) and (c) Star-shaped graph ($\degmax \in \{1,\hdots,\pdim - 1\}$).}
\label{FigGraphs}
\end{center}
\end{figure}

Figure~\ref{FigGraphs} illustrates the three types of graphs used in
our simulations: chain graphs (panel (a)), four-nearest neighbor
lattices or grids (panel (b)), and star-shaped graphs (panel (c)).
For the chain and grid graphs, the maximal node degree $\degmax$ is
fixed by definition, to $\degmax =2$ for chains, and $\degmax =4$ for
the grids.  Consequently, these graphs can capture the dependence of
the required sample size $\numobs$ only as a function of the graph
size $\pdim$, and the parameters $(\KconCov$, $\KconHess$,
$\thetamin$).  The star graph allows us to vary both $\degmax$ and
$\pdim$, since the degree of the central hub can be varied between $1$
and $\pdim -1$.  For each graph type, we varied the size of the graph
$\pdim$ in different ranges, from $\pdim =64$ upwards to $\pdim =
375$.

For the chain and star graphs, we define a covariance matrix
$\CovMatStar$ with entries $\CovMatStar_{ii} = 1$ for all $i =1,
\ldots, \pdim$, and $\CovMatStar_{ij} = \rho$ for all $(i,j) \in
\edge$ for specific values of $\rho$ specified below.  Note that these
covariance matrices are sufficient to specify the full model.  For the
four-nearest neighbor grid graph, we set the entries of the
concentration matrix $\ThetaStar_{ij} = \omega$ for $(i,j) \in \edge$,
with the value $\omega$ specified below. In all cases, we set the
regularization parameter $\regpar$ proportional to
$\sqrt{\log(\pdim)/\numobs}$, as suggested by Theorems~\ref{ThmMain}
and~\ref{ThmModel}, which is reasonable since the main purpose of
these simulations is to illustrate our theoretical results.  However,
for general data sets, the relevant theoretical parameters cannot be
computed (since the true model is unknown), so that a data-driven
approach such as cross-validation might be required for selecting the
regularization parameter $\regpar$.

Given a Gaussian graphical model instance, and the number of samples
$\numobs$, we drew $N = 100$ batches of $\numobs$ independent samples
from the associated multivariate Gaussian distribution. We estimated
the probability of correct model selection as the fraction of the $N=
100$ trials in which the estimator recovers the signed-edge set
exactly.

\def\Kcomplex{K}
Note that any multivariate Gaussian random vector is sub-Gaussian; in
particular, the rescaled variates $X_i/\sqrt{\CovMatStar_{ii}}$ are
sub-Gaussian with parameter $\csubg = 1$, so that the elementwise
$\ell_\infty$-bound from Corollary~\ref{CorEllinfSubg} applies.
Suppose we collect relevant parameters such as $\thetamin$ and the
covariance and Hessian-related terms $\KconCov$, $\KconHess$ and
$\mutinco$ into a single ``model-complexity'' term $\Kcomplex$ defined
as
\begin{eqnarray}\label{EqnKdefn}
K & \defn & \left[(1 + 8 \mutinco^{-1}) (\max_{i}\CovMatStar_{ii})
\max\{\KconCov \KconHess, \KconCov^{3} \KconHess^{2},\frac{\KconHess}{
\degmax\, \thetamin}\}\right].
\end{eqnarray}
Then, as a corollary of Theorem~\ref{ThmModel}, a sample size of order 
\begin{eqnarray}
\label{EqnCrudeBound}
\numobs & = & \Omega\left( K^{2} \; \degmax^2 \, \tunpar \log \pdim
\right),
\end{eqnarray}
is sufficient for model selection consistency with probability greater
than $1-1/\pdim^{\tunpar-2}$. In the subsections to follow, we
investigate how the empirical sample size $\numobs$ required for model
selection consistency scales in terms of graph size $\pdim$, maximum
degree $\degmax$, as well as the ``model-complexity'' term $K$ defined
above.
 
\newcommand{\simfigsize}{.48\textwidth}
\begin{figure}[h]
\begin{center}
\begin{tabular}{ccc}
\widgraph{\simfigsize}{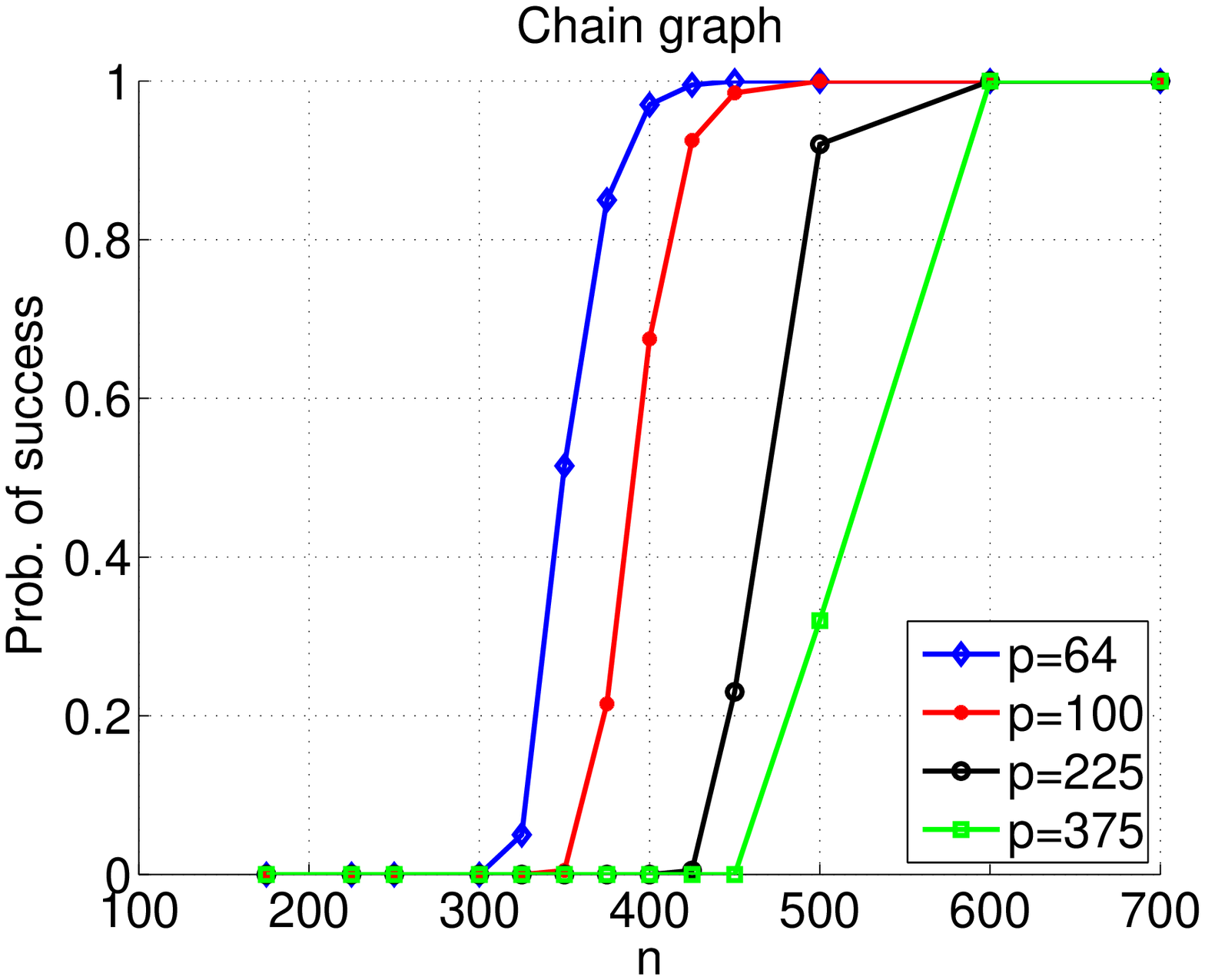} & \hspace*{.1in} &
\widgraph{\simfigsize}{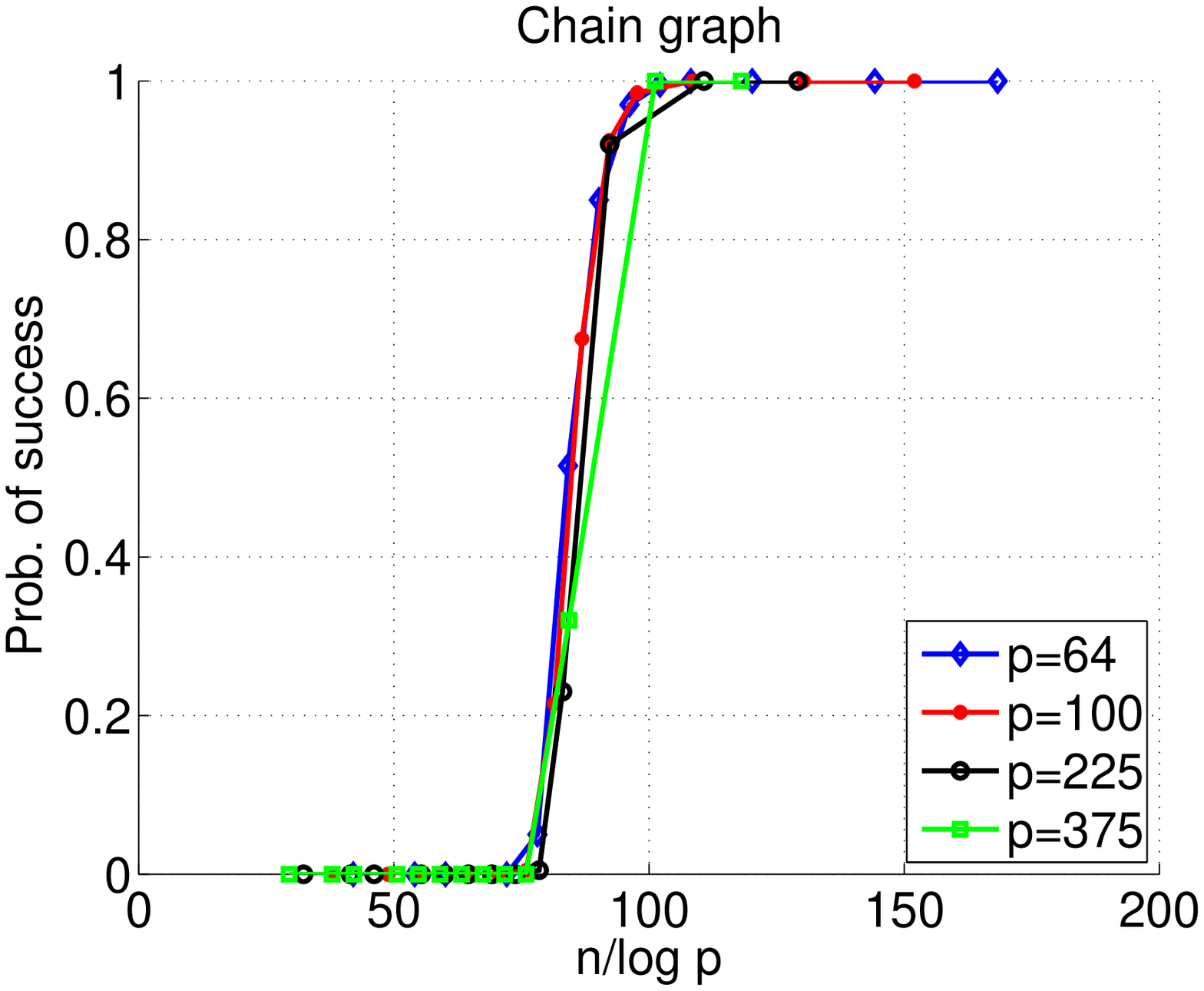} \\
(a) & & (b)
\end{tabular}
\end{center}
\caption{Simulations for chain graphs with varying number of nodes
$\pdim$, edge covariances $\CovMatStar_{ij} = 0.10$. Plots of
probability of correct signed edge-set recovery plotted versus the
ordinary sample size $\numobs$ in panel (a), and versus the rescaled
sample size $\numobs/\log \pdim$ in panel (b). Each point corresponds
to the average over $100$ trials. }
\label{FigChainProbvsNP}
\end{figure}

\subsection{Dependence on graph size}

Panel (a) of Figure~\ref{FigChainProbvsNP} plots the probability of
correct signed edge-set recovery against the sample size $\numobs$ for
a chain-structured graph of three different sizes.  For these chain
graphs, regardless of the number of nodes $\pdim$, the maximum node
degree is constant $\degmax = 2$, while the edge covariances are set
as $\CovMat_{ij} = 0.2$ for all $(i,j) \in \edge$, so that the
quantities $(\KconCov, \KconHess, \mutinco)$ remain constant.  Each of
the curve in panel (a) corresponds to a different graph size $\pdim$.
For each curve, the probability of success starts at zero (for small
sample sizes $\numobs$), but then transitions to one as the sample
size is increased.  As would be expected, it is more difficult to
perform model selection for larger graph sizes, so that (for instance)
the curve for $\pdim = 375$ is shifted to the right relative to the
curve for $\pdim = 64$.  Panel (b) of Figure~\ref{FigChainProbvsNP}
replots the same data, with the horizontal axis rescaled by $(1/\log
\pdim)$.  This scaling was chosen because for sub-Gaussian tails, our
theory predicts that the sample size should scale logarithmically with
$\pdim$ (see equation~\eqref{EqnCrudeBound}).  Consistent with this
prediction, when plotted against the rescaled sample size
$\numobs/\log \pdim$, the curves in panel (b) all stack up.
Consequently, the ratio $(\numobs/\log\pdim)$ acts as an effective
sample size in controlling the success of model selection, consistent
with the predictions of Theorem~\ref{ThmModel} for sub-Gaussian
variables.

\begin{figure}
\begin{center}
\begin{tabular}{ccc}
\widgraph{\simfigsize}{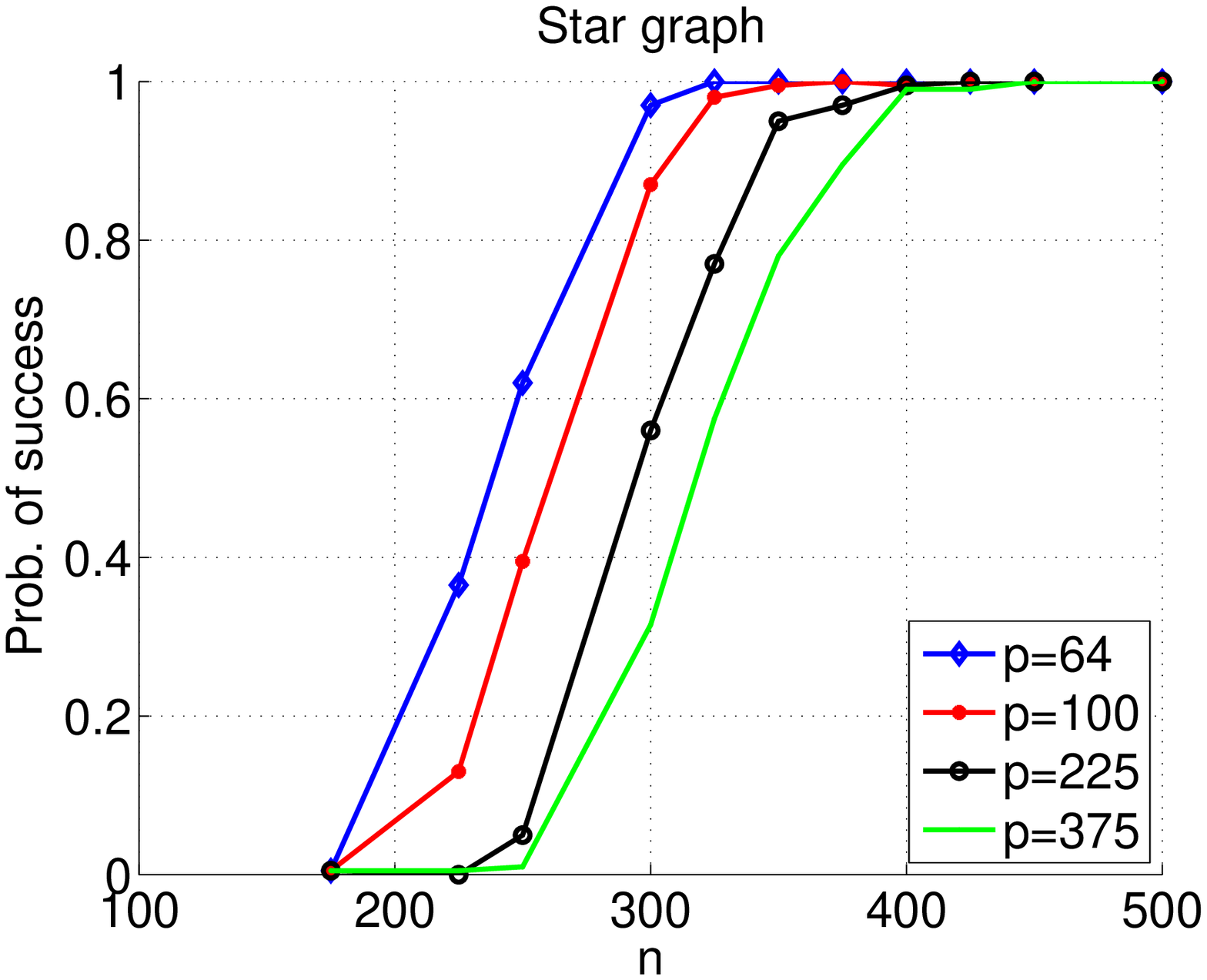} & \hspace*{.1in} &
\widgraph{\simfigsize}{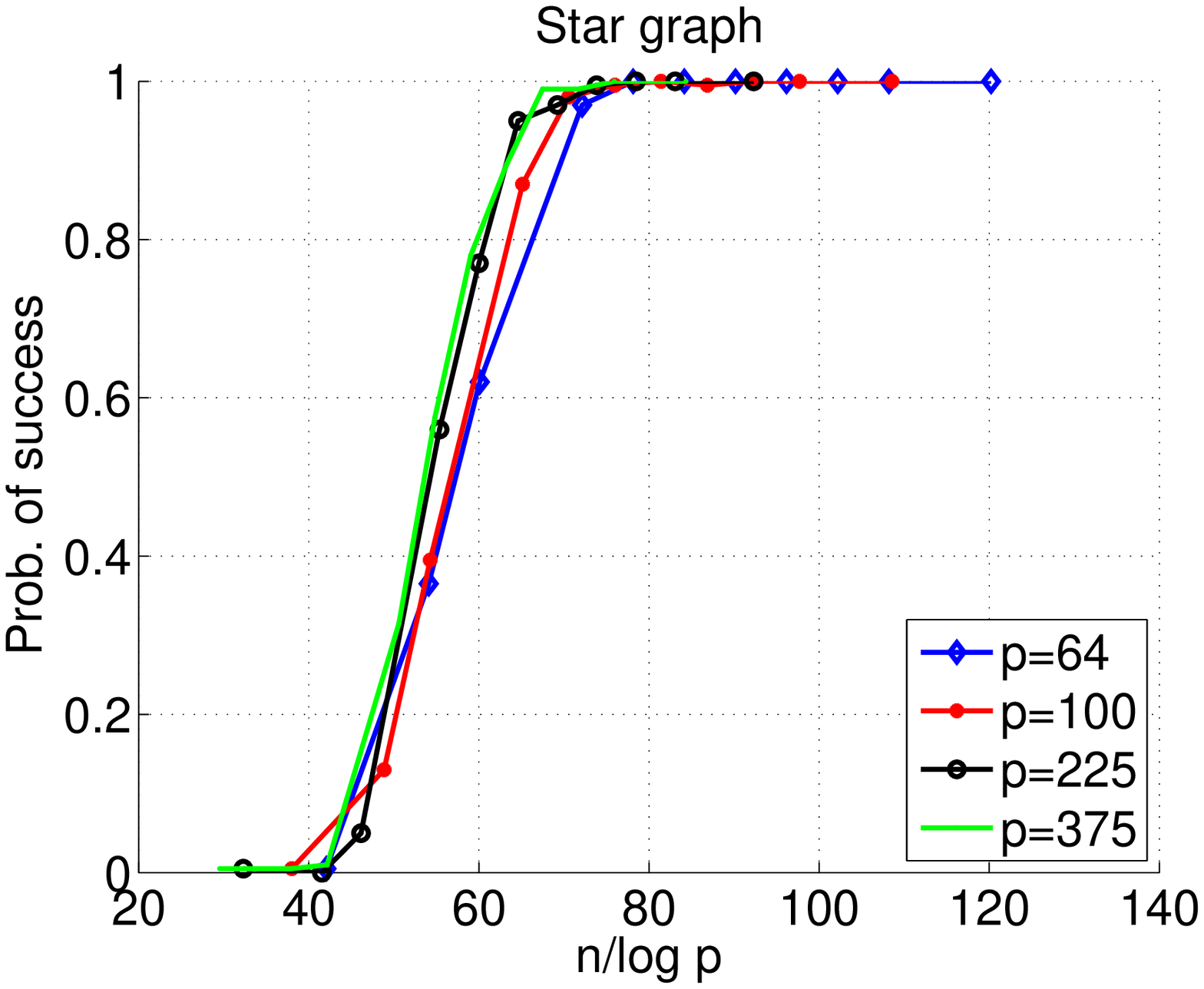} \\
(a) & & (b)
\end{tabular}
\end{center}
\caption{Simulations for a star graph with varying number of nodes
$\pdim$, fixed maximal degree $\degmax = 40$, and edge covariances
$\CovMatStar_{ij} = 1/16$ for all edges. Plots of probability of
correct signed edge-set recovery versus the sample size $\numobs$ in
panel (a), and versus the rescaled sample size $\numobs/\log \pdim$ in
panel (b).  Each point corresponds to the average over $N = 100$
trials.}
\label{FigStarProbvsNP}
\end{figure}

\begin{figure}
\begin{center}
\begin{tabular}{cc}
\widgraph{\simfigsize}{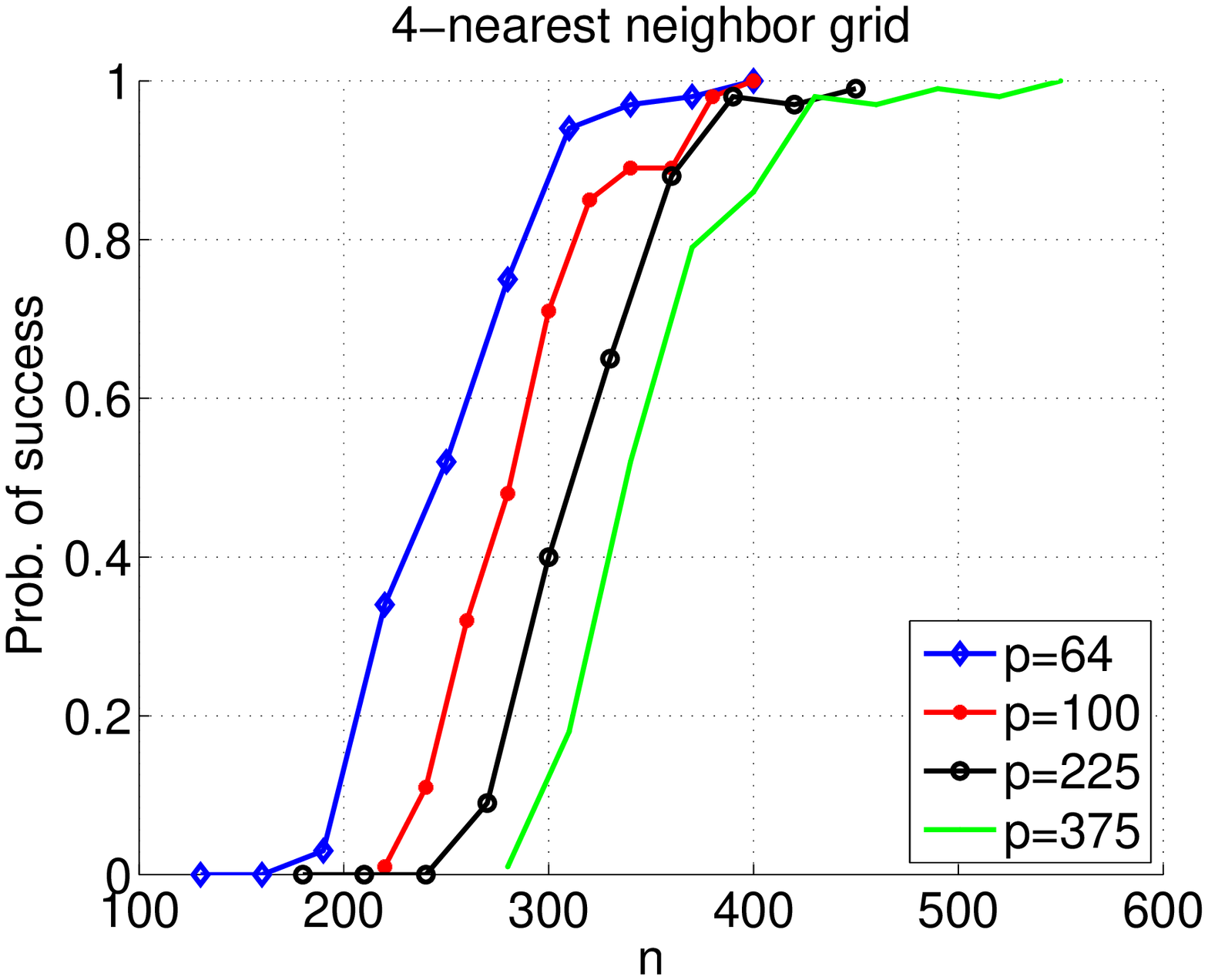} &
\widgraph{\simfigsize}{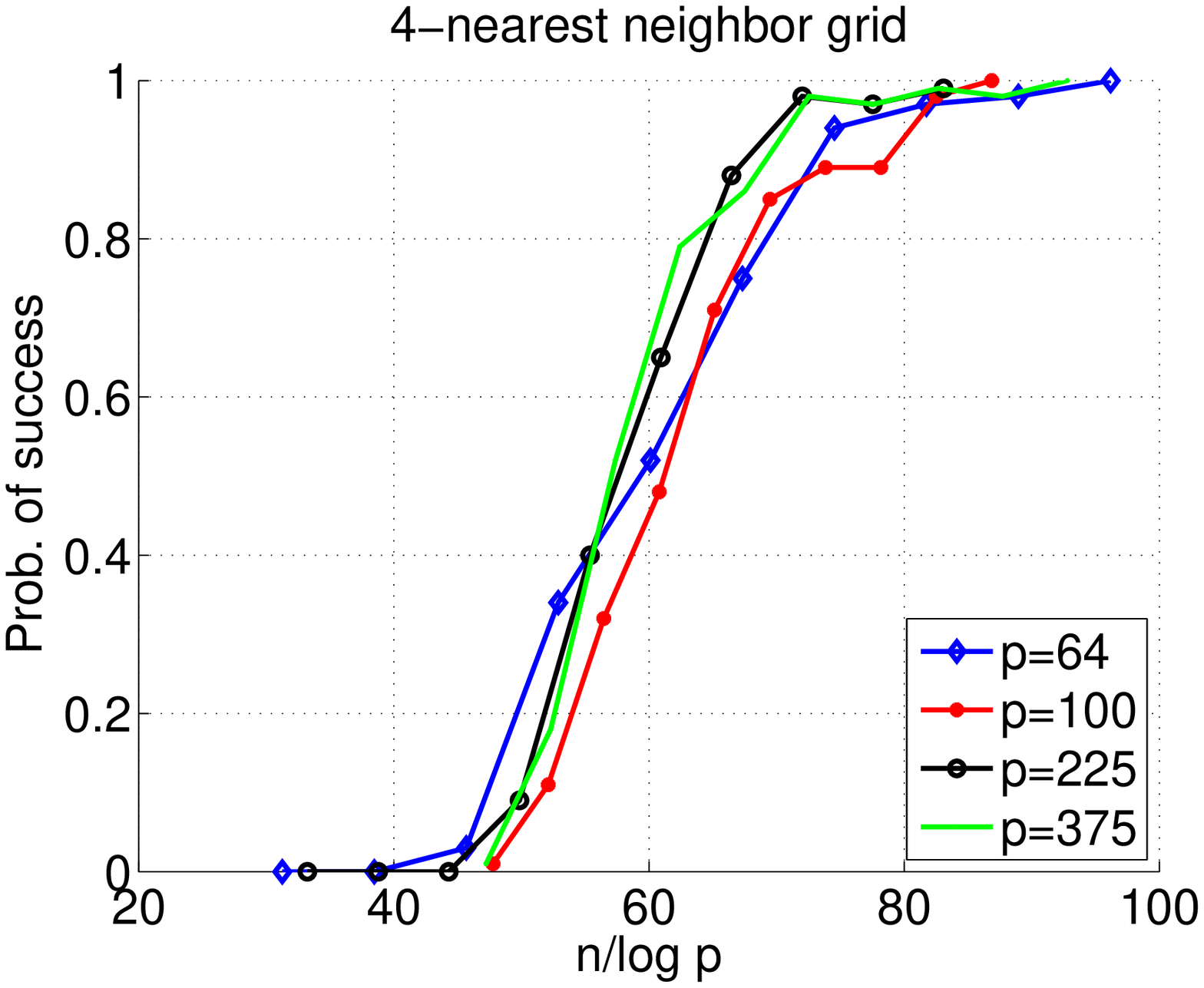} \\
(a) & (b)
\end{tabular}
\end{center}
\caption{Simulations for $2$-dimensional lattice with
$4$-nearest-neighbor interaction, edge strength interactions
$\ThetaStar_{ij} = 0.1$, and a varying number of nodes $\pdim$. Plots
of probability of correct signed edge-set recovery versus the sample
size $\numobs$ in panel (a), and versus the rescaled sample size
$\numobs/\log \pdim$ in panel (b).  Each point corresponds to the
average over $N = 100$ trials.}
\label{FigGridProbvsNP}
\end{figure}

Figure~\ref{FigStarProbvsNP} shows the same types of plots for a
star-shaped graph with fixed maximum node degree $\degmax = 40$, and
Figure~\ref{FigGridProbvsNP} shows the analogous plots for a grid
graph with fixed degree $\degmax = 4$.  As in the chain case, these
plots show the same type of stacking effect in terms of the scaled
sample size $\numobs/\log \pdim$, when the degree $\degmax$ and other
parameters ($(\mutinco, \KconHess, \KconCov)$) are held fixed.

\begin{figure}
\begin{center}
\begin{tabular}{cc}
\widgraph{\simfigsize}{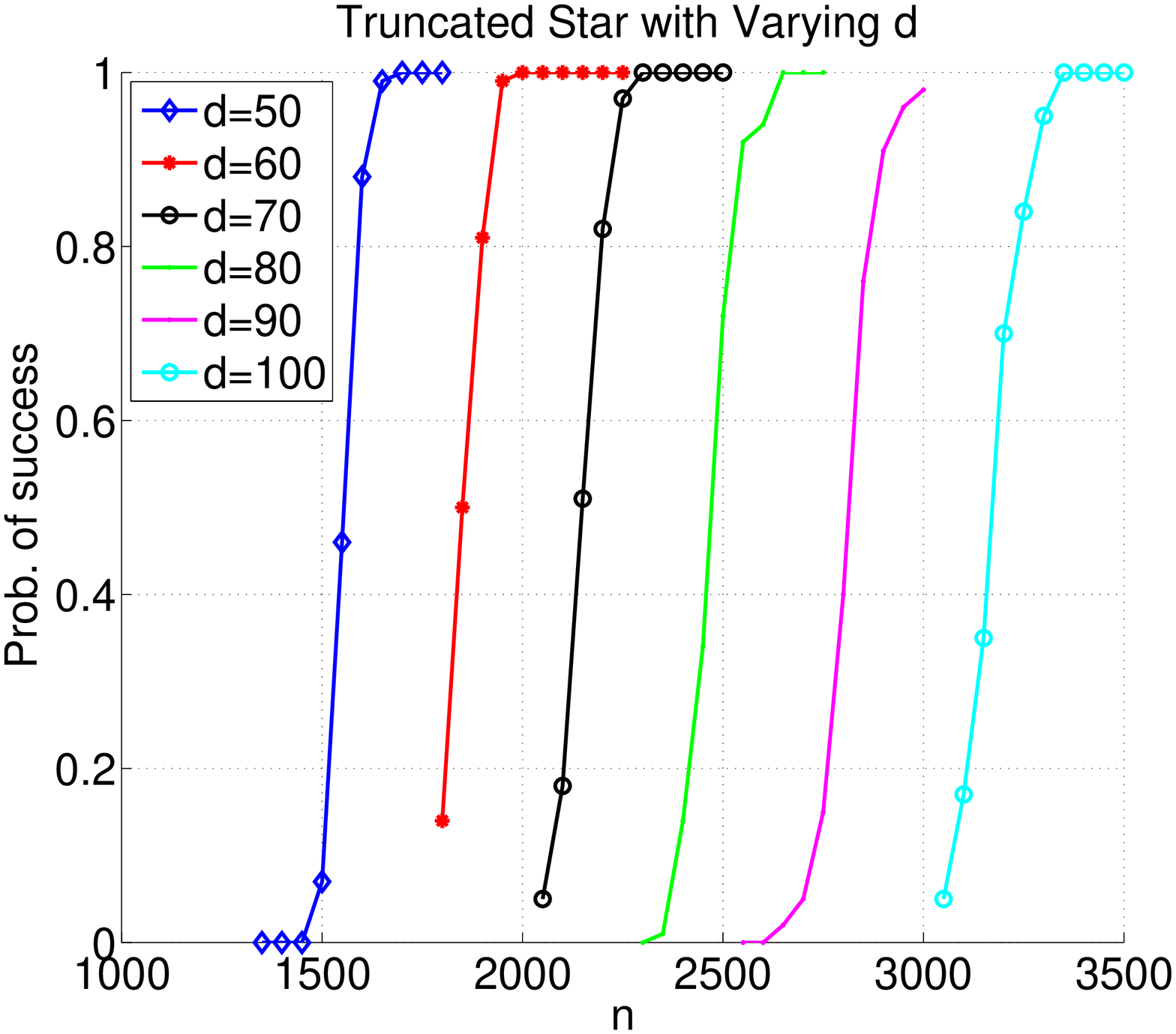} &
\widgraph{\simfigsize}{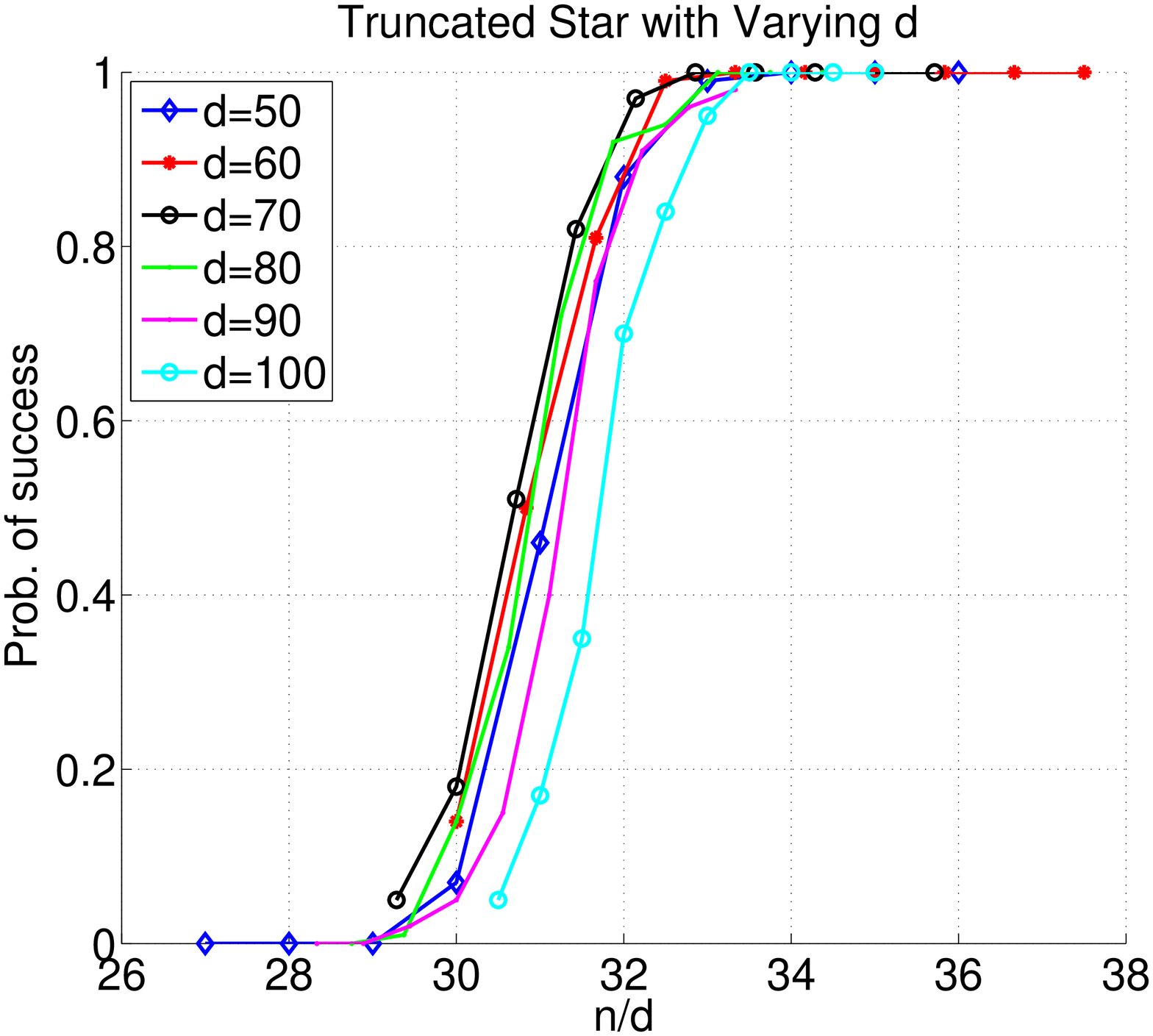} \\
(a) & (b)
\end{tabular}
\end{center}
\caption{ Simulations for star graphs with fixed number of nodes
$\pdim = 200$, varying maximal (hub) degree $\degmax$, edge
covariances $\CovMatStar_{ij} = 2.5/\degmax$.  Plots of probability of
correct signed edge-set recovery versus the sample size $\numobs$ in
panel (a), and versus the rescaled sample size $\numobs/\degmax$ in
panel (b). }
\label{FigStarProbvsND}
\end{figure}

\subsection{Dependence on the maximum node degree} 

Panel (a) of Figure~\ref{FigStarProbvsND} plots the probability of
correct signed edge-set recovery against the sample size $\numobs$ for
star-shaped graphs; each curve corresponds to a different choice of
maximum node degree $\degmax$, allowing us to investigate the
dependence of the sample size on this parameter.  So as to control
these comparisons, the models are chosen such that quantities other
than the maximum node-degree $\degmax$ are fixed: in particular, we
fix the number of nodes $\pdim = 200$, and the edge covariance entries
are set as $\CovMatStar_{ij} = 2.5/\degmax$ for $(i,j) \in \edge$ so
that the quantities $(\KconCov, \KconHess, \mutinco)$ remain constant.
The minimum value $\thetamin$ in turn scales as $1/\degmax$.  Observe
how the plots in panel (a) shift to the right as the maximum node
degree $\degmax$ is increased, showing that star-shaped graphs with
higher degrees are more difficult.  In panel (b) of
Figure~\ref{FigStarProbvsND}, we plot the same data versus the
rescaled sample size $\numobs/\degmax$.  Recall that if all the curves
were to stack up under this rescaling, then it means the required
sample size $\numobs$ scales linearly with $\degmax$.  These plots are
closer to aligning than the unrescaled plots, but the agreement is not
perfect.  In particular, observe that the curve $\degmax$ (right-most
in panel (a)) remains a bit to the right in panel (b), which suggests
that a somewhat more aggressive rescaling---perhaps
$\numobs/\degmax^\gamma$ for some $\gamma \in (1,2)$---is appropriate.

Note that for $\thetamin$ scaling as $1/\degmax$, the sufficient
condition from Theorem~\ref{ThmModel}, as summarized in
equation~\eqref{EqnCrudeBound}, is $\numobs = \Omega(\degmax^2 \log
\pdim)$, which appears to be overly conservative based on these data.
Thus, it might be possible to tighten our theory under certain
regimes.

\subsection{Dependence on covariance and Hessian terms}
\newcommand{\ModCom}{\ensuremath{T}}

Next, we study the dependence of the sample size required for model
selection consistency on the model complexity term $\Kcomplex$ defined
in \eqref{EqnKdefn}, which is a collection of the quantities
$\KconCov$, $\KconHess$ and $\mutinco$ defined by the covariance
matrix and Hessian, as well as the minimum value
$\thetamin$. Figure~\ref{FigChainNstarvsGamma} plots the probability
of correct signed edge-set recovery versus the sample size $\numobs$
for chain graphs.  Here each curve corresponds to a different setting
of the model complexity factor $\modelCompFac$, but with a fixed
number of nodes $\pdim = 120$, and maximum node-degree $\degmax = 2$.
We varied the actor $\modelCompFac$ by varying the value $\rho$ of the
edge covariances $\CovMat_{ij} = \rho,\, (i,j) \in \edge$.  Notice how
the curves, each of which corresponds to a different model complexity
factor, shift rightwards as $\modelCompFac$ is increased so that
models with larger values of $\modelCompFac$ require greater number of
samples $\numobs$ to achieve the same probability of correct model
selection.  These rightward-shifts are in qualitative agreement with
the prediction of Theorem~\ref{ThmMain}, but we suspect that our
analysis is not sharp enough to make accurate quantitative predictions
regarding this scaling.

\newcommand{\solofigsize}{.5\textwidth}

\begin{figure}
\begin{center}
\widgraph{\solofigsize}{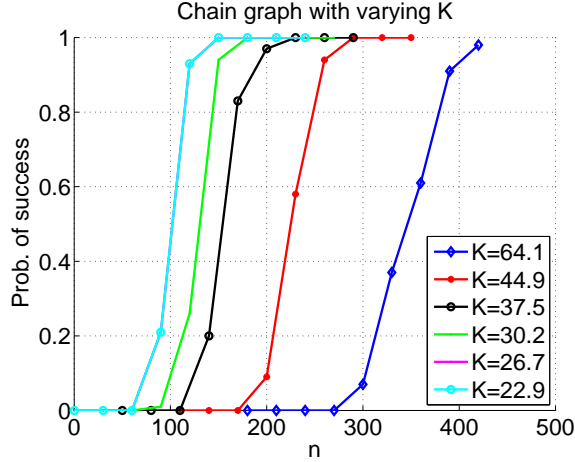} \\
\end{center}
\caption{Simulations for chain graph with fixed number of nodes $\pdim
= 120$, and varying model complexity $\modelCompFac$. Plot of
probability of correct signed edge-set recovery versus the sample size
$\numobs$.}
\label{FigChainNstarvsGamma}
\end{figure}

\subsection{Convergence rates in elementwise $\ell_\infty$-norm}

\begin{figure}
\begin{center}
\widgraph{\solofigsize}{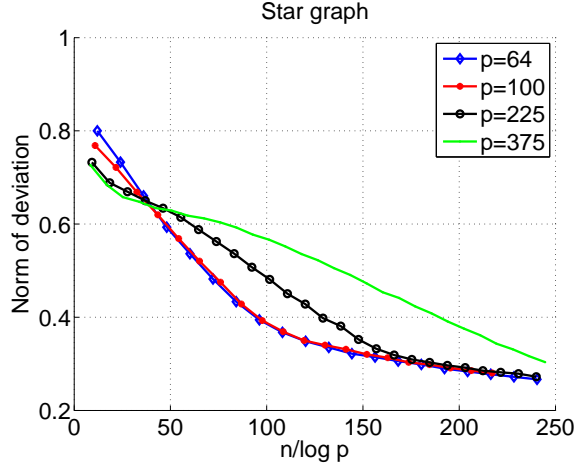} \\
\end{center}
\caption{Simulations for a star graph with varying number of nodes
$\pdim$, maximum node degree $\degmax = \lceil 0.1 \pdim \rceil$, edge
covariances $\CovMatStar_{ij}= 2.5/\degmax$. Plot of the element-wise
$\ell_\infty$ norm of the concentration matrix estimate error
$\vecnorm{\ThetaHat - \ThetaStar}{\infty}$ versus the rescaled sample
size $\numobs/\log (\pdim)$.}
\label{FigStarNormvsBaserate}
\end{figure}

Finally, we report some simulation results on the convergence rate in
elementwise $\ell_\infty$-norm.  According to
Corollary~\ref{CorEllinfSubg}, in the case of sub-Gaussian tails, if
the elementwise $\ell_\infty$-norm should decay at rate
$\order(\sqrt{\frac{\log \pdim}{\numobs}})$, once the sample size
$\numobs$ is sufficiently large.  Figure~\ref{FigStarNormvsBaserate}
shows the behavior of the elementwise $\ell_\infty$-norm for
star-shaped graphs of varying sizes $\pdim$.  The results reported
here correspond to the maximum degree $\degmax = \lceil 0.1 \pdim
\rceil$; we also performed analogous experiments for $\degmax =
\order(\log \pdim)$ and $\degmax = \order(1)$, and observed
qualitatively similar behavior.  The edge correlations were set as
$\CovMatStar_{ij} = 2.5/\degmax$ for all $(i,j) \in \edge$ so that the
quantities $(\KconCov, \KconHess, \mutinco)$ remain constant.  With
these settings, each curve in Figure~\ref{FigStarNormvsBaserate}
corresponds to a different problem size, and plots the elementwise
$\ell_\infty$-error versus the rescaled sample size $\numobs/\log
\pdim$, so that we expect to see curves of the form $f(t) =
1/\sqrt{t}$.  The curves show that when the rescaled sample size
$(\numobs/\log \pdim)$ is larger than some threshold (roughly $40$ in
the plots shown), the elementwise $\ell_\infty$ norm decays at the
rate $\sqrt{\frac{\log \pdim}{\numobs}}$, which is consistent with
Corollary~\ref{CorEllinfSubg}.

\section{Discussion}

The focus of this paper is the analysis of the high-dimensional
scaling of the $\ell_1$-regularized log determinant
problem~\eqref{EqnGaussMLE} as an estimator of the concentration
matrix of a random vector.  Our main contributions were to derive
sufficient conditions for its model selection consistency as well as
convergence rates in both elementwise $\ell_\infty$-norm, as well as
Frobenius and spectral norms.  Our results allow for a range of tail
behavior, ranging from the exponential-type decay provided by Gaussian
random vectors (and sub-Gaussian more generally), to polynomial-type
decay guaranteed by moment conditions.  In the Gaussian case, our
results have natural interpretations in terms of Gaussian Markov
random fields.

Our main results relate the i.i.d. sample size $\numobs$ to various
parameters of the problem required to achieve consistency.  In
addition to the dependence on matrix size $\pdim$, number of edges
$\spindex$ and graph degree $\degmax$, our analysis also illustrates
the role of other quantities, related to the structure of the
covariance matrix $\CovMatStar$ and the Hessian of the objective
function, that have an influence on consistency rates.  Our main
assumption is an irrepresentability or mutual incoherence condition,
similar to that required for model selection consistency of the Lasso,
but involving the Hessian of the log-determinant objective
function~\eqref{EqnGaussMLE}, evaluated at the true model. When the
distribution of $X$ is multivariate Gaussian, this Hessian is the
Fisher information matrix of the model, and thus can be viewed as an
edge-based counterpart to the usual node-based covariance matrix We
report some examples where irrepresentability condition for the Lasso
hold and the log-determinant condition fails, but we do not know in
general if one requirement dominates the other.  In addition to these
theoretical results, we provided a number of simulation studies
showing how the sample size required for consistency scales with
problem size, node degrees, and the other complexity parameters
identified in our analysis.

There are various interesting questions and possible extensions to
this paper.  First, in the current paper, we have only derived
sufficient conditions for model selection consistency. As in past work
on the Lasso~\cite{Wainwright2006_new}, it would also be interesting
to derive a \emph{converse result}---namely, to prove that if the
sample size $\numobs$ is smaller than some function of $(\pdim,
\degmax, \spindex)$ and other complexity parameters, then regardless
of the choice of regularization constant, the log-determinant method
fails to recover the correct graph structure.  Second, while this
paper studies the problem of estimating a fixed graph or concentration
matrix, a natural extension would allow the graph to vary over time, a
problem setting which includes the case where the observations are
dependent. For instance, \citet{ZhoLafWas08} study the estimation of
the covariance matrix of a Gaussian distribution in a time-varying
setting, and it would be interesting to extend results of this paper
to this more general setting.

\subsection*{Acknowledgements}  We thank Shuheng Zhou for helpful comments
on an earlier draft of this work.  Work was partially supported by NSF
grant DMS-0605165.  Yu also acknowledges support from ARO
W911NF-05-1-0104, NSFC-60628102, and a grant from MSRA.

\appendix

\section{Proof of Lemma~\ref{LEM_MLE_CHARAC}}
\label{AppLemMLECharac}

In this appendix, we show that the regularized log-determinant
program~\eqref{EqnGaussMLE} has a unique solution whenever $\regpar >
0$, and the diagonal of the sample covariance $\SigHat^\numobs$ is
strictly positive.  By the strict convexity of the log-determinant
barrier~\citep{Boyd02}, if the minimum is attained, then it is unique,
so that it remains to show that the minimum is achieved.  If $\regpar
> 0$, then by Lagrangian duality, the problem can be written in an
equivalent constrained form:
\begin{equation}
\label{EqnConForm}
\min_{\Theta \succ 0, \ellreg{\Theta} \leq C(\regpar) } \big \{
\tracer{\Theta}{\SigHat^\numobs} - \log \det(\Theta) \big \}
\end{equation}
for some $C(\regpar) < +\infty$.  Since the off-diagonal elements
remain bounded within the $\ell_1$-ball, the only possible issue is
the behavior of the objective function for sequences with possibly
unbounded diagonal entries. Since any $\Theta$ in the constraint set is
positive-definite, its diagonal entries are positive, and hence bounded below by zero.
Further, by Hadamard's inequality for positive
definite matrices~\cite{Horn1985}, we have $\log \det \Theta \leq
\sum_{i=1}^\pdim \log \Theta_{ii}$, so that
\begin{eqnarray*}
\sum_{i=1}^\pdim \Theta_{ii} \SigHat^\numobs_{ii} - \log \det \Theta &
\geq & \sum_{i=1}^\pdim \big \{\Theta_{ii} \SigHat^\numobs_{ii} - \log
\Theta_{ii} \}.
\end{eqnarray*}
As long as $\SigHat^\numobs_{ii} > 0$ for each $i = 1, \ldots, \pdim$,
this function is coercive, meaning that it diverges to infinity for
any sequence $\|(\Theta^t_{11}, \ldots, \Theta^t_{\pdim \pdim})\|
\rightarrow +\infty$.  Consequently, the minimum is attained.

Returning to the penalized form~\eqref{EqnGaussMLE}, by standard
optimality conditions for convex programs, a matrix $\SigHat \succ 0$
is optimal if and only $0$ belongs to the sub-differential of the
objective, or equivalently if and only if there exists a matrix
$\Zhat$ in the sub-differential of the off-diagonal norm
$\ellreg{\cdot}$ such that
\begin{align*}
\SigHat - \invn{\estim{\Theta}} + \lambda \Zhat = 0,
\end{align*}
as claimed.


\section{Proof of Lemma~\ref{LEM_R_CONV}}
\label{APP_LEM_R_CONV}

We write the remainder in the form
\begin{eqnarray*}
\Res(\Delta) & = & (\ThetaStar + \Delta)^{-1}- \invn{\ThetaStar} +
{\ThetaStar}^{-1} \Delta {\ThetaStar}^{-1}.
\end{eqnarray*}
By sub-multiplicativity of the $\matnorm{\cdot}{\infty}$ matrix norm,
for any two $\pdim \times \pdim$ matrices $A, B$, we have
$\matnorm{A\,B}{\infty} \le \matnorm{A}{\infty} \matnorm{B}{\infty}$,
so that
\begin{eqnarray}
\matnorm{\invn{\ThetaStar} \Delta}{\infty} & \leq &
\matnorm{\invn{\ThetaStar}}{\infty} \matnorm{\Delta}{\infty} \nonumber \\
\label{EqnFirstUpper}
& \leq & \KconCov \; \degmax \|\Delta\|_\infty \; < \; 1/3,
\end{eqnarray}
where we have used the definition of $\KconCov$, the fact that
$\Delta$ has at most $\degmax$ non-zeros per row/column, and our
assumption $\|\Delta\|_\infty \; < 1/(3 \KconCov)$.  Consequently, we
have the convergent matrix expansion
\begin{eqnarray*}
\label{EqnInvThetaExp1}
\inv{\ThetaStar + \Delta} &=& \inv{\opt{\conc}\big(I +
\invn{\opt{\conc}} \Delta\big)}\\
 & = & \invn{\big(I + \invn{\opt{\conc}} \Delta\big)}
            \inv{\opt{\conc}}\\
 & = & \sum_{k=0}^{\infty} (-1)^{k} \big(\invn{\opt{\conc}}
\Delta\big)^{k} \inv{\opt{\conc}} \\
& = & \invn{\opt{\conc}} - \invn{\opt{\conc}} \Delta
\invn{\opt{\conc}} + \sum_{k=2}^{\infty} (-1)^{k}
\big(\invn{\opt{\conc}} \Delta\big)^{k} \inv{\opt{\conc}} \\
& = & \invn{\opt{\conc}} - \invn{\opt{\conc}} \Delta
\invn{\opt{\conc}} + \invn{\opt{\conc}} \Delta \invn{\opt{\conc}}
\Delta J \invn{\opt{\conc}},
\end{eqnarray*}
where $J = \sum_{k=0}^{\infty} (-1)^{k} \big(\invn{\opt{\conc}}
\Delta\big)^{k}$.

We now prove the bound~\eqref{EqnRemBound} on the remainder as follows.
Let $e_i$ denote the unit vector with $1$ in position $i$ and zeroes
elsewhere.  From equation~\eqref{EqnRemExpand}, we have
\begin{eqnarray*}
\| \Rem(\Delta) \|_\infty & = & \max_{i, j} |e_i^T \invn{\ThetaStar}
\Delta \; \invn{\ThetaStar} \Delta J \invn{\ThetaStar} e_j| \\
& \leq & \max_i \|e_i^T \invn{\ThetaStar} \Delta \|_\infty \; \max_j
\|\invn{\ThetaStar} \Delta J \invn{\ThetaStar} e_j \|_1,
\end{eqnarray*}
which follows from the fact that for any vectors $a,b \in \real^{\pdim}$,
$|a^{T}b| \le \|a\|_{\infty} \|b\|_{1}$. This in turn can be simplified as,
\begin{eqnarray*}
\| \Rem(\Delta) \|_\infty  & \leq & \max_i \| e_i^T \invn{\ThetaStar} \|_1 \; \| \Delta\|_\infty
\; \max_j \|\invn{\ThetaStar} \Delta J \invn{\ThetaStar} e_j \|_1 
\end{eqnarray*}
since for any vector $u \in
\real^\pdim$, $\|u^T \Delta\|_\infty \leq \|u\|_1 \|\Delta\|_\infty$,
where $\|\Delta\|_\infty$ is the elementwise $\ell_\infty$-norm.
Continuing on, we have
\begin{eqnarray*}
\| \Rem(\Delta) \|_\infty & \leq & \matnorm{\invn{\ThetaStar}}{\infty} \;
\| \Delta\|_\infty \| \; \matnorm{\invn{\ThetaStar} \Delta J
\invn{\ThetaStar}}{1},
\end{eqnarray*}
where $\matnorm{A}{1} \defn \max_{\|x\|_1 = 1} \|A x\|_1$ is the
$\ell_1$-operator norm.  Since $\matnorm{A}{1} =
\matnorm{A^T}{\infty}$, we have
\begin{eqnarray}\label{EqnRtmp1}
\| \Rem(\Delta) \|_\infty & \leq & \| \Delta\|_\infty
\matnorm{\invn{\ThetaStar}}{\infty} \; \matnorm{\invn{\ThetaStar} J^T
\Delta \invn{\ThetaStar}}{\infty} \\
\nn& \leq & \|\Delta\|_\infty \; \KconCov
\matnorm{\invn{\ThetaStar}}{\infty}^2 \matnorm{J^T}{\infty}
\matnorm{\Delta}{\infty}
\end{eqnarray}
Recall that $J = \sum_{k=0}^{\infty} (-1)^{k} \big(\invn{\opt{\conc}}
\Delta\big)^{k}$.  By sub-multiplicativity of
$\matnorm{\cdot}{\infty}$ matrix norm, we have
\begin{equation*}
\matnorm{J^T}{\infty} \, \leq \, \sum_{k=0}^\infty \matnorm{\Delta
\invn{\opt{\conc}}}{\infty}^k \; \leq \; \frac{1}{1 -
\matnorm{\invn{\ThetaStar}}{\infty} \matnorm{\Delta}{\infty}} \, \leq
\, \frac{3}{2},
\end{equation*}
since $\matnorm{\invn{\ThetaStar}}{\infty} \matnorm{\Delta}{\infty} <
1/3$ from equation~\eqref{EqnFirstUpper}. Substituting this in
\eqref{EqnRtmp1}, we obtain
\begin{eqnarray*}
\| \Rem(\Delta) \|_\infty &\leq& \frac{3}{2} \|\Delta\|_\infty \; \KconCov
	\matnorm{\invn{\ThetaStar}}{\infty}^2
	\matnorm{\Delta}{\infty}\\
& \leq &  \frac{3}{2} \degmax \|\Delta\|_\infty^2 \; \KconCov^3,
\end{eqnarray*}
where the final line follows since $\matnorm{\Delta}{\infty} \leq \degmax
\|\Delta\|_\infty$, and since $\Delta$ has at most $\degmax$ non-zeroes
per row/column.


\section{Proof of Lemma~\ref{LEM_D_CONV}}
\label{APP_LEM_D_CONV}

By following the same argument as in Appendix~\ref{AppLemMLECharac},
we conclude that the restricted problem~\eqref{EqnRestricted} has a
unique optimum $\ThetaWit$.  Let $\ZWit$ be any member of the
sub-differential of $\ellreg{\cdot}$, evaluated at $\ThetaWit$.  By
Lagrangian theory, the witness $\ThetaWit$ must be an optimum of the
associated Lagrangian problem
\begin{equation*}
\min_{\Theta \succ 0, \; \Theta_{\EsetPlusComp} = 0} \big
\{\tracer{\Theta}{\SigHat} - \log \det(\Theta) + \regpar
\tracer{\Theta}{\ZWit} \big \}.
\end{equation*}
In fact, since this Lagrangian is strictly convex, $\ThetaWit$ is the
only optimum of this problem.  Since the log-determinant barrier
diverges as $\Theta$ approaches the boundary of the positive
semi-definite cone, we must have $\ThetaWit \succ 0$.  If we take
partial derivatives of the Lagrangian with respect to the unconstrained
elements $\Theta_\EsetPlus$, these partial derivatives must vanish at
the optimum, meaning that we have the zero-gradient condition
\begin{eqnarray}
\label{EqnCarefulZeroGrad}
G(\Theta_{\EsetPlus}) & = & -\invn{\Theta}_{\EsetPlus} +
\SigHat_{\EsetPlus} + \regpar \ZWit_{\EsetPlus} \; = \; 0.
\end{eqnarray}
To be clear, $\Theta$ is the $\pdim \times \pdim$ matrix with entries
in $\EsetPlus$ equal to $\Theta_{\EsetPlus}$ and entries in
$\EsetPlusComp$ equal to zero.  Since this zero-gradient condition is
necessary and sufficient for an optimum of the Lagrangian problem, it
has a unique solution (namely, $\ThetaWit_\EsetPlus$).

Our goal is to bound the deviation of this solution from
$\ThetaStar_{\EsetPlus}$, or equivalently to bound the deviation
$\Delta = \ThetaWit - \ThetaStar$.  Our strategy is to show the
existence of a solution $\Delta$ to the zero-gradient
condition~\eqref{EqnCarefulZeroGrad} that is contained inside the ball
$\Ball(\rad)$ defined in equation~\eqref{EqnDefnRad}.  By uniqueness
of the optimal solution, we can thus conclude that $\ThetaWit -
\ThetaStar$ belongs this ball.  In terms of the vector
$\myvec{\Delta}_\EsetPlus = \myvec{\Theta}_\EsetPlus -
\myvec{\ThetaStar}_\EsetPlus$, let us define a map $F:
\real^{|\EsetPlus|} \rightarrow \real^{|\EsetPlus|}$ via
\begin{eqnarray}
F(\myvec{\Delta}_{\EsetPlus}) & \defn & -
\inv{\BigHess_{\EsetPlus\EsetPlus}} \big (
\myvec{G}(\ThetaStar_{\EsetPlus} + \Delta_{\EsetPlus})\big) +
\myvec{\Delta}_{\EsetPlus},
\end{eqnarray}
where $\myvec{G}$ denotes the vectorized form of $G$.  Note that by
construction, \mbox{$F(\myvec{\Delta}_\EsetPlus) =
\myvec{\Delta}_\EsetPlus$} holds if and only if
$G(\ThetaStar_\EsetPlus + \Delta_{\EsetPlus}) = G(\Theta_{\EsetPlus})
= 0$.

We now claim that $F(\Ball(\rad)) \subseteq \Ball(\rad)$.  Since $F$
is continuous and $\Ball(\rad)$ is convex and compact, this inclusion
implies, by Brouwer's fixed point theorem~\cite{OrtegaR70}, that there
exists some fixed point $\myvec{\Delta}_\EsetPlus \in \Ball(\rad)$.
By uniqueness of the zero gradient condition (and hence fixed points
of $F$), we can thereby conclude that $\|\ThetaWit_\EsetPlus -
\ThetaStar_\EsetPlus\|_\infty \leq \rad$.

Let $\Delta \in \real^{\pdim \times \pdim}$ denote the zero-padded
matrix, equal to $\Delta_\EsetPlus$ on $\EsetPlus$ and zero on
$\EsetPlusComp$. By definition, we have
\begin{eqnarray}
G(\ThetaStar_\EsetPlus + \Delta_\EsetPlus) & = & -(\ThetaStar +
\Delta)^{-1}_{\EsetPlus} + \SigHat_{\EsetPlus} + \regpar
\ZWit_{\EsetPlus} \nonumber \\
& = & \big[-(\ThetaStar + \Delta)^{-1}_{\EsetPlus} +
\ThetaStar_{\EsetPlus} \big] + \big[ \SigHat_{\EsetPlus} -
(\ThetaStar)^{-1}_{\EsetPlus} \big] + \regpar \ZWit_\EsetPlus \nonumber \\
\label{EqnCarefulW}
& = & \big[-(\ThetaStar + \Delta)^{-1}_{\EsetPlus} +
\ThetaStar_{\EsetPlus} \big] + \Wnoise_{\EsetPlus} + \regpar
\ZWit_\EsetPlus,
\end{eqnarray}
where we have used the definition $\Wnoise = \SigHat - \CovMatStar$.

  For any $\Delta_\EsetPlus \in \Ball(\rad)$, we have
\begin{eqnarray}
\matnorm{\invn{\opt{\conc}} \Delta}{\infty} &\le &
\matnorm{\invn{\opt{\conc}}}{\infty}\matnorm{\Delta}{\infty} \nonumber
\\
\label{EqnCompBoundA}
& \leq & \KconCov \; \degmax \| \Delta \|_\infty,
\end{eqnarray}
where $\|\Delta\|_\infty$ denotes the elementwise $\ell_\infty$-norm
(as opposed to the $\ell_\infty$-operator norm
$\matnorm{\Delta}{\infty}$), and the inequality follows since $\Delta$
has at most $\degmax$ non-zero entries per row/column,

By the definition~\eqref{EqnDefnRad} of the radius $\rad$, and the
assumed upper bound~\eqref{EqnDconvAss}, we have $\|\Delta\|_\infty
\leq \rad \leq \frac{1}{3 \KconCov \degmax}$, so that the results of
Lemma~\ref{LEM_R_CONV} apply.  By using the
definition~\eqref{EqnRdefn} of the remainder, taking the vectorized
form of the expansion~\eqref{EqnRemExpand}, and restricting to entries
in $\EsetPlus$, we obtain the expansion
\begin{eqnarray}
\label{EqnMatrixExpand}
\smallvec \big(\inv{\ThetaStar + \Delta} - \invn{\opt{\conc}}
\big)_{\EsetPlus} + \BigHess_{\EsetPlus\EsetPlus}
\myvec{\Delta}_{\EsetPlus} & = & \smallvec \big((\invn{\opt{\conc}}
\Delta)^{2} J \invn{\opt{\conc}}\big)_{\EsetPlus}.
\end{eqnarray}

Using this expansion~\eqref{EqnMatrixExpand} combined with the
expression~\eqref{EqnCarefulW} for $G$, we have
\begin{eqnarray*}
F(\myvec{\Delta}_{\EsetPlus}) & = & -
\inv{\BigHess_{\EsetPlus\EsetPlus}} \myvec{G}(\ThetaStar_\EsetPlus +
\Delta_\EsetPlus) + \myvec{\Delta}_{\EsetPlus} \\
&= & \inv{\BigHess_{\EsetPlus\EsetPlus}} \smallvec \big\{
    \big[\inv{\ThetaStar + \Delta} -
    \invn{\opt{\conc}}\big]_{\EsetPlus} - W_{\EsetPlus} - \regpar
    \ZWit_{\EsetPlus}\big\} + \myvec{\Delta}_{\EsetPlus}\\
   & = & \underbrace{\inv{\BigHess_{\EsetPlus\EsetPlus}}
    \smallvec\big[(\invn{\opt{\conc}} \Delta)^{2} J
    \invn{\opt{\conc}}\big]_{\EsetPlus}} \; - \; \underbrace{
    \inv{\BigHess_{\EsetPlus\EsetPlus}} \big (
    \myvec{\Wnoise}_{\EsetPlus} + \regpar
    \myvec{\ZWit}_{\EsetPlus}\big )}. \\
& & \qquad \qquad \qquad \Term_1 \qquad \qquad \qquad \qquad \qquad
\qquad \Term_2
\end{eqnarray*}

The second term is easy to deal with: using the definition $\KconHess
= \matnorm{(\BigHess_{\EsetPlus\EsetPlus})^{-1}}{\infty}$, we have
$\|\Term_2\|_\infty \leq \KconHess \big( \|\Wnoise\|_\infty + \regpar
\big) \; = \; \rad/2$.  It now remains to show that
\mbox{$\|\Term_1\|_\infty \leq \rad/2$}.  We have
\begin{eqnarray*}
\|\Term_1\|_\infty & \leq & \KconHess \, \big\| \smallvec\big[
    (\invn{\opt{\conc}} \Delta)^2 J
    \invn{\opt{\conc}}\big]_{\EsetPlus} \big\|_\infty \nonumber \\
& \leq & \KconHess \| \Res(\Delta)\|_\infty,
\end{eqnarray*}
where we used the expanded form~\eqref{EqnRemExpand} of the remainder,
Applying the bound~\eqref{EqnRemBound} from Lemma~\ref{LEM_R_CONV}, we
obtain
\begin{eqnarray*}
\|\Term_1\|_\infty & \leq & \frac{3}{2} \degmax \KconCov^3 \KconHess
\; \|\Delta\|_\infty^2 \; \leq \; \frac{3}{2} \degmax \KconCov^3
\KconHess \; \rad^2.
\end{eqnarray*}
Since $\rad \leq \frac{1}{3 \KconCov^3 \KconHess
\degmax}$ by assumption~\eqref{EqnDconvAss}, we conclude that
\begin{eqnarray*}
\|\Term_1\|_\infty & \leq & \frac{3}{2} \degmax \KconCov^3 \KconHess
\; \frac{1}{3 \KconCov^3 \KconHess \degmax} \rad \; = \; \rad/2,
\end{eqnarray*}
thereby establishing the claim.


\section{Proof of Lemma~\ref{LEM_SAM_COV_BOUND_SUBG}}
\label{APP_LEM_SAM_COV_BOUND_SUBG}

For each pair $(i,j)$ and $\nu > 0$, define the event
\begin{eqnarray*}
\Aij(\nu) & \defn & \big \{|\frac{1}{\numobs}
\sum_{\obsind=1}^{\numobs} \SampVar_i \SampVar_j- \CovMatStar_{ij}| >
\nu \big \}.
\end{eqnarray*}
As the sub-Gaussian assumption is imposed on the variables
$\{\SampVar_{i}\}$ directly, as in Lemma~A.3 of
\citet{BickelLevina2008}, our proof proceeds by first decoupling the
products $\SampVar_{i} \SampVar_{j}$.  For each pair $(i,j)$, we
define $\rhoStar_{ij} =
\CovMatStar_{ij}/\sqrt{\CovMatStar_{ii}\CovMatStar_{jj}}$, and the
rescaled random variables \mbox{$\Xsambar{\obsind}_{i} \defn
\Xsam{\obsind}_{ij}/\sqrt{\CovMatStar_{ii}}$.}  Noting that the strict
positive definiteness of $\CovMatStar$ implies that $|\rhoStar_{ij}| <
1$, we can also define the auxiliary random variables
\begin{equation}
\label{EqnUVDefn}
\newuvar{\obsind}_{ij} \defn \Xsambar{\obsind}_i + \Xsambar{\obsind}_j
\quad \mbox{and} \quad \newvvar{\obsind}_{ij} \defn
\Xsambar{\obsind}_i -\Xsambar{\obsind}_j.
\end{equation}
With this notation, we then claim:

\blems
\label{LemMartin}
Suppose that each $\Xsambar{\obsind}_i$ is sub-Gaussian with parameter
$\sigma$.  Then for each node pair $(i,j)$, the following properties
hold:
\begin{enumerate}
\item[(a)] For all $\obsind = 1, \ldots, \numobs$, the random
variables $\newuvar{\obsind}_{ij}$ and $\newvvar{\obsind}_{ij}$ are
sub-Gaussian with parameters $2 \sigma$.
\item[(b)] For all $\nu > 0$, the probability $\mprob[\Aij(\nu)]$ is
upper bounded by
\begin{equation*}
\mprob \big [|\sum_{\obsind=1}^{\numobs} (\newuvar{\obsind}_{ij})^{2}
  - 2(1 -\rhoStar_{ij})| > \frac{2 \numobs \nu}{
    \sqrt{\CovMatStar_{ii} \CovMatStar_{jj}}} \big ] + \mprob
\big[|\sum_{\obsind=1}^{\numobs} (\newvvar{\obsind}_{ij})^{2} -
  2(1-\rhoStar_{ij}) | > \frac{2 \numobs \nu}{\sqrt{\CovMatStar_{ii}
      \CovMatStar_{jj}}} \big].
\end{equation*}
\end{enumerate}
\elems
\begin{proof}
(a) For any $r \in \real$, we have
\begin{eqnarray*}
\E[\exp(r \, \newuvar{\obsind}_{ij})] & = & \E \Big[\exp \big( r\,
\Xsambar{\obsind}_i \big) \; \exp \big(r \Xsambar{\obsind}_j\big)
\Big] \; \leq \; \E \Big[\exp \big( 2 r\, \Xsambar{\obsind}_i\big)
\Big]^{1/2} \; \Big[\exp \big(2 r \Xsambar{\obsind}_j\big)
\Big]^{1/2},
\end{eqnarray*}
where we have used the Cauchy-Schwarz inequality.  Since the variables
$\Xsambar{\obsind}_i$ and $\Xsambar{\obsind}_j$ are sub-Gaussian with
parameter $\csubg$, we have
\begin{eqnarray*}
\E \Big[\exp \big( 2 r\, \Xsambar{\obsind}_i\big) \Big]^{1/2} \; \E
\Big[\exp \big(2 r \Xsambar{\obsind}_j\big) \Big]^{1/2} & \leq & \exp(
\csubg^2 \frac{r^2}{2}) \; \exp( \csubg^2 \frac{r^2}{2}),
\end{eqnarray*}
so that $\newuvar{\obsind}_{ij}$ is sub-Gaussian with parameter $2
\csubg$ as claimed. \\
(b) By straightforward algebra, we have the decomposition
\begin{equation*}
\sum_{\obsind=1}^{\numobs} (\Xsambar{\obsind}_i \Xsambar{\obsind}_j -
\rhoStar_{ij}) \; = \; \big \{
\frac{1}{4}\sum_{i=1}^{n}\big\{(\Xsambar{\obsind}_i +
\Xsambar{\obsind}_j)^{2} - 2 (1 + \rhoStar_{st})\big\} \; - \; \big
\{\frac{1}{4}\sum_{i=1}^{n}\big\{(\SampVarStar_s - \SampVarStar_t)^{2}
- 2 (1 - \rhoStar_{st}) \big \}.
\end{equation*}
By union bound, we obtain that $\mprob[\Aij(\nu)]$ is upper bounded by
\begin{equation}
\label{EqnSampCovSumIneq}
\mprob \Big[ \big|\sum_{\obsind=1}^\numobs (\newuvar{\obsind}_{ij})^2
  - 2 \, (1+\rhoStar_{ij}) \big| \geq \frac{4 \numobs \nu}{2 \,
    \sqrt{\CovMatStar_{ii} \CovMatStar_{jj}}} \Big] + \mprob \Big[
  \big|\sum_{\obsind=1}^\numobs (\newvvar{\obsind}_{ij})^2 - 2 \,
  (1-\rhoStar_{ij}) \big| \geq \frac{4 \numobs \nu}{2 \,
    \sqrt{\CovMatStar_{ii} \CovMatStar_{jj}}} \Big],
\end{equation}
which completes the proof of Lemma~\ref{LemMartin}(b).
\end{proof}

It remains to control the terms $\sum_{\obsind=1}^\numobs
(\newuvar{\obsind}_{ij})^2$ and $\sum_{\obsind=1}^\numobs
(\newvvar{\obsind}_{ij})^2$.  We do so by exploiting tail
bounds~\cite{BulKoz} for sub-exponential random variables.  A
zero-mean random variable $Z$ is said to be \emph{sub-exponential} if
there exists a constant $\csubone \in (0, \infty)$ and $\csubtwo \in
(0,\infty]$ such that
\begin{eqnarray}
\label{EqnDefnPregauss}
\E[\exp(t Z)] & \leq & \exp(\csubone^2 \, t^2/2) \qquad \mbox{for all
$t \in (-\csubtwo, \csubtwo)$.}
\end{eqnarray}
Note that for $\csubtwo < +\infty$, this requirement is a weakening of
sub-Gaussianity, since the inequality is only required to hold on the
interval $(-\csubtwo, +\csubtwo)$.  

Now consider the variates $Z_{\obsind;ij} \defn
(\newuvar{\obsind}_{ij})^2 - 2 \, (1+\rhoStar_{ij})$. Note that they
are zero-mean; we also claim they are sub-exponential.
\blems\label{LemSampDevPreG} For all $\obsind \in
\{1,\hdots,\numobs\}$ and node-pairs $(i,j) \in \vertex \times
\vertex$, the variables 
\begin{eqnarray*}
Z_{\obsind;ij} & \defn &  (\newuvar{\obsind}_i)^2- 2
\, (1+\rhoStar_{ij})
\end{eqnarray*}
are sub-exponential with parameter $\csubone_U = 16(1 + 4 \csubg^2)$
in the interval $(-\csubtwo_U,\csubtwo_U)$, with \mbox{$\csubtwo_U =
1/(16(1 + 4 \csubg^2))$.}
\elems

We can exploit this lemma to apply tail bounds for sums of
i.i.d. sub-exponential variates (Thm. 5.1,~\cite{BulKoz}).  Doing so
yields that for $t \leq \csubone_U^2 \csubtwo_U$, we have $\mprob
\biggr [|\sum_{\obsind=1}^\numobs (\newuvar{\obsind}_{ij})^{2} - 2
(1+\rhoStar_{ij}) | > \numobs t] \leq 2 \exp \big \{ -\numobs
t^2/(2\csubone^2_U) \big \}$.  Setting $t = 2
\nu/\max_{i}\CovMatStar_{ii}$, and noting that $(2 \numobs
\nu/\sqrt{\CovMatStar_{ii}\CovMatStar_{jj}}) \ge (2 \numobs
\nu/\max_{i}\CovMatStar_{ii})$, we obtain
\begin{eqnarray*}
\mprob \biggr [|\sum_{\obsind=1}^\numobs (\newuvar{\obsind}_{ij})^{2}
  - 2 (1+\rhoStar_{ij}) | > \frac{2 \numobs
    \nu}{\sqrt{\CovMatStar_{ii}\CovMatStar_{jj}}}\biggr] & \leq & 
2 \exp
\biggr \{ -\frac{2 \numobs \nu^2}{\max_{i}(\CovMatStar_{ii})^2 \, \csubone^2_U} \biggr \} \\
& \leq & 2 \exp \biggr \{- \frac{\numobs \nu^2}{ \max_{i}
 (\CovMatStar_{ii})^2 \, 128 \, ( 1 + 4\csubg^2)^2} \biggr \},
\end{eqnarray*}
for all $\nu \leq 8 (\max_{i}\CovMatStar_{ii})\,(1 + 4 \csubg^2)$.
A similar argument yields the same tail bound for the deviation
involving $\newvvar{\obsind}_{ij}$.  Consequently, using
Lemma~\ref{LemMartin}(b), we conclude that
\begin{eqnarray*}
\mprob[\Aij(\nu)] & \leq & 4 \exp \big \{- \frac{\numobs \nu^2}{ \max_{i}
	 (\CovMatStar_{ii})^2 \, 128 \, ( 1 + 4\csubg^2)^2} \big \}.,
\end{eqnarray*}
valid for $\nu \leq 8 (\max_{i}\CovMatStar_{ii})\,(1 + 4 \csubg^2)$,
as required. It only remains to prove Lemma~\ref{LemSampDevPreG}.
\begin{proof}[Proof of Lemma~\ref{LemSampDevPreG}]
If we can obtain a bound $B > 0$ such that
\begin{eqnarray*}
\sup_{m \ge 2} \left[ \frac{\E(|Z_{\obsind;ij}|^{m}}{m!}\right]^{1/m}
& \leq  & B,
\end{eqnarray*}
it then follows (Thm.~3.2,~\cite{BulKoz}) that $Z_{\obsind;ij}$ is
sub-exponential with parameter $2 B$ in the interval $(-\frac{1}{2B},
+ \frac{1}{2B})$. We obtain such a bound $B$ as follows. Using the
inequality $(a + b)^{m} \le 2^{m}(a^m + b^m)$, valid for any real
numbers $a,b$, we have
\begin{eqnarray}\label{EqnZMoma}
\Exs(|Z_{\obsind;ij}|^{m}) \le
2^{m}\,\left(\Exs(|\newuvar{\obsind}_{ij}|^{2m}) + (2(1 +
\rhoStar_{ij}))^m\right).
\end{eqnarray}
Recalling that $\newuvar{\obsind}_{ij}$ is sub-Gaussian with parameter
$2\csubg$, from Lemma~1.4 of~\citet{BulKoz} regarding the moments of
sub-Gaussian variates, we have $\Exs[|\newuvar{\obsind}_{ij}|^{2m}]
\le 2 (2m/e)^{m} (2 \csubg)^{2m}$. Thus, noting the inequality $m! \ge
(m/e)^m$, it follows that $\Exs[|\newuvar{\obsind}_{ij}|^{2m}]/m! \le
2^{3m + 1} \csubg^{2m}$. It then follows from
equation~\eqref{EqnZMoma} that
\begin{eqnarray*}
\left[\frac{\E(|Z_{\obsind;ij}|^{m})}{m!}\right]^{1/m} &\le&
			2^{1/m}\left((2^{4m + 1} \csubg^{2m})^{1/m} +
			\frac{4(1 +
			\rhoStar_{ij})}{(m!)^{1/m}}\right)\\ &\le&
			2^{1/m}\left(2^{1/m}\, 16 \, \csubg^{2} +
			\frac{4(1 +
			\rhoStar_{ij})}{(m!)^{1/m}}\right),
\end{eqnarray*}
where we have used the inequality $(x + y)^{1/m} \le 2^{1/m}(x^{1/m} +
y^{1/m})$, valid for any integer $m \in \mathbb{N}$ and real numbers
$x,y > 0$.  Since the bound is a decreasing function of $m$, it
follows that
\begin{eqnarray*}
	\sup_{m \ge 2} \left[
	\frac{\E(|Z_{\obsind;ij}|^{m}}{m!}\right]^{1/m} &\le&
	2^{1/2}\left(2^{1/2}\, 16 \, \csubg^{2} + \frac{4(1 +
	\rhoStar_{ij})}{(2)^{1/2}}\right)\\ &\le& 32 \csubg^2 + 8 =
	8(1 + 4 \csubg^2),
\end{eqnarray*}
where we have used the fact that $|\rhoStar_{ij}| \leq 1$. The claim
of the lemma thus follows.
\end{proof}


\section{Proof of Lemma~\ref{LEM_SAM_COV_BOUND_MOMENT}}
\label{APP_LEM_SAM_COV_BOUND_MOMENT}

Define the random variables $\SamDev{\obsind} = \Xsam{\obsind}_i
\Xsam{\obsind}_j - \CovMatStar_{ij}$, and note that they have mean
zero.  By applying the Chebyshev inequality, we obtain
\begin{eqnarray}
\mprob\big[\Big|\sum_{\obsind=1}^{\numobs} \SamDev{\obsind} \Big| >
  \numobs\,\nu \big] & = & \mprob\big[
  \big(|\sum_{\obsind=1}^{\numobs} \SamDev{\obsind} \big)^{2
  \momentpow} > (\numobs\,\nu)^{2 \momentpow} \big] \nonumber \\
\label{EqnSamDevChebyshev}
& \leq & \frac{\E \big[\big( \sum_{\obsind=1}^{\numobs}
  \SamDev{\obsind} \big)^{2\momentpow}\,\big]}{\numobs^{2\momentpow}
  \,\nu^{2\momentpow}}.
\end{eqnarray}
Letting $\mathcal{A} = \{(a_1,\hdots,a_\numobs) \, \mid \, a_i \in
\{0,\hdots,2\momentpow\},\,\sum_{i=1}^{\numobs} a_i = 2\momentpow\}$,
by the multinomial theorem, we have
\begin{eqnarray*}
\E \big[\big(\sum_{\obsind=1}^{\numobs} \SamDev{\obsind}
\big)^{2\momentpow}\,\big] & = & \E \Big[\sum_{{\bf a} \in
\mathcal{A}} \binom{2\momentpow}{a_1,\hdots,a_{n}}
\prod_{\obsind=1}^{\numobs} (\SamDev{\obsind})^{a_\obsind}\Big] \\
& = & \sum_{a \in \mathcal{A}}
\binom{2\momentpow}{a_1,\hdots,a_{\numobs}}
\prod_{\obsind=1}^{\numobs}
\E\big[(\SamDev{\obsind})^{a_\obsind}\big],
\end{eqnarray*}
where the final equality uses linearity of expectation, and the
independence of the variables $\{\SamDev{\obsind}
\}_{\obsind=1}^\numobs$.

Since the variables $\SamDev{\obsind}$ are zero-mean, the product
$\prod_{\obsind=1}^{\numobs}
\E\big[(\SamDev{\obsind})^{a_\obsind}\big]$ vanishes for any
multi-index $a\in \mathcal{A}$ such that $a_\obsind= 1$ for at least
one $\obsind$.  Accordingly, defining the subset
\begin{equation*}
\mathcal{A}_{-1} \defn \{(a_1,\hdots,a_\numobs) \, \mid \, a_i \in
\{0,2,\hdots,2\momentpow\},\\\,\sum_{i=1}^{n} a_i = 2\momentpow\},
\end{equation*}
we have
\begin{eqnarray}
\label{EqnWmoma}
\E \big[\big(\sum_{\obsind=1}^{\numobs} \SamDev{\obsind}
\big)^{2\momentpow}\,\big] &= & \sum_{ a \in \mathcal{A}_{-1}}
\binom{2\momentpow}{a_1,\hdots,a_{\numobs}}
\prod_{\obsind=1}^{\numobs} \E\big[
(\SamDev{\obsind})^{a_\obsind}\big] \nonumber \\
& \leq & \underbrace{\big(\sum_{a \in \mathcal{A}_{-1}}
\binom{2\momentpow}{a_1,\hdots,a_{\numobs}}\big)} \quad
\underbrace{\max_{a \in \mathcal{A}_{-1}} \prod_{\obsind=1}^{\numobs}
\E\big[(\SamDev{\obsind})^{a_\obsind}\big]}. \\
& & \qquad \qquad \Term_1 \qquad \qquad \qquad \qquad \qquad \Term_2
\nonumber
\end{eqnarray}

The quantity $\Term_1$ is equal to the number of ways to put
$2\momentpow$ balls in $\numobs$ bins such that if a bin contains a
ball, it should have at least two balls.  Note that this implies there
can then be at most $\momentpow$ bins containing a ball. Consequently,
the term $\Term_1$ is bounded above by the product of the number of
ways in which we can choose $\momentpow$ out of $\numobs$ bins, and
the number of ways in which we can put $2\momentpow$ balls into $m$
bins---viz. 
\begin{eqnarray*}
\Term_1 & \leq & \binom{\numobs}{\momentpow} \quad
\momentpow^{2\momentpow} \; \leq \; \numobs^{\momentpow} \;
\momentpow^{2\momentpow}.
\end{eqnarray*}

Turning now to the second term $\Term_2$, note the following
inequality: for any numbers $(v_1, \ldots, v_\ell) \in
\real_{+}^{\ell}$ and non-negative integers $(a_1, \ldots, a_\ell)$,
we have
\begin{eqnarray*}
\prod_{\obsind = 1}^{\ell} v_{\obsind}^{a_\obsind} & \leq &
(\max_{\obsind = 1, \ldots, \ell} v_{\obsind})^{a_+} \; \leq \;
\sum_{\obsind=1}^{\ell} v_{\obsind}^{a_+}, \qquad \mbox{where $a_+
\defn \sum \limits_{\obsind=1}^{\ell} a_\obsind$.}
\end{eqnarray*}
 Using this inequality, for any $a \in \mathcal{A}_{-1}$, we have
\begin{eqnarray*}
\prod_{\obsind=1}^{\numobs} \E
\big[(\SamDev{\obsind})^{a_\obsind}\big] & \leq &
\prod_{\obsind=1}^{\numobs }\E\big[|\SamDev{\obsind}|^{a_\obsind}\big]
\\
& \leq & \sum_{ \{ \obsind \, \mid \, a_\obsind \neq 0 \} } \E
\big[(\SamDev{\obsind})^{2\momentpow} \big] \\
& \leq & \momentpow \; \max_{\obsind \in \{1, \ldots, \numobs\}} \E
\big[(\SamDev{\obsind})^{2\momentpow} \big],
\end{eqnarray*}
where the last inequality follows since any multi-index $a \in
\mathcal{A}_{-1}$ has at most $\momentpow$ non-zero entries.  Thus, we
have shown that $\Term_2 \leq \momentpow \max_{\obsind \in \{1,
\ldots, \numobs \}} \E \big[(\SamDev{\obsind})^{2\momentpow}\big]$.

Substituting our bounds on $\Term_1$ and $\Term_2$ into
equation~\eqref{EqnWmoma}, we obtain
\begin{eqnarray}
\label{EqnWmomb}
\E \big[\big|\sum_{\obsind=1}^{\numobs} \big(\SamDev{\obsind}
\big)^{2\momentpow}\,\big] & \leq & \numobs^{\momentpow} \;
\momentpow^{2\momentpow+1} \; \max_{\obsind \in \{1,\ldots, \numobs
\}} \E \big[(\SamDev{\obsind})^{2\momentpow}\big].
\end{eqnarray}
It thus remains to bound the moments of $\SamDev{\obsind}$. 
We have
\begin{eqnarray*}
\E \big[ (\SamDev{\obsind})^{2\momentpow}\big] & \leq &
\E\big[(\Xsam{\obsind}_i \Xsam{\obsind}_j -
\CovMatStar_{ij})^{2\momentpow}\big] \; \leq \; 2^{2\momentpow} \big\{
\E[\big(\Xsam{\obsind}_i \Xsam{\obsind}_j \big)^{2\momentpow}] +
[\CovMatStar_{ij}]^{2\momentpow} \big\},
\end{eqnarray*}
where we have used the inequality $(a+b)^{2\momentpow} \; \leq \;
2^{2\momentpow} (a^{2\momentpow} + b^{2\momentpow})$, valid for all
real numbers $a$ and $b$.  An application of the Cauchy-Schwarz
inequality yields
\begin{eqnarray*}
\E \big[ (\SamDev{\obsind})^{2\momentpow} \big] & \leq &
2^{2\momentpow} \big\{ \sqrt{ \E [\big(
\Xsam{\obsind}_{i}\big)^{4\momentpow}]\,\E [\big( \Xsam{\obsind}_{j}
\big)^{4\momentpow}]} + [\CovMatStar_{ij}]^{2\momentpow} \big\} \\
& \leq & 2^{2\momentpow} \big( \KconMom [\CovMatStar_{ij}
\CovMatStar_{ij}]^{\momentpow} + [ \CovMatStar_{oj}]^{2\momentpow}
\big),
\end{eqnarray*}
where we have used the assumed moment bound
$\E[\big(\Xsam{\obsind}_i/\sqrt{\CovMatStar_{ii}}\big)^{4\momentpow}]
\le \KconMom$. Equation~\eqref{EqnWmomb} thus reduces to
\begin{eqnarray*}
\E \big[\big|\sum_{\obsind=1}^{\numobs} \SamDev{\obsind}
\big|^{2\momentpow}\,\big] & \leq &
\numobs^{\momentpow}\,\momentpow^{2\momentpow+1}\,2^{2\momentpow}\big(
\KconMom [\CovMatStar_{ii}\CovMatStar_{jj}]^{\momentpow} +
[\CovMatStar_{ij}]^{2\momentpow} \big).
\end{eqnarray*}
Substituting back into equation~\eqref{EqnSamDevChebyshev} yields
\begin{eqnarray*}
\mprob\big[ \Big|\frac{1}{\numobs} \sum_{\obsind=1}^{\numobs}
\Big(\Xsam{\obsind}_i \Xsam{\obsind}_j - \CovMatStar_{ij}\Big)\Big| >
\nu \big] & \leq & \frac{\momentpow^{2\momentpow+1}
\numobs^{\momentpow} 2^{2\momentpow}\big( \KconMom
[\CovMatStar_{ii}\CovMatStar_{jj}]^{\momentpow} +
[\CovMatStar_{ij}]^{2\momentpow} \big) }{\numobs^{2\momentpow}\,
\nu^{2\momentpow}}\\ 
& = & \frac{\big\{\momentpow^{2\momentpow+1} 2^{2\momentpow} \,
(\KconMom [\CovMatStar_{ii}\CovMatStar_{jj}]^{\momentpow} +
[\CovMatStar_{ij}]^{2\momentpow} ) \big\}}{\numobs^{\momentpow}\,
\nu^{2\momentpow}}.
\end{eqnarray*}

Noting that $\CovMatStar_{ij}$, $\CovMatStar_{jj}$, and
$\CovMatStar_{ij}$ are all bounded above by $\max_{i}
\CovMatStar_{ii}$, we obtain
\begin{eqnarray*}
\mprob\big[ \Big|\frac{1}{\numobs} \sum_{\obsind=1}^{\numobs}
\Big(\Xsam{\obsind}_i \Xsam{\obsind}_j - \CovMatStar_{ij}\Big)\Big| >
\nu \big] & \leq & \frac{\big\{\momentpow^{2\momentpow+1}
2^{2\momentpow} (\max_i \CovMatStar_{ii}) ^{2\momentpow}\, (\KconMom +
1 ) \big\}}{\numobs^{\momentpow}\, \nu^{2\momentpow}},
\end{eqnarray*}
as claimed.


\bibliographystyle{plainnat}

\bibliography{covsel}

\begin{thebibliography}{27}
\providecommand{\natexlab}[1]{#1}
\providecommand{\url}[1]{\texttt{#1}}
\expandafter\ifx\csname urlstyle\endcsname\relax
  \providecommand{\doi}[1]{doi: #1}\else
  \providecommand{\doi}{doi: \begingroup \urlstyle{rm}\Url}\fi

\bibitem[Bickel and Levina(2008{\natexlab{a}})]{BickelLevina2007}
P.~J. Bickel and E.~Levina.
\newblock Covariance regularization by thresholding.
\newblock \emph{Ann. Statist.}, 2008{\natexlab{a}}.

\bibitem[Bickel and Levina(2008{\natexlab{b}})]{BickelLevina2008}
P.J. Bickel and E.~Levina.
\newblock Regularized estimation of large covariance matrices.
\newblock \emph{Ann. Statist.}, 36\penalty0 (1):\penalty0 199--227,
  2008{\natexlab{b}}.

\bibitem[Boyd and Vandenberghe(2004)]{Boyd02}
S.~Boyd and L.~Vandenberghe.
\newblock \emph{Convex optimization}.
\newblock Cambridge University Press, Cambridge, UK, 2004.

\bibitem[Bregman(1967)]{Bregman67a}
L.~M. Bregman.
\newblock The relaxation method for finding the common point of convex sets and
  its application to the solution of problems in convex programming.
\newblock \emph{USSR Computational Mathematics and Mathematical Physics},
  7:\penalty0 191--204, 1967.

\bibitem[Brown(1986)]{Brown86}
L.D. Brown.
\newblock \emph{Fundamentals of statistical exponential families}.
\newblock Institute of Mathematical Statistics, Hayward, CA, 1986.

\bibitem[Buldygin and Kozachenko(2000)]{BulKoz}
V.~V. Buldygin and Y.~V. Kozachenko.
\newblock \emph{Metric characterization of random variables and random
  processes}.
\newblock American Mathematical Society, Providence, RI, 2000.

\bibitem[Censor and Zenios(1988)]{Censor}
Y.~Censor and S.~A. Zenios.
\newblock \emph{Parallel Optimization: Theory, Algorithms, and Applications}.
\newblock Numerical Mathematics and Scientific Computation. Oxford University
  Press, 1988.

\bibitem[d'Aspr\'{e}mont et~al.(2008)d'Aspr\'{e}mont, Banerjee, and
  Ghaoui]{AspreBanG2008}
A.~d'Aspr\'{e}mont, O.~Banerjee, and L.~El Ghaoui.
\newblock First-order methods for sparse covariance selection.
\newblock \emph{SIAM J. Matrix Anal. Appl.}, 30\penalty0 (1):\penalty0 56--66,
  2008.

\bibitem[{El Karoui}(2008)]{Karoui2007}
N.~{El Karoui}.
\newblock Operator norm consistent estimation of large dimensional sparse
  covariance matrices.
\newblock \emph{Ann. Statist.}, To appear, 2008.

\bibitem[Friedman et~al.(2007)Friedman, Hastie, and
  Tibshirani]{FriedHasTib2007}
J.~Friedman, T.~Hastie, and R.~Tibshirani.
\newblock Sparse inverse covariance estimation with the graphical {L}asso.
\newblock \emph{Biostat.}, 9\penalty0 (3):\penalty0 432--441, 2007.

\bibitem[Furrer and Bengtsson(2007)]{Furrer2007}
R.~Furrer and T.~Bengtsson.
\newblock Estimation of high-dimensional prior and posterior covariance
  matrices in kalman filter variants.
\newblock \emph{J. Multivar. Anal.}, 98\penalty0 (2):\penalty0 227--255, 2007.

\bibitem[Horn and Johnson(1985)]{Horn1985}
R.~A. Horn and C.~R. Johnson.
\newblock \emph{Matrix Analysis}.
\newblock Cambridge University Press, Cambridge, 1985.

\bibitem[Huang et~al.(2006)Huang, Liu, Pourahmadi, and Liu]{Huang2006}
J.Z. Huang, N.~Liu, M.~Pourahmadi, and L.~Liu.
\newblock Covariance matrix selection and estimation via penalised normal
  likelihood.
\newblock \emph{Biometrika}, 93\penalty0 (1):\penalty0 85--98, 2006.

\bibitem[Johnstone(2001)]{Johnstone2001}
I.~M. Johnstone.
\newblock On the distribution of the largest eigenvalue in principal components
  analysis.
\newblock \emph{Ann. Statist.}, 29\penalty0 (2):\penalty0 295--327, 2001.

\bibitem[Johnstone and Lu(2004)]{JohnstoneLu2004}
I.~M. Johnstone and A.~Y. Lu.
\newblock Sparse principal components analysis.
\newblock \emph{Unpublished Manuscript}, 2004.

\bibitem[Ledoit and Wolf(2003)]{LedoitWolf2003}
O.~Ledoit and M.~Wolf.
\newblock A well-conditioned estimator for large-dimensional covariance
  matrices.
\newblock \emph{J. Multivar. Anal.}, 88:\penalty0 365–411, 2003.

\bibitem[Ledoux(2001)]{Ledoux01}
M.~Ledoux.
\newblock \emph{The {C}oncentration of {M}easure {P}henomenon}.
\newblock Mathematical Surveys and Monographs. American Mathematical Society,
  Providence, RI, 2001.

\bibitem[Meinshausen(2008)]{Meins2008}
N.~Meinshausen.
\newblock A note on the {L}asso for graphical {G}aussian model selection.
\newblock \emph{Statistics and Probability Letters}, 78\penalty0 (7):\penalty0
  880--884, 2008.

\bibitem[Meinshausen and B\"{u}hlmann(2006)]{MeinsBuhl2006}
N.~Meinshausen and P.~B\"{u}hlmann.
\newblock High-dimensional graphs and variable selection with the {L}asso.
\newblock \emph{Ann. Statist.}, 34\penalty0 (3):\penalty0 1436--1462, 2006.

\bibitem[Ortega and Rheinboldt(1970)]{OrtegaR70}
J.~M. Ortega and W.~G. Rheinboldt.
\newblock \emph{Iterative Solution of Nonlinear Equations in Several
  Variables}.
\newblock Academic Press, NY, 1970.

\bibitem[Rothman et~al.(2008)Rothman, Bickel, Levina, and Zhu]{Rothman2007}
A.J. Rothman, P.J. Bickel, E.~Levina, and J.~Zhu.
\newblock Sparse permutation invariant covariance estimation.
\newblock \emph{Electron. J. Statist.}, 2:\penalty0 494--515, 2008.

\bibitem[Tropp(2006)]{Tropp2006}
J.~A. Tropp.
\newblock Just relax: {C}onvex programming methods for identifying sparse
  signals.
\newblock \emph{IEEE Trans. Info. Theory}, 51\penalty0 (3):\penalty0
  1030--1051, 2006.

\bibitem[Wainwright(2006)]{Wainwright2006_new}
M.~J. Wainwright.
\newblock Sharp thresholds for high-dimensional and noisy recovery of sparsity
  using the {L}asso.
\newblock Technical Report 709, UC Berkeley, May 2006.
\newblock To appear in IEEE Trans. Info. Theory.

\bibitem[Wu and Pourahmadi(2003)]{Wu2003}
W.~B. Wu and M.~Pourahmadi.
\newblock Nonparametric estimation of large covariance matrices of longitudinal
  data.
\newblock \emph{Biometrika}, 90\penalty0 (4):\penalty0 831--844, 2003.

\bibitem[Yuan and Lin(2007)]{YuanLin2007}
M.~Yuan and Y.~Lin.
\newblock Model selection and estimation in the {G}aussian graphical model.
\newblock \emph{Biometrika}, 94\penalty0 (1):\penalty0 19--35, 2007.

\bibitem[Zhao and Yu(2006)]{Zhao06}
P.~Zhao and B.~Yu.
\newblock On model selection consistency of {L}asso.
\newblock \emph{Journal of Machine Learning Research}, 7:\penalty0 2541--2567,
  2006.

\bibitem[Zhou et~al.(2008)Zhou, Lafferty, and Wasserman]{ZhoLafWas08}
S.~Zhou, J.~Lafferty, and L.~Wasserman.
\newblock Time-varying undirected graphs.
\newblock In \emph{21st Annual Conference on Learning Theory (COLT)}, Helsinki,
  Finland, July 2008.

\end{thebibliography}


\end{document}